\documentclass[10pt,journal,compsoc]{IEEEtran}
\ifCLASSOPTIONcompsoc
  \usepackage[nocompress]{cite}
\else
  \usepackage{cite}
\fi

\def\equationautorefname~#1\null{Equation~(#1)\null}
\def\appendixautorefname~#1\null{Appendix~#1\null}
\def\subappendixautorefname~#1\null{Appendix~#1\null}
\def\sectionautorefname~#1\null{Section~#1\null}
\def\subsectionautorefname~#1\null{Section~#1\null}
\def\figureautorefname~#1\null{Figure~#1\null}
\def\tableautorefname~#1\null{Table~#1\null}
\def\observationautorefname~#1\null{Observation~#1\null}
\def\algorithmautorefname~#1\null{Algorithm~#1\null}

\ifCLASSINFOpdf
\else
\fi
\usepackage{amssymb}
\usepackage{array}
\usepackage{placeins}
\usepackage{amsmath,bm}

\usepackage{array}
\usepackage{amsmath}
\usepackage{amsthm}
\usepackage{amsfonts}
\usepackage{booktabs}
\usepackage{algorithm}
\usepackage{algorithmic}

\PassOptionsToPackage{table}{xcolor}
\usepackage{xcolor}
\usepackage{graphicx}  
\usepackage{dsfont}

\usepackage{url}
\usepackage{enumerate}
\usepackage{multirow}
\usepackage{booktabs}  
\usepackage{makecell}
\usepackage{bm}
\usepackage{pifont}
\usepackage{threeparttable} 
\usepackage{ragged2e}  
\usepackage[edges]{forest}
\usepackage{tikz}
\usepackage{verbatim}
\usepackage{multicol}
\usepackage[numbers]{natbib}
\usepackage{tabularx}
\usepackage{subfigure}
\usepackage{pifont}

\definecolor{mygreen}{RGB}{0,150,80}
\definecolor{mypurple}{RGB}{0,153,204}
\definecolor{output-black}{RGB}{122,122,122}
\definecolor{myblue_1}{RGB}{230, 255, 255} 
\definecolor{mycolor_0}{RGB}{230, 225, 200}
\definecolor{mycolor_1}{RGB}{210, 240, 230}
\definecolor{mycolor_2}{RGB}{190, 220, 230}
\definecolor{mycolor_3}{RGB}{200, 215, 240}
\definecolor{mycolor_4}{RGB}{200, 224, 214}
\definecolor{mycolor_5}{RGB}{190, 230, 240}
\definecolor{mycolor_6}{RGB}{230, 220, 255}
\definecolor{mycolor_7}{RGB}{177, 225, 175}
\definecolor{mycolor_8}{RGB}{180, 220, 250}
\definecolor{mycolor_box}{RGB}{50, 50, 50}
\definecolor{mycolor_tab-1}{RGB}{255, 255, 255}
\definecolor{mycolor_tab-2}{RGB}{234, 248, 253}
\definecolor{darkcyan}{RGB}{0, 139, 139}

\definecolor{revision-color_1}{RGB}{165, 0, 0} 

\definecolor{pros}{RGB}{175, 60, 115}   %
\definecolor{cons}{RGB}{30, 140, 100}    %

\usepackage{fontawesome5}
\usepackage{academicons}

\usepackage[pagebackref=true,breaklinks=true,colorlinks,bookmarks=false,linkcolor=blue,citecolor=blue,urlcolor=blue]{hyperref}

\begin{document}

\title{Discrete Tokenization for Multimodal LLMs: A Comprehensive Survey}

\author{
        Jindong Li,
        Yali Fu,
        Jiahong Liu,
        Linxiao Cao, 
        Wei Ji,
        Menglin Yang,
        Irwin King,
        Ming-Hsuan Yang

\IEEEcompsocitemizethanks{
\IEEEcompsocthanksitem Jindong Li, Linxiao Cao, and Menglin Yang are with The Hong Kong University of Science and Technology (Guangzhou), Guangzhou, China. 
E-mail: jli839@connect.hkust-gz.edu.cn, lcao950@connect.hkust-gz.edu.cn, menglin.yang@outlook.com. 
\IEEEcompsocthanksitem Yali Fu is with Jilin University, Changchun, China. 
E-mail: fuyl23@mails.jlu.edu.cn. 
\IEEEcompsocthanksitem Jiahong Liu and Irwin King are with The Chinese University of Hong Kong, Hong Kong, China. 
E-mail: jiahong.liu21@gmail.com, king@cse.cuhk.edu.hk.
\IEEEcompsocthanksitem Wei Ji is with the Nanjing University, China. Email: weiji0523@gmail.com
\IEEEcompsocthanksitem Ming-Hsuan Yang is with the University of California at Merced, United States. Email: mhyang@ucmerced.edu
\IEEEcompsocthanksitem Jindong Li and Yali Fu contribute equally as co-first authors. Menglin Yang is the corresponding author.

}
\thanks{Manuscript received xxxx xx, 20xx; revised xxxx xx, 20xx.}}

%
%

\markboth{}%
{Shell \MakeLowercase{\textit{et al.}}: Bare Demo of IEEEtran.cls for Computer Society Journals}
%



\IEEEtitleabstractindextext{
\begin{abstract}
\justifying
The rapid advancement of large language models (LLMs) has intensified the need for effective mechanisms to transform continuous multimodal data into discrete representations suitable for language-based processing. Discrete tokenization, with vector quantization (VQ) as a central approach, offers both computational efficiency and compatibility with LLM architectures. Despite its growing importance, there is a lack of a comprehensive survey that systematically examines VQ techniques in the context of LLM-based systems. This work fills this gap by presenting the first structured taxonomy and analysis of discrete tokenization methods designed for LLMs. We categorize 8 representative VQ variants that span classical and modern paradigms and analyze their algorithmic principles, training dynamics, and integration challenges with LLM pipelines. Beyond algorithm-level investigation, we discuss existing research in terms of classical applications without LLMs, LLM-based single-modality systems, and LLM-based multimodal systems, highlighting how quantization strategies influence alignment, reasoning, and generation performance. In addition, we identify key challenges including codebook collapse, unstable gradient estimation, and modality-specific encoding constraints. Finally, we discuss emerging research directions such as dynamic and task-adaptive quantization, unified tokenization frameworks, and biologically inspired codebook learning. This survey bridges the gap between traditional vector quantization and modern LLM applications, serving as a foundational reference for the development of efficient and generalizable multimodal systems. A continuously updated version is available at: \url{https://github.com/jindongli-Ai/LLM-Discrete-Tokenization-Survey}.
\end{abstract}

\begin{IEEEkeywords}
Discrete Tokenization, Vector Quantization (VQ), Multimodal Learning, Large Language Models (LLMs).
\end{IEEEkeywords}}

\maketitle

\IEEEdisplaynontitleabstractindextext

%
\IEEEpeerreviewmaketitle

\IEEEraisesectionheading{
\section{Introduction}
\label{sec_Introduction}}


%
%
%
%

 

\IEEEPARstart{R}{e}cent advances in large language models (LLMs)~\cite{T5-and-variants_Huggingface, 2024_arXiv_LLaMA-3_The-LLama-3-Herd-of-Models, GPT-series_Huggingface, Mistral_Huggingface, InternVL_Huggingface, DeepSeek_Huggingface, 2025_arXiv_Qwen-3_Qwen3-Technical-Report} have significantly transformed the way machines understand and generate human language. These models have demonstrated exceptional capabilities in language comprehension and generation, driving their adoption across a wide range of applications. As research continues to evolve, there is growing interest in extending the capabilities of LLMs beyond text to encompass multimodal data, including images~\cite{2024_CVPR_V2L-Tokenizer_Beyond-Text=Frozen-Large-Language-Models-in-Visual-Signal-Comprehension, 2025_AAAI_MARS_MARS=Mixture-of-Auto-regressive-Models-for-Fine-grained-Text-to-image-Synthesis}, audio~\cite{2023_NeurlPS_TWIST_Textually-Pretrained-Speech-Language-Models, 2025_arXiv_Kimi-Audio_Kimi-Audio-Technical-Report}, and video~\cite{2024_ICML_Video-LaVIT_Video-LaVIT=Unified-Video-Language-Pre-training-with-Decoupled-Visual-motional-Tokenization, 2024_ICML_VideoPoet_VideoPoet=A-Large-Language-Model-for-Zero-shot-Video-Generation}, thus introducing new challenges in unifying heterogeneous modalities within a common framework.

\begin{figure}
    \centering
    \includegraphics[width=0.99\linewidth]{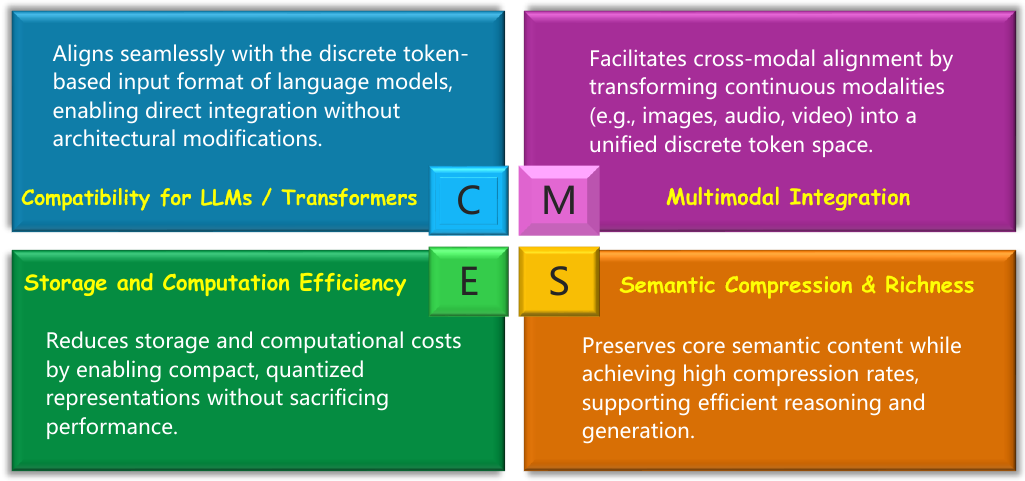}
    \caption{Discrete tokenization enables seamless integration with language models and supports efficient, scalable, and semantically meaningful processing for multimodal LLMs (MLLMs).}
    \label{fig:fig_1}
\end{figure}

Discrete tokenization based on vector quantization (VQ) has emerged as a key technique to address these challenges, offering significant advantages for multimodal integration in LLMs~\cite{2024_Preprints_Survey_Continuous-or-Discrete-That-Is-the-Question=A-Survey-on-Large-Multi-modal-Models-from-the-Perspective-of-Input-output-Space-Extension, 2024_arXiv_Survey_Next-Token-Prediction-Towards-Multimodal-Intelligence=A-Comprehensive-Survey}. As illustrated in Fig.~\ref{fig:fig_1}, by transforming high-dimensional continuous inputs into compact discrete tokens, it enables non-text modalities to be processed in a format aligned with the inherently token-based structure of language models. This design not only improves computational efficiency through compression, but also retains semantic granularity essential for cross-modal reasoning. Due to these strengths, discrete tokenization has become a core component in many state-of-the-art multimodal LLMs (MLLMs).

\begin{figure*}[t!]
	\centering
	\resizebox{0.95\linewidth}{!}{
    	\input{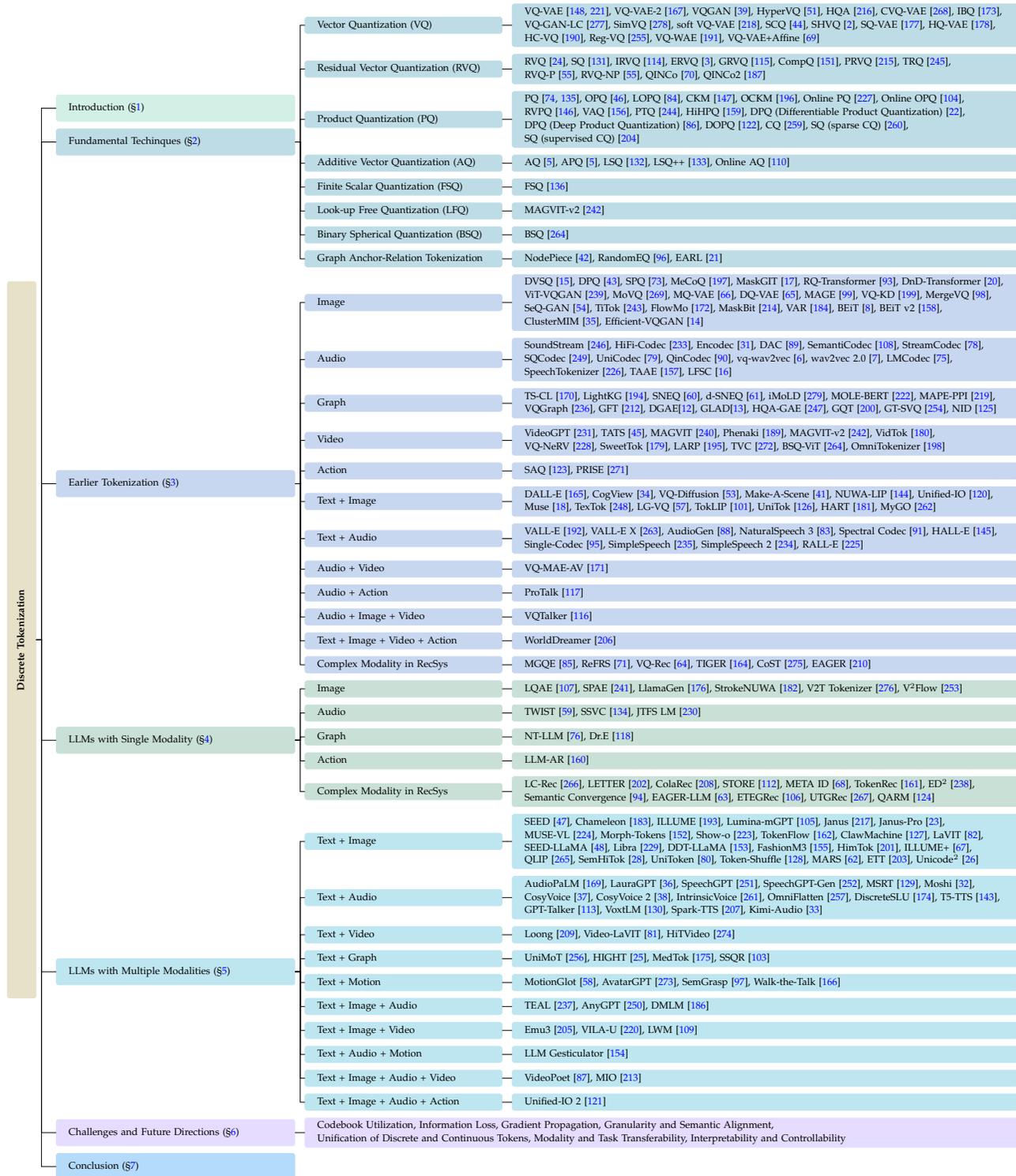}
    	}
	\caption{Taxonomy of this survey with representative works. Specifically, it is organized from the perspective of modality.}
	\label{fig:fig_tree}
\end{figure*}

Despite the growing relevance of discrete tokenization, existing surveys remain limited in both scope and technical depth. Several earlier reviews~\cite{1988_TOC_Survey_Image-Coding-Using-Vector-Quantization=A-Review, 1996_TIP_Survey_Advances-in-Residual-Vector-Quantization=A-Review, 2006_IEEE-Potentials_Survey_A-Review-of-Vector-Quantization-Techniques, 2019_FITEE_Survey_Vector-Quantization=A-Review, 2020_AIIHMSP_Survey_A-Survey-of-Data-Hiding-based-on-Vector-Quantization} cover topics before the emergence of LLMs and are no longer adequate in the context of today’s rapidly evolving AI landscape. While recent works have offered broader overviews of multimodal learning systems~\cite{2025_arXiv_Survey_From-Principles-to-Applications=A-Comprehensive-Survey-of-Discrete-Tokenizers-in-Generation-Comprehension-Recommendation-and-Information-Retrieval}, the treatment of quantization techniques remains insufficient.
Other surveys are narrowly scoped and restricted to individual modalities or tasks. For instance,~\citet{2025_arXiv_Survey_A-Survey-of-Quantized-Graph-Representation-Learning=Connecting-Graph-Structures-with-Large-Language-Models} provides a comprehensive account of quantization methods for graph-structured data, yet does not generalize beyond this domain. Similarly,~\cite{2024_arXiv_Survey_Vector-Quantization-for-Recommender-Systems=a-Review-and-Outlook} centers exclusively on recommender systems, emphasizing efficiency and representation quality, while~\cite{2025_arXiv_Survey_Recent-Advances-in-Discrete-Speech-Tokens=A-Review} focuses solely on discrete speech tokens for representation learning. This fragmentation and lack of cross-modal integration pose challenges for researchers aiming to design general-purpose, LLM-based multimodal systems.

In this work, VQ-based discrete tokenization is systematically discussed to better understand its role in addressing the limitations of current multimodal LLM systems. The analysis connects tokenization design choices to key integration requirements of LLMs, such as maintaining token alignment and ensuring effective gradient propagation through quantized representations. By analyzing applications across all major modalities within a unified analytical framework, this survey offers comparative insights that have been lacking in prior literature. Furthermore, it identifies and clarifies key challenges in current implementations, providing practical insights for enhancing quantization quality and system robustness. The overall structure of our survey is shown in Fig.~\ref{fig:fig_tree}.
Our main contributions are summarized as follows:
\begin{itemize}
    \item We establish a comprehensive taxonomy that organizes existing discrete tokenization methods based on their codebook learning paradigms and compatibility with LLM integration requirements.
    \item Representative applications in non-LLM settings are reviewed to reveal how their design principles can inform the construction of modality-specific tokenization strategies suitable for LLMs.
    \item A detailed modality-wise analysis is provided that compares discrete tokenization approaches across various data types within LLM systems.
    \item Key challenges in current techniques are identified and future research directions are outlined, including strategies to mitigate codebook collapse and to enable dynamic and adaptive quantization.
\end{itemize}

\section{Fundamental Techniques}
\label{Sec_Fundamental-Techniques}

\begin{figure}[!t]
    \centering
    \includegraphics[width=0.9\linewidth]{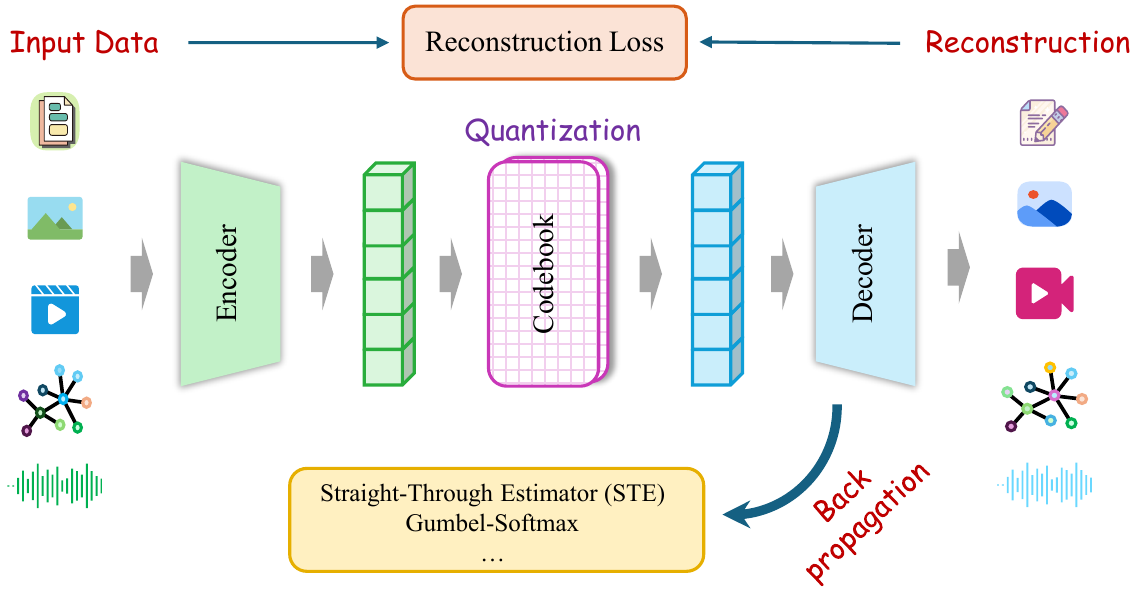}
\caption{General pipeline of discrete quantization based on VAE~\cite{2017_NeurlPS_VQ-VAE_Neural-discrete-representation-learning}, involving three main stages: encoding, quantization, and decoding.}
    \label{fig:VQ-pipeline}
\end{figure}

In the context of LLMs, discrete tokenization (quantization in non-LLM models) serves as the fundamental unit of representation, enabling efficient processing and generation of complex data across modalities. 
Tokens are derived through quantization techniques, which map continuous or high-dimensional data to a discrete, finite set of representations known as a codebook.
A typical formulation of discrete quantization follows the pipeline shown in Fig.~\ref{fig:VQ-pipeline}.

\noindent\textbf{General Formulation.}
\textit{
The discrete quantization pipeline begins with input data $\mathbf{x}$ (e.g., image, audio), which is processed by an encoder into a continuous latent representation $\mathbf{z}$. This continuous representation $\mathbf{z}$ is then discretized to a specific representation $\mathbf{c}_q$ in the codebook through a quantization process $\mathcal{Q}$. Finally, the discrete representation $\mathbf{c}_q$ is passed to a decoder, which outputs $\hat{\mathbf{x}}$ to approximate $\mathbf{x}$ as much as possible.
}

The process involves transforming continuous data into discrete tokens, which are encouraged to retain sufficient information and then used for downstream tasks such as generation or classification. The encoder and decoder typically consist of a deep neural network (e.g., convolutional or transformer-based layers)~\cite{2022_ICLR_ViT-VQGAN_Vector-quantized-Image-Modeling-with-Improved-VQGAN, 2024_ICLR_MAGVIT-v2_Language-Model-Beats-Diffusion-Tokenizer-is-Key-to-Visual-Generation, 2024_ICLR_SpeechTokenizer_SpeechTokenizer=Unified-Speech-Tokenizer-for-Speech-Language-Models, 2025_AAAI_Dr.E_Multi-View-Empowered-Structural-Graph-Wordification-for-Language-Models}, depending on the data modality.

To effectively implement the discrete quantization pipeline, three critical questions need to be addressed: \textbf{Q1:} How to train the entire pipeline? \textbf{Q2:} How to flow the gradient through discrete bottleneck? \textbf{Q3:} How to implement the quantization process $\mathcal{Q}$? These questions are key to enabling efficient, end-to-end training and implementation of discrete tokenization.

\subsubsection*{\textbf{Q1: How to Train the Entire Pipeline?}}

There are three primary methods for training the discrete quantization process: reconstruction-based, adversarial-based, and contrastive-based methods.

\noindent \textbf{Reconstruction-based Methods.}
This paradigm usually refers to a variational autoencoder (VAE)-based framework, which learns discrete representations by optimizing the reconstruction quality of the original input data.

The classical and fundamental models, VQ-VAE~\cite{2017_NeurlPS_VQ-VAE_Neural-discrete-representation-learning} and hierarchical VQ-VAE (i.e., VQ-VAE-2~\cite{2019_NeurlPS_vq_vae_2_Generating-diverse-high-fidelity-images-with-vq-vae-2}) jointly optimize the whole quantization pipeline by minimizing a combined loss:
\begin{equation}
    \mathcal{L}_{\mathrm{vq-vae}} = \|\mathbf{x} - \hat{\mathbf{x}}\|_2^2 + \|\text{sg}[\mathbf{z}] - \mathbf{c}_{q}\|_2^2 + \beta \|\mathbf{z} - \text{sg}[\mathbf{c}_{q}]\|_2^2, 
\label{Eq:VQ-VAE_loss}
\end{equation}
\begin{equation}
    \text{sg}(x) = 
        \begin{cases} 
        x & \text{forward pass, identity function} \\
        0 & \text{backward pass, gradient is stopped}
        \end{cases} ,
\end{equation}
where the three terms correspond to reconstruction loss $\mathcal{L}_{\mathrm{reconstruction}}$, codebook loss $\mathcal{L}_{\mathrm{codebook}}$, and commitment loss $\mathcal{L}_{\mathrm{commitment}}$ (scaled by weight $\beta$), and $\text{sg}(\cdot)$ denotes the stop-gradient operator.

\noindent \textbf{Adversarial-based Methods.}
VQGAN~\cite{2021_CVPR_VQGAN_Taming-transformers-for-high-resolution-image-synthesis} extends the standard VQ-VAE by introducing adversarial training and a perceptual loss for learning a perceptually rich codebook.

The network is optimized by combining the VQ-VAE loss $\mathcal{L}_{\text{vq-vae}}$ in Eq.~(\ref{Eq:VQ-VAE_loss}) with an adversarial loss $\mathcal{L}_{\text{gan}}$:
\begin{equation}
    \mathcal{L}_{\text{vqgan}} = \mathcal{L}_{\text{vq-vae}} + \lambda \, \mathcal{L}_{\text{gan}}, \quad
    \lambda = \frac{\nabla_{\mathcal{D}_L} [\mathcal{L}_{\text{per}}]}{\nabla_{\mathcal{D}_L} [\mathcal{L}_{\text{gan}]} + \delta},
\end{equation}
\begin{equation}
\mathcal{L}_{\text{gan}} = \log \mathbb{D}(\mathbf{x}) + \log(1 - \mathbb{D}(\hat{\mathbf{x}})),
\end{equation}
where $\mathbb{D}$ is the patch-based discriminator, $\lambda$ is the weighting coefficient, $\mathcal{D}$ denotes the decoder, $\mathcal{L}_{\text{per}}$ is the perceptual loss, $\nabla_{\mathcal{D}_L}[\cdot]$ denotes the gradient of its input with respect to the last layer $L$ of the decoder, and $\delta$ is a small constant for numerical stability.

\subsubsection*{\textbf{Q2: How to Flow Gradient Through Discrete Bottleneck?}}

The argmax operation in quantization is non-differentiable (detailed in Section~\ref{SubsubSec:Code-Assignment}), which blocks the gradient flow during training. To address this, various strategies have been proposed.

\noindent \textbf{Straight-Through Estimator (STE).}
STE~\cite{2013_arXiv_STE} offers a heuristic method that enables gradient flow through a non-differentiable discrete bottleneck. It treats the quantization as an identity function during the backward pass, and directly copies the gradients from the decoder input to the encoder output. This leads to the following approximation:
\begin{equation}
    \nabla_{\mathbf{z}} \mathcal{L} \approx \nabla_{\mathbf{c}_q} \mathcal{L},
\end{equation}

\noindent \textbf{Gumbel-Softmax.}
The Gumbel-Softmax~\cite{2016_arXiv_Gumbel-Softmax,2019_arXiv_2020_ICLR_vq-wav2vec_vq-wav2vec-Self-supervised-learning-of-discrete-speech-representations} provides a differentiable approximation to categorical sampling by replacing non-differentiable discrete sampling of quantization with a differentiable continuous relaxation perturbed by Gumbel noise during training.

Specifically, given a categorical distribution with class probabilities $\boldsymbol{\pi} = (\pi_1, \dots, \pi_K)$, the discrete one-hot sample can be approximated by a differentiable softmax function:
\begin{equation}
    \mathbf{y} = \text{softmax}\left( \frac{\log \boldsymbol{\pi} + \mathbf{g}}{\tau} \right),
\end{equation}
where $\mathbf{g} = (g_1, \dots, g_K)$ is a vector of i.i.d. samples drawn from the $Gumbel(0, 1)$ distribution:
\begin{equation}
g_k = -\log\left(-\log(u_k)\right), \quad u_k \sim \text{Uniform}(0,1),
\end{equation}
and $\tau > 0$ is the temperature parameter that controls the smoothness of the probability distribution.

During testing, as $\tau \to 0$, the distribution is closer to a one-hot vector by the non-differentiable argmax operation:
\begin{equation}
\lim_{\tau \to 0} \mathbf{y} = \text{one-hot} \left( \arg\max_k (\log \pi_k + g_k) \right),
\end{equation}
where the exact categorical sample is recovered via the Gumbel-Max trick.

This relaxation allows gradients to flow through the discrete sampling process, enabling gradient-based optimization. During training, $\tau$ is typically annealed from a high value (for smoother distributions) to a low value (for near one-hot outputs), bridging the gap between continuous and discrete representations.

\noindent \textbf{Rotation Trick.}
\citet{2025_ICLR_rotation-trick_Restructuring-vector-quantization-with-the-rotation-trick} proposes a rotation trick for gradient propagation, aligning encoder outputs to their nearest codebook vectors via rotation, rescaling linear transformation and encoding relative magnitude and angle between encoder output and codebook vector in the gradient.

\begin{figure}[!t]
    \centering
    \includegraphics[width=0.31\linewidth]{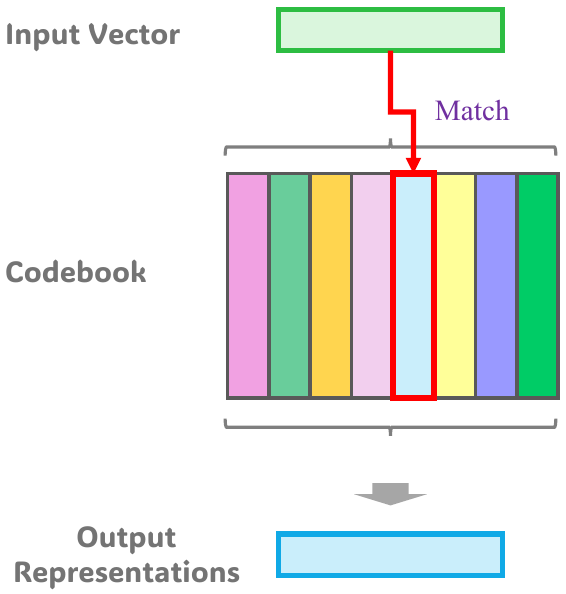}
    \hspace{1.5em}
    \includegraphics[width=0.32\linewidth]{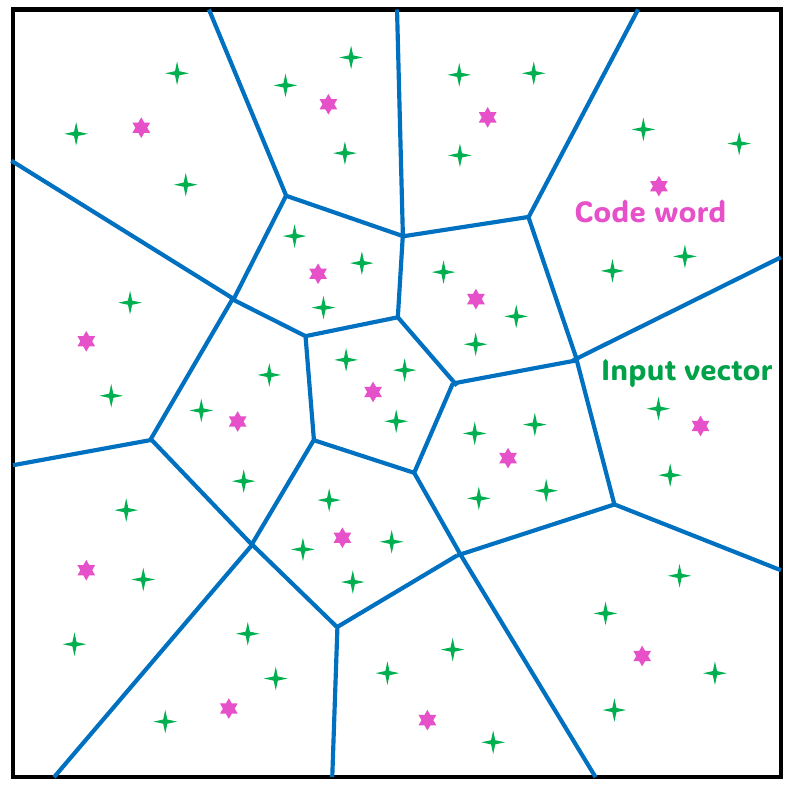}
    \caption{Illustration of vector quantization (VQ): each input vector is matched to its nearest codeword in a finite codebook (left), corresponding to the Voronoi partitions of continuous space (right).}
    \label{fig:VQ}
\end{figure}

\subsubsection*{\textbf{Q3: How to Implement the Quantization Process $\mathcal{Q}$?}}

There are eight primary methods (Section~\ref{subsec:Vanilla-Vector-Quantization(VQ)}-Section~\ref{subsec:Graph-Anchor-relation-Tokenization}) for implementing the quantization process $\mathcal{Q}$, each offering unique approaches to discretizing continuous data.
The following subsections systematically review fundamental quantization methods from classical algorithms to modern innovations by highlighting their unique mechanisms.

\begin{figure}[t]
    \centering
    \includegraphics[width=0.99\linewidth]{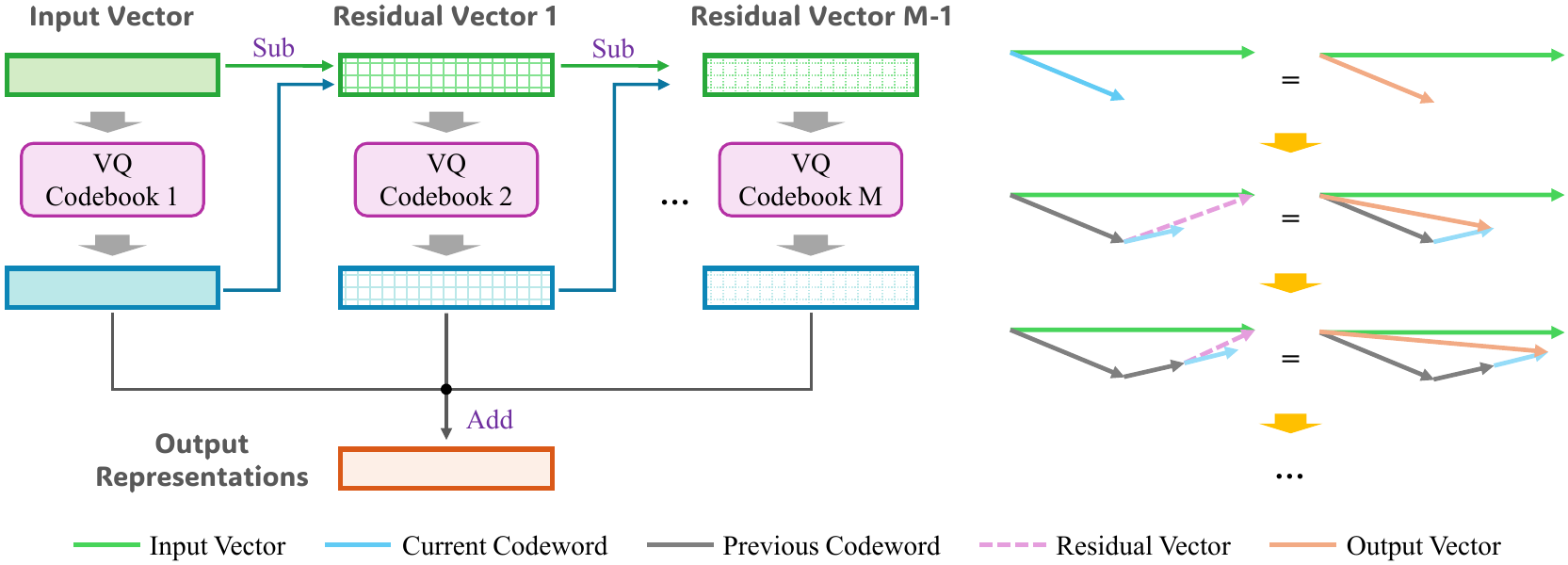}
    \caption{Illustration of RVQ with multi-stage quantization. Each stage quantizes the residual vector from the previous stage (left), progressively approximating the input vector as shown in the geometric visualization of vector operations (right). }
    \label{fig:RVQ}
\end{figure}

\subsection{Vector Quantization}
\label{subsec:Vanilla-Vector-Quantization(VQ)}

In LLMs, Vector Quantization (VQ)~\cite{2019_FITEE_Survey_Vector-Quantization=A-Review, 2017_NeurlPS_VQ-VAE_Neural-discrete-representation-learning, 2021_CVPR_VQGAN_Taming-transformers-for-high-resolution-image-synthesis} is a technique that discretizes continuous latent representations by mapping them to the closest entries in a finite codebook, as illustrated in Fig.~\ref{fig:VQ}. 
It plays a key role that bridges between continuous and discrete representations, enabling compact and interpretable modeling.

\noindent \textbf{Definition [Vector Quantization].}
\textit{
Let $\mathcal{Z} \subseteq \mathbb{R}^D$ be the continuous input space, $\mathbf{z} \in \mathcal{Z}$ be a $D$-dimensional input vector, and $\mathcal{C} = \{\mathbf{c}_1, \mathbf{c}_2, \dots, \mathbf{c}_K\} \subseteq \mathbb{R}^D$ be a codebook with $K$ \textbf{codewords (also called codevectors or codes)}. Vector Quantization defines a mapping function $q: \mathcal{Z} \to \mathcal{C}$, which assigns a continuous vector $\mathbf{z}$ to its nearest codeword $\mathbf{c}_{k^*}$, i.e., $q(\mathbf{z}) = \mathbf{c}_{k^*}$. 
}

\subsubsection{Codebook Initialization}
Each codevector $\mathbf{c}_k$ in the codebook is usually a prototype vector. For the initialization of $K$ codevectors $\{\mathbf{c}_k\}_{k=1}^K$, a common practice is to sample from a Gaussian distribution $\mathcal{N}(0, I)$ or use uniform initialization in a small range (e.g., $\mathcal{U}(-0.1, 0.1)$)~\cite{2024_arXiv_SimVQ_Addressing-representation-collapse-in-vector-quantized-models-with-one-linear-layer, 2022_ICLR_ViT-VQGAN_Vector-quantized-Image-Modeling-with-Improved-VQGAN}. 
In addition, $K$-means clustering method can be applied to find the cluster centroids of the training embeddings~\cite{2017_NeurlPS_VQ-VAE_Neural-discrete-representation-learning, 2025_ICLR_GQT_Learning-Graph-Quantized-Tokenizers}. 
HyperVQ~\cite{2024_arXiv_HyperVQ_Hypervq-Mlr-based-vector-quantization-in-hyperbolic-space} defines geometrically constrained codevectors by performing hyperbolic multinomial logistic regression and selecting a representative point in the decision hyperplane.

\subsubsection{Code Assignment: Embedding to Code Mapping}
\label{SubsubSec:Code-Assignment}

As illustrated in Fig.~\ref{fig:VQ}, given the continuous latent embedding $\mathbf{z}$, vector quantization assigns it to the nearest code in the codebook by argmax operation:
\begin{equation}
    k^\star = \arg\min_{k} \|\mathbf{z} - \mathbf{c}_k\|_2,
\label{Eq:Arg}
\end{equation}
and the quantized output is:
\begin{equation}
    \mathbf{c}_q = \mathbf{c}_{k^\star}.
\end{equation}

The above argmax assignment is typically referred to as \textit{deterministic quantization}, where identical input is always assigned to the same codeword. Additionally, some methods~\cite{2023_CVPR_Reg-VQ_Regularized-vector-quantization-for-tokenized-image-synthesis, 2024_arXiv_LARP_LARP=Tokenizing-Videos-with-A-Learned-Autoregressive-Generative-Prior, 2021_ICML_DALL-E_Zero-shot-Text-to-image-Generation, 2019_arXiv_2020_ICLR_vq-wav2vec_vq-wav2vec-Self-supervised-learning-of-discrete-speech-representations} employ the Gumbel-Softmax operation to introduce stochasticity or noise during training, named \textit{stochastic quantization}, where it assigns codewords based on probability distribution and can assign different codewords for identical input, helping to escape local optima.

\begin{figure*}[t]
    \centering
    \includegraphics[width=0.99\linewidth]{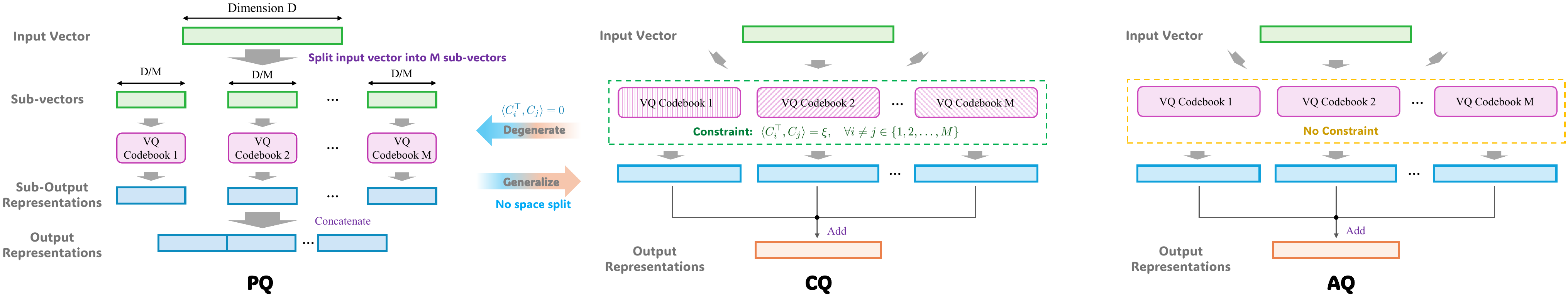}
    \caption{Illustration of PQ, CQ, and AQ. Each sub-vector of the input vector is quantized independently in its own subspace (left). Both CQ and AQ quantize the input vector via multiple full-dimensional codebooks without dimension split while CQ particularly introduces a constant product constraint between codebooks (middle and right). When the constant equals to 0, CQ degenerates to PQ.}
    \label{fig:PQ-CQ-AQ}
\end{figure*}

\subsubsection{Codebook Updating}

Updating the codebook during training is critical to ensure it remains representative and stable. In addition to codebook loss, one commonly adopted approach is EMA (exponential moving average) updating.

\noindent \textbf{Codebook Loss.}
The codebook can be updated by codebook loss, second term in Eq.~(\ref{Eq:VQ-VAE_loss}), which pulls codewords toward the encoder outputs. This loss encourages the codebook to better cover the distribution of encoded features and improves quantization quality.

\noindent \textbf{Exponential Moving Average (EMA) Updating.}
EMA strategy updates the codebook by progressively reflecting the distribution of encoder outputs through running averages that track both the assignment counts and the cumulative encoder outputs for each codevector $\mathbf{c}_i$~\cite{2017_NeurlPS_VQ-VAE_Neural-discrete-representation-learning,2019_NeurlPS_vq_vae_2_Generating-diverse-high-fidelity-images-with-vq-vae-2,2018_arXiv_VQ-VAE-with-soft-EM_Theory-and-experiments-on-vector-quantized-autoencoders}.

For training step $t$, the following statistics are updated:
\begin{equation}
N_i^{(t)} := \gamma N_i^{(t-1)} + (1 - \gamma) n_i^{(t)},
\end{equation}
\begin{equation}
m_i^{(t)} := \gamma m_i^{(t-1)} + (1 - \gamma) \sum_{j=1}^{n_i^{(t)}} E(x)_{i,j}^{(t)},
\end{equation}
where $E(x)$ denotes the encoder output vectors in the current mini-batch, $n_i^{(t)}$ is the number of vectors in $E(x)$ which will be quantized to the codevector $\mathbf{c}_i$, and $\gamma$ is a decay parameter (typically 0.99).

After accumulating these statistics, the codevector $\mathbf{c}_i$ is updated as:
\begin{equation}
\mathbf{c}_i^{(t)} := \frac{m_i^{(t)}}{N_i^{(t)}}.
\end{equation}
This EMA approach ensures that the codebook evolves smoothly during training by integrating information across multiple mini-batches, leading to a more stable and representative set of codevectors.
Furthermore, \citet{2018_arXiv_VQ-VAE-with-soft-EM_Theory-and-experiments-on-vector-quantized-autoencoders} introduces a soft Expectation Maximization (EM) algorithm, which assigns the input embedding to a probabilistic distribution over codevectors and updates the involved codevectors, instead of updating only the nearest codevector.

The above vanilla vector quantization serves as the basis for a wide range of quantization techniques, where it maps inputs to the nearest codeword to achieve compact and discrete representations. 
We also discuss the prevalent codebook collapse issue in vector quantization at the end of this section, especially representative solutions under the vanilla VQ mechanism, such as HQA~\cite{2020_NeurlPS_HQA_Hierarchical-quantized-autoencoders}, CVQ-VAE~\cite{2013_CVQ_VAE_Online-clustered-codebook}, VQ-WAE~\cite{2023_ICML_VQ-WAE_Vector-Quantized-Wasserstein-Auto-Encoder}, SQ-VAE~\cite{2022_arXiv_SQ_VAE_Sq-vae-Variational-bayes-on-discrete-representation-with-self-annealed-stochastic-quantization}, and HQ-VAE~\cite{2024_TMLR_HQ-VAE_Hq-vae=Hierarchical-discrete-representation-learning-with-variational-bayes}.

\subsection{Residual Vector Quantization}
\label{subsec:Residual-Vector-Quantization(RVQ)}

Residual Vector Quantization (RVQ)~\cite{1996_TIP_Survey_Advances-in-Residual-Vector-Quantization=A-Review,2010_MDPI_RVQ_Approximate-nearest-neighbor-search-by-residual-vector-quantization} introduces a multi-stage quantization mechanism to gradually reduce the quantization error. As depicted in Fig.~\ref{fig:RVQ}, instead of mapping the input vector to a single codeword, RVQ encodes the input through a sequence of residual quantization stages, where each stage encodes the quantization residual from the previous stage.

\noindent\textbf{Definition [Residual Vector Quantization].}
\textit{
Let $\mathbf{z} \in \mathbb{R}^D$ be a $D$-dimensional input vector, and $\{\mathcal{C}^{(i)}\}_{i=1}^M$ be a set of $M$ codebooks, where each codebook $\mathcal{C}^{(i)} = \{\mathbf{c}^{(i)}_1, \dots, \mathbf{c}^{(i)}_{K_i}\} \subseteq \mathbb{R}^D$ contains $K_i$ codewords. Residual Vector Quantization defines $M$ sequential quantization stages. For the stage $(i+1)$, RVQ quantizes the residual vector $\mathbf{r}^{(i+1)} = \mathbf{r}^{(i)} - \mathbf{c}^{(i)}_{k^*}$ to its nearest codeword $\mathbf{c}^{(i+1)}_{k^*}$ in the $(i+1)$-th codebook $\mathcal{C}^{(i+1)}$. In particular, the first residual $\mathbf{r}^{(1)} = \mathbf{z}$. The final quantized output $\mathbf{z}_q$ is obtained by summing the selected codewords: $\mathbf{z}_q = \sum_{i=1}^M \mathbf{c}^{(i)}_{k^*}$.
}

\subsubsection{Codebook Structure and Optimization}

The effectiveness of RVQ heavily depends on the structure and optimization of its stage-wise codebooks.
SQ~\cite{2014_arXiv_SQ_Stacked-quantizers-for-compositional-vector-compression} presents a hierarchical dependency structure among subcodebooks, utilizing greedy coarse-to-fine encoding, and employing hierarchical k-means and top-down refinement to initialize and update subcodebooks, thereby reducing quantization error.
In~\cite{2015_arXiv_IRVQ_Improved-residual-vector-quantization-for-high-dimensional-approximate-nearest-neighbor-search}, IRVQ combines subspace clustering with warm-started k-means to learn high-entropy codebooks and introduces a multi-path encoding strategy that mitigates greedy encoding errors.
On the other hand, 
ERVQ~\cite{2017_Multimedia-Systems_ERVQ_Optimized-residual-vector-quantization-for-efficient-approximate-nearest-neighbor-search} proposes a joint optimization to iteratively optimize all stage codebooks by the others, instead of training them sequentially. 
\citet{2017_IEEE-Multimedia_GRVQ_Generalized-residual-vector-quantization-and-aggregating-tree-for-large-scale-search} propose the generalized RVQ (GRVQ) by introducing transition clustering to improve k-means and multipath encoding for lower quantization error. 
CompQ~\cite{2016_TKDE_CompQ_Competitive-quantization-for-approximate-nearest-neighbor-search} introduces a competitive quantization to jointly train all codebooks via redefining "winner codevector" and stochastic gradient descent.

\subsubsection{Projected or Transformed Residual Quantization}

RVQ can be enhanced by applying projections or transformations to the residual vectors, improving alignment and quantization accuracy across stages.
PRVQ~\cite{2014_IEEE-multimedia_PRVQ_Projected-residual-vector-quantization-for-ANN-search} enhances RVQ by incorporating PCA projections with dimensionality reduction before residual quantization, ensuring that the discarded projection information is retained and used.
Transformed Residual Quantization (TRQ)~\cite{2015_arXiv_TRQ_Transformed-residual-quantization-for-approximate-nearest-neighbor-search} introduces cluster-wise transforms in RVQ by learning a local rotation matrix for each residual cluster and aligns residual vectors via the proposed iterative alignment (IA) to reduce quantization noise and improve subsequent quantization accuracy. 
\citet{2016_Neurocomputing_RVQ-P-RVQ-NP_Parametric-and-nonparametric-residual-vector-quantization-optimizations-for-ANN-search} optimize RVQ by projection of data with an orthogonal matrix, proposing the non-parametric RVQ-NP and the parametric RVQ-P.

\subsubsection{Implicit and Neural Codebook Generation}

Implicit and neural methods construct RVQ codebooks in a data-adaptive manner during quantization.
QINCo~\cite{2024_arXiv_QINCo_Residual-quantization-with-implicit-neural-codebooks} replaces the fixed codebooks in RVQ with neural networks that generate step-specific codebooks conditioned on partial reconstructions, allowing the codebooks to adapt to residual distributions.
QINCo2~\cite{2025_arXiv_QinCo2_Qinco2-Vector-Compression-and-Search-with-Improved-Implicit-Neural-Codebooks} further improves QINCo by introducing the codeword pre-selection with beam search for improved vector encoding, a lookup-based decoder for efficient large-scale search, and an optimized training procedure and network architecture.

\subsection{Product Quantization}
\label{subsec:Product-Quantization(PQ)}

Product Quantization (PQ)~\cite{2018_TMTA_Survey_A-survey-of-product-quantization,2011_TPAMI_PQ_Product-Quantization-for-Nearest-Neighbor-Search} decomposes the original vector space into multiple lower-dimensional subspaces and quantizes each subspace independently, as illustrated in Fig.~\ref{fig:PQ-CQ-AQ}. This approach drastically reduces quantization error while maintaining compact representations, and is particularly effective in high-dimensional scenarios.

\noindent\textbf{Definition [Product Quantization].}  
\textit{  
Let $\mathcal{Z} \subseteq \mathbb{R}^D$ be the input space, and let $\mathbf{z} \in \mathcal{Z}$ be a $D$-dimensional input vector. Product Quantization partitions $\mathbf{z}$ into $M$ disjoint sub-vectors: $\mathbf{z} = [\mathbf{z}^{(1)}, \mathbf{z}^{(2)}, \dots, \mathbf{z}^{(M)}]$, where each $\mathbf{z}^{(m)} \in \mathbb{R}^{D/M}$. For each subspace, a separate sub-codebook $\mathcal{C}^{(m)} = \{\mathbf{c}_1^{(m)}, \dots, \mathbf{c}_{K}^{(m)}\} \subseteq \mathbb{R}^{D/M}$ is trained. The overall codebook $\mathcal{C}$ is then defined as the Cartesian product $\mathcal{C} = \mathcal{C}^{(1)} \times \mathcal{C}^{(2)} \times \dots \times \mathcal{C}^{(M)}$, resulting in $K^M$ possible composite codewords. The sub-vector $\mathbf{z}^{(m)}$ is quantized independently by mapping it to its nearest sub-codeword $\mathbf{c}_{k^*}^{(m)}$ in m-th subspace. The final quantized output $\mathbf{z}_q$ is given by concatenating those codewords:
\begin{equation}
    \mathbf{z}_q = [\mathbf{c}_{k^*}^{(1)}, \mathbf{c}_{k^*}^{(2)}, \dots, \mathbf{c}_{k^*}^{(M)}].
\end{equation}
}

\subsubsection{Space Decomposition Optimization}

Optimizing subspace decomposition plays a central role in enhancing PQ performance by reducing quantization distortion.
Optimized Product Quantization (OPQ)~\cite{2013_CVPR_OPQ_Optimized-Product-Quantization-for-Approximate-Nearest-Neighbor-Search} considers the optimal space decomposition issue and transforms the input space by a rotation matrix $\mathbf{R}$, allowing optimal subspace partitioning. 
Locally Optimized Product Quantization (LOPQ)~\cite{2014_CVPR_LOPQ-Locally-Optimized-Product-Quantization-for-Approximate-Nearest-Neighbor-Search} employs a coarse quantizer to assign data to cells, then independently optimizes a rotation matrix and a product quantizer to encode residuals within each cell.
CKM~\cite{2013_CVPR_CKM_Cartesian-k-means} optimally rotates the original space, enabling lower distortion.
OCKM~\cite{2015_TKDE_OCKM_Optimized-cartesian-k-means} further optimizes CKM by introducing multiple sub-codebooks in each subspace and the multi-codeword selection in each sub-codebook.

\subsubsection{Codebook Structure and Update}

The design and maintenance of sub-codebooks are crucial for ensuring efficient and accurate quantization in PQ.
Online PQ~\cite{2018_TKDE_Online-PQ_Online-product-quantization} presents two budget constraints to update the partial codebooks incrementally.
In addition, Online OPQ~\cite{2020_ICDM_Online-OPQ_Online-optimized-product-quantization} extends to dynamically update the quantization codebooks and the rotation matrix via the Orthogonal Procrustes problem.
Residual Vector Product Quantization (RVPQ)~\cite{2023_ESWA_RVPQ_Residual-vector-product-quantization-for-approximate-nearest-neighbor-search} introduces residual codebooks within each subspace and optimizes them jointly, enhancing the quantization structure.
On the other hand, Variance-Aware Quantization (VAQ)~\cite{2022_ICDE_VAQ_Fast-adaptive-similarity-search-through-variance-aware-quantization} adapts codebook sizes to subspaces based on subspace importance via linear dimensionality reduction, and 
Product Tree Quantization (PTQ)~\cite{2015_ICIP_PTQ_Product-tree-quantization-for-approximate-nearest-neighbor-search} introduces the tree-structured codebooks and relaxes the subspace independence assumption of PQ, thereby reducing distortion.
Recently, HiHPQ~\cite{2024_AAAI_HiHPQ_HiHPQ=Hierarchical-Hyperbolic-Product-Quantization-for-Unsupervised-Image-Retrieval} proposes a hyperbolic product quantizer by a Cartesian product of hyperbolic subspaces and a soft hyperbolic codebook quantization based on Lorentzian distance.

\subsubsection{End-to-end Learning-Based PQ}

End-to-end learning-based PQ enables joint optimization of quantization and task-specific objectives through differentiable formulations.
Differentiable Product Quantization (DPQ)~\cite{2020_ICML_Differentiable-Product-Quantization-(DPQ)_Differentiable-product-quantization-for-end-to-end-embedding-compression} jointly learns discrete codes and task-specific objectives in an end-to-end differentiable manner via a differentiable softmax operation and a centroid-based approximation.
Deep Product Quantization (DPQ)~\cite{2019_CVPR_Deep-Product-Quantization-(DPQ)_End-to-end-supervised-product-quantization-for-image-search-and-retrieval} introduces a supervised end-to-end learnable PQ framework that leverages the supervised signal to learn soft and hard representations through a straight-through estimator jointly.
Differentiable Optimized Product Quantization (DOPQ)~\cite{2023_WWW_DOPQ_Differentiable-Optimized-Product-Quantization-and-Beyond} optimizes the non-differentiable argmax operation based on direct loss minimization for end-to-end training.

\begin{figure}[t]
    \centering
    \includegraphics[width=0.33\linewidth]{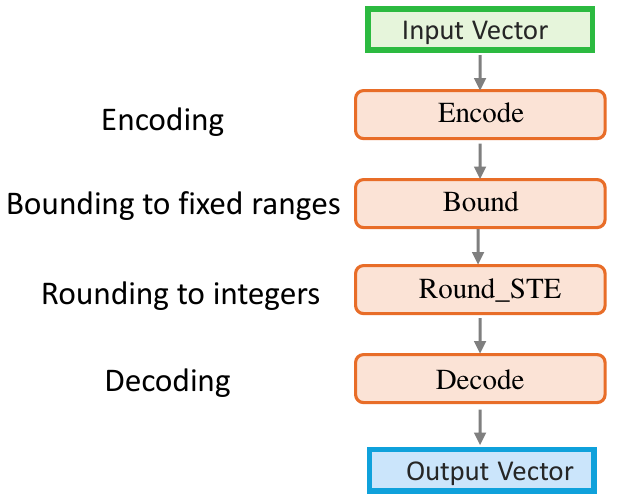}
    \hspace{3em}
    \includegraphics[width=0.27\linewidth]{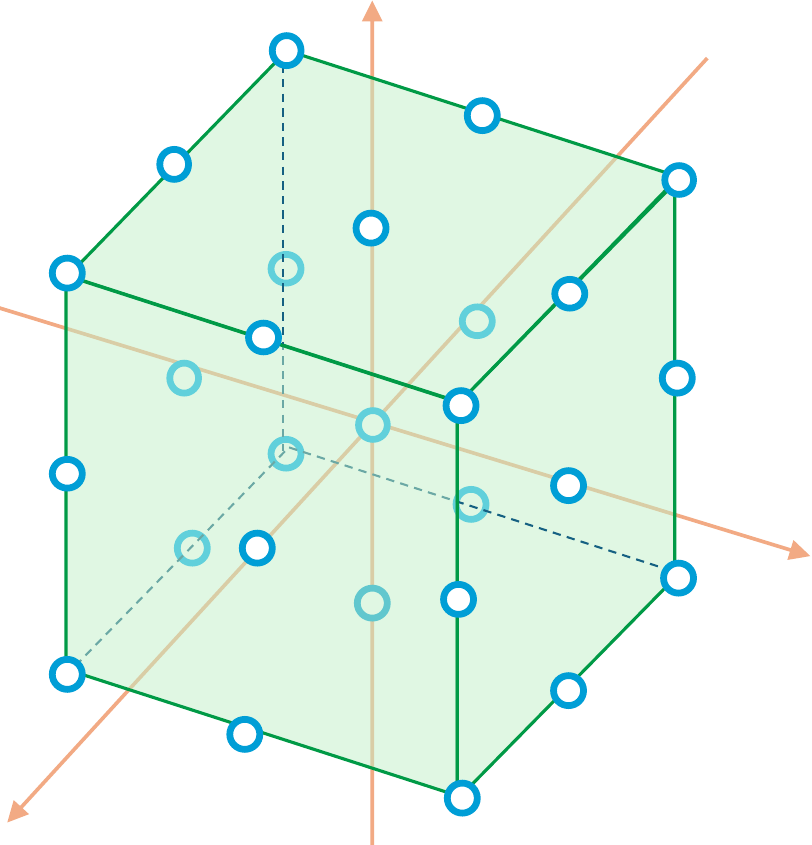}
    \caption{Illustration of FSQ. Each dimension of $D$-dimensional input vector is bounded and rounded to $L$ corresponding integers (left). The formed codebook has a size $L^D$, like the hypercube visualization of codeword distribution for $D=3$ and $L=3$ (right).}
    \label{fig:FSQ}
\end{figure}

\subsubsection{Generalized Product Quantization}

Composite Quantization (CQ)~\cite{2014_ICML_CQ_Composite-quantization-for-approximate-nearest-neighbor-search} can be viewed as a generalized formulation of PQ. Unlike PQ, CQ has no subspace decomposition and orthogonality constraint, and instead introduces a constant interdictionary-element-product constraint between codebooks (see Fig.~\ref{fig:PQ-CQ-AQ}). When the codebooks are constrained to be mutually orthogonal and codewords are zero-padded outside the designated subspace, CQ degenerates to PQ.

\noindent\textbf{Definition [Composite Quantization].}  
\textit{  
Let $\mathcal{Z} \subseteq \mathbb{R}^D$ be the input space and let $\mathbf{z} \in \mathcal{Z}$ be a $D$-dimensional input vector. Composite Quantization quantizes $\mathbf{z}$ in the original space as a summation of selected codewords $\mathbf{c}_{k^*}^{(m)}$ from $M$ global codebooks $\{\mathcal{C}^{(m)}\}_{m=1}^M$, where each codebook $\mathcal{C}^{(m)} = \{\mathbf{c}_1^{(m)}, \dots, \mathbf{c}_K^{(m)}\} \subseteq \mathbb{R}^D$ contains $K$ codewords.
In addition, codebooks satisfy a constant inter-dictionary-element-product constraint, i.e.,
\begin{equation}
    \langle C_i^\top, C_j \rangle = \xi,  \quad \forall i \ne j \in \{1, 2, \dots, M\},
\end{equation}
where $\langle \cdot, \cdot \rangle$ denotes the inner product, $\xi$ is a constant, and CQ degenerates to PQ when $\xi=0$. The quantized output $\mathbf{z}_q$ is $\mathbf{z}_q = \sum_{m=1}^{M} \mathbf{c}_{k^*}^{(m)}$.
}

\citet{2015_CVPR_Sparse-CQ(SQ)_Sparse-composite-quantization} introduce a sparse CQ (i.e., SQ) to construct sparse codebooks via the constant inter-dictionary-element-product constraint and sparsity regularization.
Supervised Quantization (SQ)~\cite{2016_CVPR_Supervised-CQ(SQ)_Supervised-quantization-for-similarity-search} is a supervised CQ method through quantization of the input in a linearly transformed discriminative subspace.

\begin{figure}[t]
    \centering
    \includegraphics[width=0.7\linewidth]{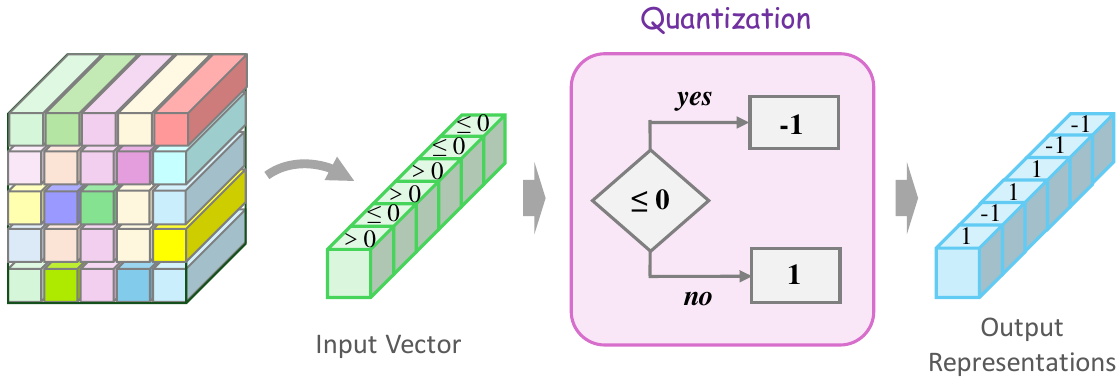}
    \caption{Illustration of LFQ. Each dimension of an input vector is directly quantized into 1 or -1.}
    \label{fig:LFQ}
\end{figure}

\subsection{Additive Vector Quantization}
\label{subsec:Additive-Vector-Quantization(AQ)}

Additive Vector Quantization (AQ)~\cite{2014_CVPR_AQ_Additive-quantization-for-extreme-vector-compression} quantizes the input as a sum of codewords selected from multiple full-dimensional codebooks, as illustrated in Fig.~\ref{fig:PQ-CQ-AQ}.

\noindent\textbf{Definition [Additive Vector Quantization].}
\textit{
Let $\mathcal{Z} \subseteq \mathbb{R}^D$ be the input space, and let $\mathbf{z} \in \mathcal{Z}$ be a $D$-dimensional input vector. Additive Vector Quantization quantizes $\mathbf{z}$ as a summation of $M$ codewords $\{\mathbf{c}^{(m)}_{k^*}\}_{m=1}^M$ selected from $M$ codebooks $\{\mathcal{C}^{(m)}\}_{m=1}^M$, where each codebook $\mathcal{C}^{(m)} = \{\mathbf{c}^{(m)}_1, \dots, \mathbf{c}^{(m)}_K\} \subseteq \mathbb{R}^D$. The quantized output $\mathbf{z}_q$ is $\mathbf{z}_q = \sum_{m=1}^{M} \mathbf{c}^{(m)}_{k^*}$.
}

\citet{2014_CVPR_AQ_Additive-quantization-for-extreme-vector-compression} introduce the additive product quantization (APQ) method that uses OPQ optimization to rotate the data and then applies AQ encoding to different parts of the rotated vector.
The local search quantization (LSQ)~\cite{2016_ECCV_LSQ_Revisiting-additive-quantization} enhances AQ by incorporating iterated local search (ILS) to efficiently handle the NP-hard encoding problem, and enforces the sparsity of the codebooks.
On the other hand, LSQ++~\cite{2018_ECCV_LSQ++_LSQ++-Lower-running-time-and-higher-recall-in-multi-codebook-quantization} improves LSQ by introducing a fast codebook update for a lower running time and stochastic relaxation techniques for greater recall.
In addition, Online AQ~\cite{2021_KDD_Online-AQ_Online-additive-quantization} dynamically updates codebooks for streaming data, and introduces a randomized block beam search algorithm to assign discrete codes to incoming data efficiently, with a better regret bound than online PQ.

\subsection{Finite Scalar Quantization}
\label{subsec:Finite-Scalar-Quantization(FSQ)}

Finite Scalar Quantization (FSQ)~\cite{2024_ICLR_FSQ_Finite-scalar-quantization-Vq-vae-made-simple} projects latent inputs to a few dimensions (typically less than 10) by the final encoder layer and quantizes each dimension independently to a small set of fixed scalar values by rounding to integers, using STE to propagate the gradient through the non-differentiable rounding operation, as shown in Fig.~\ref{fig:FSQ}. FSQ is a simple yet effective alternative to VQ in VQ-VAEs without any auxiliary losses and codebook collapse.

\noindent\textbf{Definition [Finite Scalar Quantization].}
\textit{
Given a $D$-dimensional input vector $\mathbf{z} = [z_1, z_2, \dots, z_D] \in \mathbb{R}^D$ (typically with $D < 10$), Finite Scalar Quantization quantizes each dimension $z_i$ into one of $L$ values $\{-\lfloor L/2 \rfloor, \dots, \lfloor L/2 \rfloor\}$. Specifically, for each dimension $z_i$, FSQ firstly applies a bounding function $f(\cdot)$ (e.g., $f(z_i) = \frac{L}{2} \cdot \tanh(z_i)$), and then rounds to integers, i.e.,
\begin{equation}
    q(z_i) = \operatorname{round}(f(z_i)) \in \{-\lfloor \frac{L}{2} \rfloor, \dots, \lfloor \frac{L}{2} \rfloor\}.
\end{equation}
The final quantized output $\hat{\mathbf{z}}$ is $\hat{\mathbf{z}} = [q(z_1), \dots, q(z_D)]$. For $\mathbf{z} \in \mathbb{R}^D$, there are $L^D$ possible codewords in the final implicit codebook. More generally, each dimension $z_i$ can be mapped to distinct $L_i$ values $\{-\lfloor L_i/2 \rfloor, \dots, \lfloor L_i/2 \rfloor\}$, finally forming an implicit codebook with size $\prod_{i=1}^{D} L_i$.
}

\begin{figure}[t]
    \centering
    \includegraphics[width=0.7\linewidth]{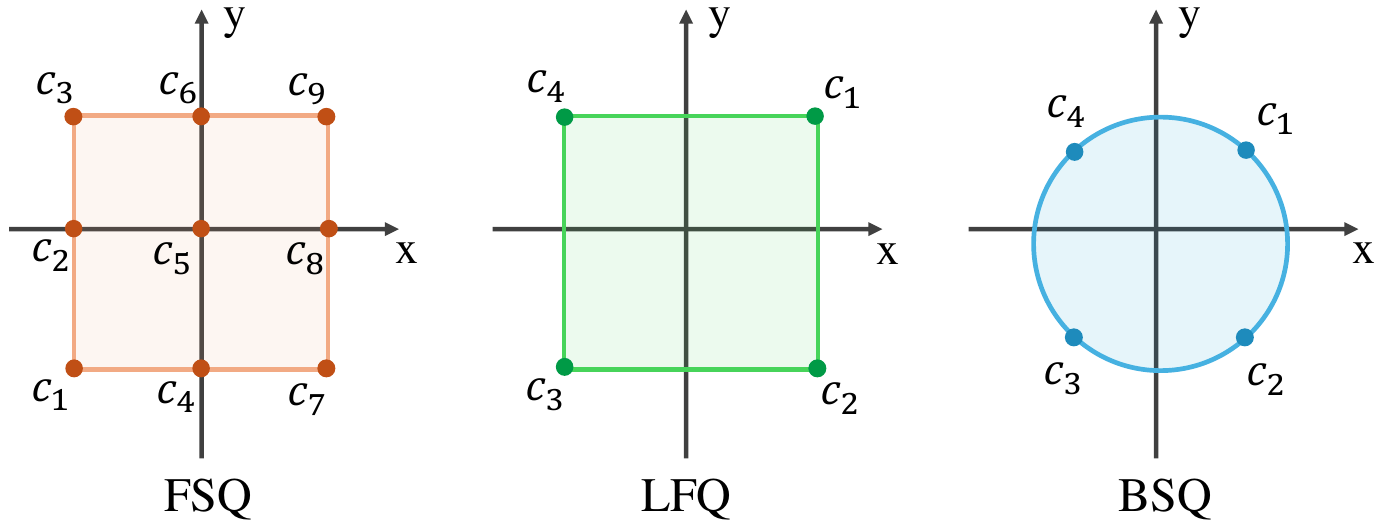}
    \caption{Comparison of FSQ, LFQ, and BSQ in 2D space. While the codewords of FSQ and LFQ partition the space into axis-aligned hypercubic cells, those of BSQ are uniformly distributed on the unit hypersphere.}
    \label{fig:BSQ}
\end{figure}

\subsection{Lookup-Free Quantization}
\label{subsec:Look-up-Free-Quantization(LFQ)}

Unlike the above VQ-based approaches, such as VQ in Section~\ref{subsec:Vanilla-Vector-Quantization(VQ)}, which need to look up all codewords in the codebook to find the nearest codeword for quantization, Lookup-Free Quantization (LFQ)~\cite{2024_ICLR_MAGVIT-v2_Language-Model-Beats-Diffusion-Tokenizer-is-Key-to-Visual-Generation} reduces the codebook’s embedding dimension to zero and maps the input to a binary integer set without lookup, as illustrated in Fig.~\ref{fig:LFQ}.

\noindent\textbf{Definition [Lookup-Free Quantization].}
\textit{
Given a D-dimensional input vector $\mathbf{z} = [z_1, z_2, \dots, z_D] \in \mathbb{R}^D$, Lookup-Free Quantization constructs an implicit codebook $\mathcal{C}$ as the Cartesian product of $D$ binary sets:
\begin{equation}
    \mathcal{C} = \times_{i=1}^{D} \mathcal{C}_i, \quad \text{where } \mathcal{C}_i = \{-1, +1\}, |\mathcal{C}| = 2^D.
\end{equation}
Each dimension $z_i$ is quantized independently into binary codebook $\mathcal{C}_i$ via the sign function $\text{sign}(\cdot)$:
\begin{equation}
    q(z_i) = \text{sign}(z_i) = -1 \cdot \mathbb{I}_{[z_i \le 0]} + 1 \cdot \mathbb{I}_{[z_i > 0]},
\end{equation}
where $\mathbb{I}_{[\cdot]}$ is the indicator function.
The quantized binary code $q(\mathbf{z}) \in \{-1, +1\}^D$ defines a unique codeword in $\mathcal{C}$, and the corresponding token index is given by:
\begin{equation}
    \text{Index}(\mathbf{z}) = \sum_{i=1}^{D} 2^{i-1} \cdot \mathbb{I}_{[z_i > 0]}.
\end{equation}
}

LFQ thus avoids explicit codebook lookup and enables efficient discrete tokenization with binary latent representations, growing the vocabulary size in a way.

\subsection{Binary Spherical Quantization}
\label{subsec:Binary-Spherical-Quantization(BSQ)}

Binary Spherical Quantization (BSQ)~\cite{2024_arXiv_2025_ICLR_BSQ_Image-and-Video-Tokenization-with-Binary-Spherical-Quantization} employs a spherical projection-based quantization with binary encoding. An illustrative comparison between FSQ, LFQ, and BSQ in 2D is shown in Fig.~\ref{fig:BSQ}.
Compared with LFQ, BSQ has bounded reconstruction error by constraining the codebook on the unit hypersphere, enabling faster and better convergence for large-scale visual and video modeling.

\noindent\textbf{Definition [Binary Spherical Quantization].}
\textit{
Given an input vector $\mathbf{z} \in \mathbb{R}^D$, it is normalized on a unit sphere by $\ell_2$ normalization, i.e., $\mathbf{u} = \frac{\mathbf{v}}{\|\mathbf{v}\|_2} \in \mathbb{S}^{D-1}$. 
Binary Spherical Quantization defines an implicit codebook $\mathcal{C} = \left\{-\frac{1}{\sqrt{D}}, \frac{1}{\sqrt{D}}\right\}^D \subseteq \mathbb{S}^{D-1}$, where each codeword $\mathbf{c} \in \mathbb{R}^D$ satisfies $\|\mathbf{c}\|_2 = 1$.
BSQ quantizes $\mathbf{u}$ along each dimension via
\begin{equation}
\mathbf{c}_k=\hat{\mathbf{u}} = \frac{1}{\sqrt{D}} \cdot \text{sign}(\mathbf{u}) \in \mathcal{C}, 
k = \sum_{i=1}^{D} \mathbb{I}_{[u_i > 0]} \cdot 2^{i-1}, 
\end{equation}    
where sign($\cdot$) is sign function with sign(0) $= 1$, and $\mathbb{I}_{[\cdot]}$ is the indicator function.
}

\subsection{Graph Anchor-Relation Tokenization}
\label{subsec:Graph-Anchor-relation-Tokenization}

For the graphs, there is a common tokenization method, which we call Graph Anchor-Relation Tokenization (GART). It exploits the anchor-based graph representation learning technique, and tokenizes nodes into a composition of selected anchor nodes and relational context to form compact and discrete representations~\cite{2022_ICLR_NodePiece_NodePiece=Compositional-and-Parameter-Efficient-Representations-of-Large-Knowledge-Graphs, 2023_AAAI_EARL_Entity-agnostic-Representation-Learning-for-Parameter-efficient-Knowledge-Graph-Embedding, 2023_EMNLP_RandomEQ_Random-Entity-Quantization-for-Parameter-efficient-Compositional-Knowledge-Graph-Representation}. This tokenization design drastically reduces the vocabulary size while retaining expressiveness, especially for knowledge graphs.

\noindent\textbf{Definition [Graph Anchor-Relation Tokenization].}
\textit{
Let $\mathcal{V}$ denote the set of nodes in a graph and $\mathcal{R}$ be the set of relation types between nodes. The codebook is constructed by $M$ anchors and $|\mathcal{R}|$ relation types, where the anchor set $\mathcal{A} = \{a_1, a_2, \dots, a_M\}$ is pre-selected from $\mathcal{V}$ according to certain strategies and $M \ll |\mathcal{V}|$.
Given a target node $v \in \mathcal{V}$, it is matched to a composition of k-nearest anchors from $\mathcal{A}$ and its $d$ connected relations from $\mathcal{R}$, denoted as $\mathcal{W}=\{a_{i_1}, a_{i_2}, \dots, a_{i_k},  r_{j_1}, r_{j_2}, \dots, r_{j_d}\}$ where $a_{i_k} \in \mathcal{A}$ and $r_{j_d} \in \mathcal{R}$. Finally, node $v$ is tokenized into the codevector $E(\mathcal{W})$ through an encoder function $E(\cdot)$.
}

In practice, the encoder can adopt MLP, Transformer, or GNNs~\cite{2022_ICLR_NodePiece_NodePiece=Compositional-and-Parameter-Efficient-Representations-of-Large-Knowledge-Graphs, 2023_AAAI_EARL_Entity-agnostic-Representation-Learning-for-Parameter-efficient-Knowledge-Graph-Embedding, 2023_EMNLP_RandomEQ_Random-Entity-Quantization-for-Parameter-efficient-Compositional-Knowledge-Graph-Representation}. 
For anchor selection strategies, random selection works well, with Personalized PageRank (PPR) and degree-based strategies as alternatives.
And some methods also encode the distances between target nodes and anchors~\cite{2022_ICLR_NodePiece_NodePiece=Compositional-and-Parameter-Efficient-Representations-of-Large-Knowledge-Graphs} or multi-hop neighbors~\cite{2023_AAAI_EARL_Entity-agnostic-Representation-Learning-for-Parameter-efficient-Knowledge-Graph-Embedding} to retain topological information and semantic context.

\subsection{Discussion}

\textbf{(a) Codebook Collapse.}
Despite its strengths, discrete quantization faces a critical challenge—\textit{\textbf{codebook collapse}}, which limits codebook diversity and the expressiveness of quantized representations. 
\textit{
\textbf{Codebook collapse} refers to a situation where only a small subset of codewords are actively utilized during training, while the majority remain unused and don't get updated, called "dead codewords", resulting in low codebook utilization.
}

Beyond multi-codebooks, hierarchical structures and codebook-free designs in Table~\ref{tab:Codebook-Collapse_Comparison}, numerous methods have been developed to mitigate codebook collapse:
\textbf{\textit{(i) Code Reset.}}
HQA~\cite{2020_NeurlPS_HQA_Hierarchical-quantized-autoencoders} reinitializes unused codes near frequently used ones during training to mitigate under-utilization.
CVQ-VAE~\cite{2013_CVQ_VAE_Online-clustered-codebook} further introduces an online clustering strategy, dynamically reinitializing the unused codes based on running average statistics.
\textbf{\textit{(ii) Linear Reparameterization.}}
Recent methods apply linear transformation over code vectors to optimize codebooks for the collapse issue.
For example, \citet{2023_ICML_Straightening-Out-the-Straight-Through-Estimator=Overcoming-Optimization-Challenges-in-Vector-Quantized-Networks} proposes an affine reparameterization of codes by shared mean and standard deviation, assigning affine parameters to enable gradients to flow through unused codes.
VQGAN-LC~\cite{2024_arXv_VQGAN-LC_Scaling-the-codebook-size-of-vqgan-to-100000-with-a-utilization-rate-of-99} and SimVQ~\cite{2024_arXiv_SimVQ_Addressing-representation-collapse-in-vector-quantized-models-with-one-linear-layer} employ a learnable linear layer and reparameterize the codes to ensure all codes remain active.
\textbf{\textit{(iii) Soft Quantization.}}
Unlike nearest-neighbor hard assignments in standard VQ methods, soft quantization quantizes inputs as a weighted combination of codewords to improve codebook utilization.
IBQ~\cite{2024_arXiv_IBQ_Taming-scalable-visual-tokenizer-for-autoregressive-image-generation} applies STE on the one-hot categorical distribution between the encoded feature and codebook, letting all codes be selected equally.
The soft VQ-VAE~\cite{2020_AAAI_soft-VQ-VAE_Vector-quantization-based-regularization-for-autoencoders} and SCQ~\cite{2024_L4DC_SCQ_Soft-Convex-Quantization=Revisiting-Vector-Quantization-with-Convex-Optimization} quantize inputs as a convex combination problem of codewords.
SHVQ~\cite{2017_NeurlPS_SHVQ_Soft-to-hard-vector-quantization-for-end-to-end-learning-compressible-representations} and SQ-VAE~\cite{2022_arXiv_SQ_VAE_Sq-vae-Variational-bayes-on-discrete-representation-with-self-annealed-stochastic-quantization} also introduce the anneal mechanism, gradually approaching hard assignment from soft quantization during training.
Furthermore, HQ-VAE~\cite{2024_TMLR_HQ-VAE_Hq-vae=Hierarchical-discrete-representation-learning-with-variational-bayes} incorporates a hierarchical structure into SQ-VAE to mitigate the collapse issue.
\textbf{\textit{(iv) Regularization.}}
Several works introduce different regularization terms to improve codebook utilization.
HC-VQ~\cite{2022_arXiv_HC-VQ_Homology-constrained-vector-quantization-entropy-regularizer} proposes an entropy regularization based on persistent homology, indicating that higher entropy of VQ latent space is associated with higher codebook utilization.
On the other hand, Reg-VQ~\cite{2023_CVPR_Reg-VQ_Regularized-vector-quantization-for-tokenized-image-synthesis} introduces a prior distribution regularization where all codevectors are used, preventing collapse of the predicted token distribution.
VQ-WAE~\cite{2023_ICML_VQ-WAE_Vector-Quantized-Wasserstein-Auto-Encoder} combines a KL-regularization with Wasserstein distance approximated entropic regularized dual form, to match the codebook with latent data distribution.
Some methods~\cite{2024_ICLR_MAGVIT-v2_Language-Model-Beats-Diffusion-Tokenizer-is-Key-to-Visual-Generation, 2024_arXiv_IBQ_Taming-scalable-visual-tokenizer-for-autoregressive-image-generation, 2024_arXiv_2025_ICLR_BSQ_Image-and-Video-Tokenization-with-Binary-Spherical-Quantization} additionally add an entropy penalty to encourage codebook utilization.

\begin{table}[t]
    \centering
    \caption{Comparison of codebook collapse across quantization techniques.}
    \label{tab:Codebook-Collapse_Comparison}
    \renewcommand{\arraystretch}{1.5}
    \rowcolors{0}{mycolor_tab-1}{mycolor_tab-2}
    \scalebox{0.71}{
    \begin{tabular}{l l l}
    \toprule
    \textbf{Technique} &   \textbf{\makecell[l]{Collapse \\Mitigation}} &   \textbf{Reason / Comment} \\
    \midrule
    VQ (\S\ref{subsec:Vanilla-Vector-Quantization(VQ)}) &   No &   Learnable codebook easily collapses if not regularized \\
      RVQ (\S\ref{subsec:Residual-Vector-Quantization(RVQ)})       &   Partially &   Each stage has a learnable codebook; collapse still occurs \\
      PQ (\S\ref{subsec:Product-Quantization(PQ)})        &   Partially &   Multiple sub-codebooks reduce impact, but each can still collapse \\
      AQ (\S\ref{subsec:Additive-Vector-Quantization(AQ)})        &   Partially &   Additive structure softens collapse impact, but doesn't prevent it \\
      FSQ (\S\ref{subsec:Finite-Scalar-Quantization(FSQ)})       &   Yes &   Implicit fixed codebook; fixed scalar values \\
      LFQ (\S\ref{subsec:Look-up-Free-Quantization(LFQ)})       &   Yes &   Implicit fixed codebook; binary quantization \\
      BSQ (\S\ref{subsec:Binary-Spherical-Quantization(BSQ)})       &   Yes &   Implicit fixed codebook; binary quantization on the unit sphere \\
      GART (\S\ref{subsec:Graph-Anchor-relation-Tokenization}) &   Yes &   Uses shared anchors and relation types, avoiding the codebook \\
    \bottomrule
    \end{tabular}
    }
\end{table}

\begin{table*}[t]
    \centering
    \caption{Comprehensive comparison of quantization techniques on structural characteristics, significant pros and cons, commonly-supported modalities, and primary application scenarios.
    Denote Image: \faImage, Audio: \faMicrophone, Video: \faFilm, Action: \faCogs, Motion: \faWalking, Graph: \faProjectDiagram.
    Notably, for multi-codebook methods, we define $\bar{K} = \frac{1}{M} \sum_{i=1}^{M} K_i$ and 
    $\tilde{K} = \left( \prod_{i=1}^{M} K_i \right)^{\frac{1}{M}}$, 
    where $i \in \{1, 2, \ldots, M\}$.
    When $K_i = K$ for all $i$, $\bar{K} = \tilde{K} = K$.
    }
    \label{tab:VQs-Comprehensive-Comparison}
    \renewcommand{\arraystretch}{0.75}
    \scalebox{0.63}{
    \begin{tabular}{
    >{\centering\arraybackslash}m{1.3cm} |
    >{\centering\arraybackslash}m{0.8cm}
    >{\centering\arraybackslash}m{1.3cm}
    >{\centering\arraybackslash}m{1.3cm}
    >{\centering\arraybackslash}m{1.3cm}
    >{\centering\arraybackslash}m{1.3cm}
    >{\centering\arraybackslash}m{1.4cm} |
    >{\centering\arraybackslash}m{2.4cm} |
    >{\centering\arraybackslash}m{7cm} |
    >{\centering\arraybackslash}m{4cm}
    }
    \toprule
    \makecell[c]{\textbf{Technique}}
        & \makecell[c]{\textbf{\#Input} \\ \textbf{Dim.} }
        & \makecell[c]{\textbf{\#Codebook} \\ \textbf{Dim.}}
        & \makecell[c]{\textbf{\#Codebook}}
        & \makecell[c]{\textbf{\#Code} \\ \textbf{/Codebook}}
        & \makecell[c]{\textbf{\#Codebook}  \\ \textbf{Param.} }
        & \makecell[c]{\textbf{\#Codebook}  \\  \textbf{Size} }
        & \makecell[c]{\textbf{Modalities}}
        & \makecell[c]{\textbf{Significant \textcolor{pros}{Pros,} \textcolor{cons}{Cons}}}
        & \makecell[c]{\textbf{Primary Scenarios} \\  \textbf{(Tasks)}} \\
    \midrule

    \textbf{VQ}
        & \makecell[c]{$D$}
        & \makecell[c]{$D$}
        & \makecell[c]{$1$}
        & \makecell[c]{$K$}
        & \makecell[c]{$KD$}
        & \makecell[c]{$K$}
        & \makecell[c]{\faImage, \faMicrophone, \faFilm, \faCogs, \faWalking, \faProjectDiagram}
        & \makecell[c]{\textcolor{pros}{well-established;} \\ \textcolor{cons}{easy codebook collapse}}
        & \makecell[l]{Reconstruction, Compression, \\Generation, Understanding, etc.} \\
    \midrule

    \textbf{RVQ}
        & \makecell[c]{$D$}
        & \makecell[c]{$D$}
        & \makecell[c]{$M$}
        & \makecell[c]{$K_i$}
        & \makecell[c]{$M\bar{K}D$}
        & \makecell[c]{$\tilde{K}^M$}
        & \makecell[c]{\faImage, \faMicrophone, \faFilm, \faWalking, \faProjectDiagram}
        & \makecell[c]{\textcolor{pros}{support for variable bitrates;} \\ \textcolor{cons}{error propagation and accumulation}}
        & \makecell[l]{Reconstruction, Compression, \\Generation, etc.} \\
    \midrule

    \textbf{PQ}
        & \makecell[c]{$D$}
        & \makecell[c]{$D/M$}
        & \makecell[c]{$M$}
        & \makecell[c]{$K_i$}
        & \makecell[c]{$\bar{K}D$}
        & \makecell[c]{$\tilde{K}^M$}
        & \makecell[c]{\faImage}
        & \makecell[c]{\textcolor{pros}{efficient high-dimensional search;} \\ \textcolor{cons}{increased distortion due to independence assumption}}
        & \makecell[l]{ANN Search} \\
    \midrule

    \textbf{AQ}
        & \makecell[c]{$D$}
        & \makecell[c]{$D$}
        & \makecell[c]{$M$}
        & \makecell[c]{$K_i$}
        & \makecell[c]{$M\bar{K}D$}
        & \makecell[c]{$\tilde{K}^M$}
        & \makecell[c]{\faImage}
        & \makecell[c]{\textcolor{pros}{lower reconstruction error than PQ;} \\ \textcolor{cons}{higher training computational cost than PQ}}
        & \makecell[l]{ANN Search} \\
    \midrule

    \textbf{FSQ}
        & \makecell[c]{$D$}
        & \makecell[c]{$D$}
        & \makecell[c]{$1$}
        & \makecell[c]{$L^D$}
        & \makecell[c]{$0$}
        & \makecell[c]{$L^D$}
        & \makecell[c]{\faImage, \faMicrophone, \faFilm}
        & \makecell[c]{\textcolor{pros}{stable training, no collapse with no auxiliary losses,} \\ \textcolor{pros}{fewer parameters;} \textcolor{cons}{limited expressiveness}}
        & \makecell[l]{Reconstruction, Compression,\\ Generation, etc.} \\
    \midrule

    \textbf{LFQ}
        & \makecell[c]{$D$}
        & \makecell[c]{$0$}
        & \makecell[c]{$1$}
        & \makecell[c]{$2^D$}
        & \makecell[c]{$0$}
        & \makecell[c]{$2^D$}
        & \makecell[c]{\faImage, \faFilm}
        & \makecell[c]{\textcolor{pros}{no lookup;} \\ \textcolor{cons}{unbounded quantization error, limited expressiveness}}
        & \makecell[l]{Reconstruction, Compression,\\Generation, etc.} \\
    \midrule

    \textbf{BSQ}
        & \makecell[c]{$D$}
        & \makecell[c]{$D$}
        & \makecell[c]{$1$}
        & \makecell[c]{$2^D$}
        & \makecell[c]{$0$}
        & \makecell[c]{$2^D$}
        & \makecell[c]{\faImage, \faFilm}
        & \makecell[c]{\textcolor{pros}{upper-bounded quantization error, no $\mathcal{L}_{\mathrm{commitment}}$,} \\ \textcolor{pros}{more stable training than LFQ;} \textcolor{cons}{limited expressiveness}}
        & \makecell[l]{Reconstruction, Compression,\\ Generation, etc.} \\
    \midrule

    \textbf{GART}
        & \makecell[c]{$D$}
        & \makecell[c]{\textbf{-}}
        & \makecell[c]{\textbf{-}}
        & \makecell[c]{\textbf{-}}
        & \makecell[c]{\textbf{-}}
        & \makecell[c]{\textbf{-}}
        & \makecell[c]{\faProjectDiagram}
        & \makecell[c]{\textcolor{pros}{preserving graph topology;} \\ \textcolor{cons}{only graph-specific}}
        & \makecell[l]{Generation, Classification,\\ Prediction, etc.} \\
    \bottomrule
    \end{tabular}
    }
\end{table*}

\noindent 
\textbf{(b) Qualitative Comparison}. We provide a systematic comparison of quantization techniques in Table~\ref{tab:VQs-Comprehensive-Comparison}.

\textit{(i) Codebook Expressivity.} 
Codebook size in Table~\ref{tab:VQs-Comprehensive-Comparison} characterizes the upper bound of discrete representation space. Multi-codebook methods achieve exponentially greater expressivity via compositional codebooks than single-codebook counterparts. Methods without learnable codebooks project inputs to fixed values and are less adaptive to data distributions, inherently limiting their expressivity. But these methods simplify training and ensure stability.

\textit{(ii) Training Stability.}
Learnable codebook methods require auxiliary losses (e.g., commitment loss) and update strategies (e.g., EMA), offering stronger data adaptability but potentially introducing optimization instability. In contrast, FSQ/LFQ/BSQ eliminates codebook parameters and auxiliary losses, inherently ensuring training stability and being an attractive choice for training large-scale MLLMs.

\textit{(iii) Modality Coverage and Applicability.}
Despite the broad applicability of vanilla VQ across modalities, practical selection of quantization strategies has shifted towards modality-specific preferences.
For instance, RVQ is preferred in audio modeling due to its hierarchical coarse-to-fine design and support for variable bitrates. GART specifically provides a topology-preserving quantization mechanism for graphs. In visual modeling, recent trends~\cite{2025_arXiv_QLIP_QLIP=Text-Aligned-Visual-Tokenization-Unifies-Auto-Regressive-Multimodal-Understanding-and-Generation,2024_ICML_VideoPoet_VideoPoet=A-Large-Language-Model-for-Zero-shot-Video-Generation,2025_arXiv_HiTVideo_HiTVideo=Hierarchical-Tokenizers-for-Enhancing-Text-to-Video-Generation-with-Autoregressive-Large-Language-Models, 2024_arXiv_2025_ICLR_Show-o_Show-o=One-Single-Transformer-to-Unify-Multimodal-Understanding-and-Generation} also favor simpler implicit-codebook strategies for training stability.

\noindent 
\textbf{(c) Empirical Comparison.}
To compare the effectiveness of quantization methods, we evaluate mainstream methods in Table~\ref{tab:Performance-VQs-Reconstruction}. It's observed that FSQ/BSQ achieves competitive reconstruction and generation performance compared to VQGAN. Specifically, FSQ achieves the best generation performance, while BSQ achieves the highest rFID score, indicating that high-quality reconstruction and generation don't strictly depend on explicit learnable codebooks, particularly when optimization instability or codebook collapse may affect training.
Classical quantizations (e.g., RVQ and PQ) are originally designed for approximate nearest neighbor (ANN) search~\cite{2010_MDPI_RVQ_Approximate-nearest-neighbor-search-by-residual-vector-quantization, 2011_TPAMI_PQ_Product-Quantization-for-Nearest-Neighbor-Search}. Thus, we also compare their basic retrieval capabilities in Table~\ref {tab:Performance_VQs-ANN} to fundamentally assess the inherent differences in information fidelity. RVQ/CQ/AQ consistently outperforms PQ, suggesting that full-dimensional multi-codebook methods preserve feature discriminability more effectively than subspace splitting. 

Overall, multi-codebook designs favor information fidelity and structural expressivity, whereas methods with implicit codebooks offer stability and scalability for MLLMs.

\begin{table}[t]
    \centering
    \caption{Performance comparison of image reconstruction and generation on the ImageNet dataset (Res.: Resolution)~\cite{2025_arXiv_GQ_Vector-Quantization-using-Gaussian-Variational-Autoencoder}.}
    \label{tab:Performance-VQs-Reconstruction}
    \renewcommand{\arraystretch}{1.25}
    \scalebox{0.72}{
    \begin{tabular}{l | c c c c | c c}
    \toprule
    \multirow{2}{*}{\textbf{Technique}}
        & \multicolumn{4}{c}{\textbf{Reconstruction \scriptsize(Res. $256 \times 256$)}} 
        & \multicolumn{2}{c}{\textbf{Generation \scriptsize(Res. $128\times128$)}} \\
    \cmidrule(lr){2-5} \cmidrule(lr){6-7}
        & \textbf{PSNR} $\uparrow$
        & \textbf{LPIPS} $\downarrow$
        & \textbf{SSIM} $\uparrow$
        & \textbf{rFID} $\downarrow$
        & \textbf{gFID} $\downarrow$
        & \textbf{IS} $\uparrow$ \\
    \midrule
    VQGAN~\cite{2021_CVPR_VQGAN_Taming-transformers-for-high-resolution-image-synthesis}
        & 27.86
        & 0.062
        & 0.823
        & 1.228
        & 8.01
        & 151.40 \\
        
    FSQ~\cite{2024_ICLR_FSQ_Finite-scalar-quantization-Vq-vae-made-simple}
        & 28.83
        & 0.055
        & 0.842
        & 1.067
        & 7.33
        & 224.88 \\
        
    LFQ~\cite{2024_ICLR_MAGVIT-v2_Language-Model-Beats-Diffusion-Tokenizer-is-Key-to-Visual-Generation}
        & 27.87
        & 0.068
        & 0.829
        & 2.444
        & 7.73
        & 142.09 \\
        
    BSQ~\cite{2024_arXiv_2025_ICLR_BSQ_Image-and-Video-Tokenization-with-Binary-Spherical-Quantization}
        & 28.44
        & 0.051
        & 0.852
        & 0.700
        & 7.82
        & 221.64 \\
        
    \bottomrule
    \end{tabular}
    }
\end{table}

\section{Earlier Tokenization}
\label{Sec_Earlier-Tokenization}

Before the advent of LLMs, discrete tokenization via vector quantization, was widely used for efficient data compression and representation learning. This section reviews applications in image, audio, video, graph, and recommendation systems~\cite{2024_arXiv_Survey_Next-Token-Prediction-Towards-Multimodal-Intelligence=A-Comprehensive-Survey, 2025_ICLR_TAAE_Scaling-transformers-for-low-bitrate-high-quality-speech-coding, 2025_ICLR_HART_HART=Efficient-Visual-Generation-with-Hybrid-Autoregressive-Transformer, 2025_WWW_HQA-GAE_Hierarchical-Vector-Quantized-Graph-Autoencoder-with-Annealing-Based-Code-Selection, 2024_RecSys_CoST_CoST=Contrastive-Quantization-based-Semantic-Tokenization-for-Generative-Recommendation}, which lay the groundwork for modern multimodal systems by demonstrating the effectiveness of quantized representations on diverse data types, and continue to provide transferable insights and readily adaptable techniques for LLM multimodal modeling.

\begin{table}[t]
    \centering
    \caption{Performance comparison of fundamental approximate nearest neighbor (ANN) search on the SIFT1M dataset under 64-bit codes~\cite{2018_TMTA_Survey_A-survey-of-product-quantization}.}
    \label{tab:Performance_VQs-ANN}
    \renewcommand{\arraystretch}{1.3}
    \scalebox{0.72}{
    \begin{tabular}{l | c c c c c}
    \toprule
    \textbf{Technique}
        & \textbf{\#Code Per Book}
        & \textbf{\#Codebook}
        & \textbf{Recall@1~$\uparrow$}
        & \textbf{Recall@10~$\uparrow$}
        & \textbf{Recall@100~$\uparrow$} \\
    \midrule
    
    RVQ~\cite{2010_MDPI_RVQ_Approximate-nearest-neighbor-search-by-residual-vector-quantization}  
        & 256
        & 8
        & 0.259
        & 0.658
        & 0.954 \\
        
    PQ~\cite{2011_TPAMI_PQ_Product-Quantization-for-Nearest-Neighbor-Search} 
        & 256
        & 8
        & 0.224
        & 0.599
        & 0.924 \\
        
    CQ~\cite{2014_ICML_CQ_Composite-quantization-for-approximate-nearest-neighbor-search} 
        & 256
        & 8
        & 0.288
        & 0.716
        & 0.967 \\
        
    AQ~\cite{2014_CVPR_AQ_Additive-quantization-for-extreme-vector-compression} 
        & 256
        & 8
        & 0.298
        & 0.741
        & 0.972 \\
        
    \bottomrule
    \end{tabular}
    }
\end{table}

\subsection{Image}
\label{sec_non-LLM_Image}
Discrete tokenization has been widely used in image retrieval~\cite{2021_ICCV_SPQ_Self-supervised-product-quantization-for-deep-unsupervised-image-retrieval, 2022_AAAI_MeCoQ_Contrastive-quantization-with-code-memory-for-unsupervised-image-retrieval}, generation~\cite{2024_NeurlPS_TiTok_An-image-is-worth-32-tokens-for-reconstruction-and-generation, 2024_TMLR_MaskBit_MaskBit=Embedding-free-Image-Generation-via-Bit-Tokens}, and representation learning~\cite{2022_ICLR_BEiT_BEIT=BERT-Pre-Training-of-Image-Transformers, 2024_ICLR_ClusterMIM_On-the-role-of-discrete-tokenization-in-visual-representation-learning}. Quantized visual tokens enabled compact and expressive image modeling.

\textbf{\textit{(i) Image Retrieval.}}
DVSQ~\cite{2017_CVPR_DVSQ_Deep-visual-semantic-quantization-for-efficient-image-retrieval} jointly learns visual-semantic embeddings and quantizers for efficient image retrieval with compact binary codes.
Deep Progressive Quantization (DPQ)~\cite{2019_IJCAI_Deep-progressive-quantization-(DPQ)_Beyond-product-quantization=Deep-progressive-quantization-for-image-retrieval} learns codes of varying lengths by progressively approximating the feature space for large-scale image retrieval.
In~\cite{2021_ICCV_SPQ_Self-supervised-product-quantization-for-deep-unsupervised-image-retrieval}, SPQ achieves self-supervised product quantization via cross-quantized contrastive learning for unsupervised image retrieval.
MeCoQ~\cite{2022_AAAI_MeCoQ_Contrastive-quantization-with-code-memory-for-unsupervised-image-retrieval} introduces contrastive unsupervised quantization with code memory for reduced drift and regularization to prevent degeneration.

\textbf{\textit{(ii) Image Synthesis.}}
MaskGIT~\cite{2022_CVPR_MaskGIT_Maskgit=Masked-generative-image-transformer} tokenizes images via VQGAN and uses a bidirectional Transformer decoder to predict masked tokens for synthesis.
On the other hand, RQ-Transformer~\cite{2022_CVPR_RQ-VAE_RQ-Transformer_Autoregressive-image-generation-using-residual-quantization} and DnD-Transformer~\cite{2025_ICLR_DnD-Transformer_A-spark-of-vision-language-intelligence=2-dimensional-autoregressive-transformer-for-efficient-finegrained-image-generation} quantize feature maps of images by RQ-VAE based on RVQ for 2D autoregressive generation.
In addition, ViT-VQGAN~\cite{2022_ICLR_ViT-VQGAN_Vector-quantized-Image-Modeling-with-Improved-VQGAN} and Efficient-VQGAN~\cite{2023_ICCV_Efficient-VQGAN_Efficient-vqgan=Towards-high-resolution-image-generation-with-efficient-vision-transformers} replace the CNN with a vision Transformer for improved reconstruction, with ViT-VQGAN also introducing factor and $\ell_{2}$-normalized codes for better codebook usage.
Both MQ-VAE~\cite{2023_CVPR_MQ-VAE_Not-all-image-regions-matter=Masked-vector-quantization-for-autoregressive-image-generation} and DQ-VAE~\cite{2023_CVPR_DQ-VAE_Towards-accurate-image-coding-Improved-autoregressive-image-generation-with-dynamic-vector-quantization} consider the codebook redundancy issue caused by ignoring different perceptual importance of image regions. MQ-VAE designs adaptive mask and de-mask modules while DQ-VAE proposes variable-length codes.
To unify image generation and representation learning, additional methods have also made efforts~\cite{2022_NeurlPS_MoVQ_Movq=Modulating-quantized-vectors-for-high-fidelity-image-generation, 2023_CVPR_MAGE_Mage=Masked-generative-encoder-to-unify-representation-learning-and-image-synthesis, 2024_Neurl_VQ-KD_Image-understanding-makes-for-a-good-tokenizer-for-image-generation, 2024_CVPR_SeQ-GAN_Rethinking-the-objectives-of-vector-quantized-tokenizers-for-image-synthesis, 2025_CVPR_MergeVQ_MergeVQ=A-Unified-Framework-for-Visual-Generation-and-Representation-with-Disentangled-Token-Merging-and-Quantization, 2024_NeurlPS_VAR_Visual-autoregressive-modeling=Scalable-image-generation-via-next-scale-prediction, 2024_NeurlPS_TiTok_An-image-is-worth-32-tokens-for-reconstruction-and-generation, 2025_arXiv_FlowMo_Flow-to-the-Mode=Mode-Seeking-Diffusion-Autoencoders-for-State-of-the-Art-Image-Tokenization, 2024_TMLR_MaskBit_MaskBit=Embedding-free-Image-Generation-via-Bit-Tokens}, more details can be found in Appendix~\ref{appendix_subsec-non-LLM_Image}.

\textbf{\textit{(iii) Image Classification.}}
BEiT~\cite{2022_ICLR_BEiT_BEIT=BERT-Pre-Training-of-Image-Transformers} introduces BERT-style masked image modeling for vision Transformers by predicting discrete tokens from masked patches.
Building on BEiT,
BEiT v2~\cite{2022_arXiv_BEiT-v2_Beit-v2=masked-image-modeling-with-vector-quantized-visual-tokenizers} proposes VQ-KD to train a semantic visual tokenizer, pushing masked image modeling (MIM) beyond pixel-level targets.
Further exploring tokenizer design,
ClusterMIM~\cite{2024_ICLR_ClusterMIM_On-the-role-of-discrete-tokenization-in-visual-representation-learning} introduces a label-free clustering tokenizer for MIM and the TCAS metric to evaluate its quality.

\subsection{Audio}
\label{sec_non-LLM_Audio}
Recent developments in audio modeling leverage discrete tokenization for efficient compression and self-supervised representation learning~\cite{2019_arXiv_2020_ICLR_vq-wav2vec_vq-wav2vec-Self-supervised-learning-of-discrete-speech-representations, 2020_NeurlPS_wav2vec-2.0_wav2vec-2.0=A-Framework-for-Self-Supervised-Learning-of-Speech-Representations}, primarily through neural codecs~\cite{2024_J-STSP_SemantiCodec_SemantiCodec=An-Ultra-Low-Bitrate-Semantic-Audio-Codec-for-General-Sound, 2025_IEEE-Signal-Processing-Letter_StreamCodec_A-Streamable-Neural-Audio-Codec-with-Residual-Scalar-Vector-Quantization-for-Real-Time-Communication, 2025_arXiv_SQCodec_One-Quantizer-is-Enough=Toward-a-Lightweight-Audio-Codec} and quantized speech tokens.

\textbf{\textit{(i) Self-Supervised Speech Representation via Discrete Units.}}
VQ-wav2vec~\cite{2019_arXiv_2020_ICLR_vq-wav2vec_vq-wav2vec-Self-supervised-learning-of-discrete-speech-representations} and Wav2vec~\cite{2020_NeurlPS_wav2vec-2.0_wav2vec-2.0=A-Framework-for-Self-Supervised-Learning-of-Speech-Representations} introduce a BERT-style self-supervised contrastive paradigm for modeling speech representations from raw audio.

\textbf{\textit{(ii) High-Fidelity Audio Compression with Discrete Tokens.}}
SoundStream~\cite{2022_TASLP_SoundStream_SoundStream=An-End-to-End-Neural-Audio-Codec} uses RVQ with structured dropout, trained by the VQGAN formulation for unified codecs at variable bitrates.
HiFi-Codec~\cite{2023_arXiv_HiFi-Codec_HiFi-Codec=Group-residual-Vector-quantization-for-High-Fidelity-Audio-Codec} introduces group-residual vector quantization with only four codebooks, Encodec~\cite{2023_TMLR_EnCodec_High-Fidelity-Neural-Audio-Compression} develops a multiscale spectrogram adversary and loss balancer, and DAC~\cite{2023_NeurlPS_DAC_High-fidelity-neural-audio-compression-with-Improved-RVQGAN} adds periodic inductive biases.
LMCodec~\cite{2023_ICASSP_LMCodec_LMCodec=A-Low-Bitrate-Speech-Codec-with-Causal-Transformer-Models} introduces a fully causal transformer with conditional entropy coding for the low-bitrate speech codec.
WavTokenizer~\cite{2025_ICLR_WavTokenizer_WavTokenizer=An-Efficient-Acoustic-Discrete-Codec-Tokenizer-for-Audio-Language-Modeling} proposes a single-quantizer acoustic codec by expanding the VQ space and redesigning the decoder with an inverse-Fourier-based architecture.

\textbf{\textit{(iii) Semantic-Aware and General-Purpose Tokenization.}}
SemantiCodec~\cite{2024_J-STSP_SemantiCodec_SemantiCodec=An-Ultra-Low-Bitrate-Semantic-Audio-Codec-for-General-Sound} consists of semantic and acoustic encoders, dual-layer vector quantization, and a diffusion-based decoder, supporting diverse audio types.
Along the same lines,
SpeechTokenizer~\cite{2024_ICLR_SpeechTokenizer_SpeechTokenizer=Unified-Speech-Tokenizer-for-Speech-Language-Models} unifies semantic and acoustic tokens for speech language modeling, hierarchically disentangling speech information across RVQ layers.

\textbf{\textit{(iv) Real-Time and Lightweight Audio Codecs.}}
StreamCodec~\cite{2025_IEEE-Signal-Processing-Letter_StreamCodec_A-Streamable-Neural-Audio-Codec-with-Residual-Scalar-Vector-Quantization-for-Real-Time-Communication} is a streamable causal audio codec for real-time communication with residual scalar-vector quantization to enhance codebook utilization.
Similarly,
SQCodec~\cite{2025_arXiv_SQCodec_One-Quantizer-is-Enough=Toward-a-Lightweight-Audio-Codec} designs a single-quantizer architecture based on TConv module and FSQ for the lightweight audio codec.
A broader set of representative works and further details~\cite{2025_arXiv_UniCodec_UniCodec=Unified-Audio-Codec-with-Single-Domain-Adaptive-Codebook, 2025_arXiv_QinCodec_QINCODEC=Neural-Audio-Compression-with-Implicit-Neural-Codebooks, 2025_ICLR_TAAE_Scaling-transformers-for-low-bitrate-high-quality-speech-coding, 2025_ICASSP_LFSC_Low-Frame-rate-Speech-Codec=A-Codec-Designed-for-Fast-High-quality-Speech-LLM-Training-and-Inference} can be found in Appendix~\ref{appendix_subsec-non-LLM_Audio}.

\subsection{Graph}
\label{sec_non-LLM_Graph}
Graphs are ubiquitous in domains such as knowledge graphs~\cite{2020_ACL_TS-CL_Knowledge-graph-embedding-compression} and molecular systems~\cite{2023_NeurlPS_iMoLD_Learning-invariant-molecular-representation-in-latent-discrete-space}. Their non-Euclidean structure and the requirement for permutation invariance pose fundamental challenges for scalable and effective modeling~\cite{2022_ICLR_NodePiece_NodePiece=Compositional-and-Parameter-Efficient-Representations-of-Large-Knowledge-Graphs, 2023_NeurlPS_iMoLD_Learning-invariant-molecular-representation-in-latent-discrete-space, 2024_TMLR_DGAE_Discrete-Graph-Auto-Encoder, 2025_AAAI_GLAD_Glad=Improving-latent-graph-generative-modeling-with-simple-quantization, 2025_ICLR_NID_Node-Identifiers=Compact-Discrete-Representations-for-Efficient-Graph-Learning}. To address these issues, discrete tokenization techniques have emerged as a compact and interpretable alternative for graph representation, enabling scalable modeling and structure-aware representation learning.

\textbf{\textit{(i) Graph Representation Compression.}} TS-CL~\cite{2020_ACL_TS-CL_Knowledge-graph-embedding-compression} and LightKG~\cite{2021_CIKM_LightKG_A-lightweight-knowledge-graph-embedding-framework-for-efficient-inference-and-storage} leverage discrete codes to compress knowledge graph embeddings for efficient storage and inference. Similarly, SNEQ~\cite{2020_AAAI_SNEQ_SNEQ=Semi-supervised-attributed-network-embedding-with-attention-based-quantisation} and d-SNEQ~\cite{2021_TNNLS_d-SNEQ_Semisupervised-network-embedding-with-differentiable-deep-quantization} learn low-dimensional network embeddings under semi-supervised settings by self-attention-based and autoencoder-based PQ, respectively. NID~\cite{2025_ICLR_NID_Node-Identifiers=Compact-Discrete-Representations-for-Efficient-Graph-Learning} learns compact discrete node codes by compressing GNN layers, enabling interpretable graph tokenization.

\textbf{\textit{(ii) Molecular Representation Learning.}}
iMoLD~\cite{2023_NeurlPS_iMoLD_Learning-invariant-molecular-representation-in-latent-discrete-space} learns distribution-invariant molecular representations via a first-encode-then-separate paradigm and task-agnostic self-supervised objective.
Similarly,
MOLE-BERT~\cite{2023_ICLR_Mole-BERT_Mole-BERT=Rethinking-Pre-training-Graph-Neural-Networks-for-Molecules} introduces a context-aware tokenizer with group VQ-VAE and a joint pretraining framework combining masked atom modeling and contrastive learning.
For efficient PPI modeling,
MAPE-PPI~\cite{2024_ICLR_MAPE-PPI_MAPE-PPI=Towards-Effective-and-Efficient-Protein-Protein-Interaction-Prediction-via-Microenvironment-Aware-Protein-Embedding} encodes protein microenvironments into discrete codes and masks the codebook.

\textbf{\textit{(iii) Graph Generation.}}
DGAE~\cite{2024_TMLR_DGAE_Discrete-Graph-Auto-Encoder} and GLAD~\cite{2025_AAAI_GLAD_Glad=Improving-latent-graph-generative-modeling-with-simple-quantization} both improve permutation-invariant graph generation in discrete latent spaces, where DGAE introduces a graph-to-set autoencoder with an autoregressive 2D-Transformer, while GLAD introduces a diffusion model with diffusion bridges.
Additionally,
Appendix~\ref{appendix_subsec-non-LLM_Graph} further discusses representative methods with varied design choices and objectives~\cite{2024_ICLR_VQGraph_VQGraph=Rethinking-Graph-Representation-Space-for-Bridging-GNNs-and-MLPs, 2024_NeurlPS_GFT_Graph-Foundation-Model-with-Transferable-Tree-Vocabulary, 2025_WWW_HQA-GAE_Hierarchical-Vector-Quantized-Graph-Autoencoder-with-Annealing-Based-Code-Selection, 2025_ICLR_GQT_Learning-Graph-Quantized-Tokenizers, 2025_arXiv_GT-SVQ_GT-SVQ=A-Linear-Time-Graph-Transformer-for-Node-Classification-Using-Spiking-Vector-Quantization} on other directions.

\subsection{Video}
\label{sec_non-LLM_Video}
Discrete tokenization is also an essential component in video modeling, enabling compact representations of spatio-temporal information. Recent work explores its use in video synthesis, compression, and unified representation learning across diverse temporal scales.
VideoGPT~\cite{2021_arXiv_VideoGPT_VideoGPT=Video-Generation-using-VQ-VAE-and-Transformers} leverages VQ-VAE with 3D convolutions to obtain spatio-temporally aware discrete latent representations of videos.
To better support flexible long video generation, TATS~\cite{2022_ECCV_TATS_Long-Video-Generation-with-Time-agnostic-VQGAN-and-Time-sensitive-Transformer} uses a time-agnostic VQGAN to tokenize videos into temporally agnostic codes.
MAGVIT~\cite{2023_CVPR_MAGVIT_MAGVIT=Masked-Generative-Video-Transformer} introduces a 3D tokenizer that quantizes videos into spatiotemporal tokens for unified video synthesis.
As a subsequent extension,
MAGVIT-v2~\cite{2024_ICLR_MAGVIT-v2_Language-Model-Beats-Diffusion-Tokenizer-is-Key-to-Visual-Generation} shows that strong visual tokenizers enable autoregressive LMs to outperform diffusion models in image and video generation.
BSQ-ViT~\cite{2024_arXiv_2025_ICLR_BSQ_Image-and-Video-Tokenization-with-Binary-Spherical-Quantization} is a unified image-video tokenizer using a transformer with block-wise causal masking and BSQ for variable-length inputs.
Several additional efforts have been made from different perspectives~\cite{2023_ICLR_Phenaki_Phenaki=Variable-Length-Video-Generation-from-Open-Domain-Textual-Description, 2024_NeurlPS_OmniTokenizer_OmniTokenizer=A-Joint-Image-Video-Tokenizer-for-Visual-Generation, 2024_arXiv_LARP_LARP=Tokenizing-Videos-with-A-Learned-Autoregressive-Generative-Prior, 2024_arXiv_VidTok_VidTok=A-Versatile-and-Open-source-Video-Tokenizer, 2024_arXiv_VQ-NeRV_VQ-NeRV=A-Vector-Quantized-Neural-Representation-for-Videos, 2024_arXiv_SweetTokenizer_SweetTokenizer=Semantic-aware-Spatial-temporal-Tokenizer-for-Compact-Visual-Discretization, 2025_arXiv_TVC_TCV=Tokenized-Video-Compression-with-Ultra-low-Bitrate}, and further details are in Appendix~\ref{appendix_subsec-non-LLM_Video}.

\subsection{Action}
\label{sec_non-LLM_Action}
Discrete tokenization has also been explored in action modeling, particularly for encoding continuous control signals into compact action tokens. Early efforts focus on quantization for reinforcement learning and efficient temporal abstraction.  
SAQ~\cite{2023_CoRL_SAQ_Action-quantized-Offline-Reinforcement-Learning-for-Robotic-Skill-Learning} introduces a VQ-VAE~\cite{2017_NeurlPS_VQ-VAE_Neural-discrete-representation-learning}-based offline RL framework that discretizes actions by state, enabling more stable policy learning.
To enhance temporal abstraction in control,
PRISE~\cite{2024_ICML_PRISE_PRISE=LLM-style-Sequence-Compression-for-Learning-Temporal-Action-Abstractions-in-Control} employs VQ for action discretization and designs byte pair encoding (BPE) to extract skill tokens for efficient sequence modeling.

\subsection{Multiple Modalities}
\label{sec:non-LLM_MM}
\vspace{0.3em}
\noindent \textbf{(a) Text + Image.}
In text-image tasks, discrete tokenization serves as a bridge between visual and linguistic modalities. It enables unified token spaces for text-to-image generation and multimodal representation learning.

\textbf{\textit{(i) Unified Discrete Token Spaces for Generation.}}
DALL-E~\cite{2021_ICML_DALL-E_Zero-shot-Text-to-image-Generation} and CogView~\cite{2021_NeruIPS_CogView_CogView=Mastering-Text-to-image-Generation-via-Transformer} both model text and image tokens jointly via a transformer for text-to-image generation, where image tokens are obtained through dVAE and VQ-VAE~\cite{2017_NeurlPS_VQ-VAE_Neural-discrete-representation-learning}, respectively.
Incorporating VQ-VAE~\cite{2017_NeurlPS_VQ-VAE_Neural-discrete-representation-learning} with DDPM in the token space, VQ-Diffusion~\cite{2022_CVPR_VQ-Diffusion_Vector-Quantized-Diffusion-Model-for-Text-to-image-Synthesis} enables efficient, high-fidelity text-to-image generation.
For more controllable generation, 
Make-A-Scene~\cite{2022_ECCV_Make-A-Scene_Make-A-Scene=Scene-based-Text-to-image-Generation-with-Human-Priors} uses discrete prompts and layouts to guide aligned scene construction.
Unified-IO~\cite{2023_ICLR_Unified-IO_Unified-IO=A-Unified-Model-for-Vision-Language-and-Multi-modal-Tasks} unifies vision and language tasks by converting all inputs and outputs into discrete sequences for unified sequence modeling.
Muse~\cite{2023_ICML_Muse_Muse=Text-to-image-Generation-via-Masked-Generative-Transformers} models masked VQ tokens conditioned on text and reconstructs them iteratively in base and super-resolution stages.
Focusing on tokenizer design,
UniTok~\cite{2025_arXiv_UniToK_UniTok=A-Unified-Tokenizer-for-Visual-Generation-and-Understanding} introduces a unified tokenizer with multi-codebook quantization for high-fidelity generation and semantic understanding.
Recently, HART~\cite{2025_ICLR_HART_HART=Efficient-Visual-Generation-with-Hybrid-Autoregressive-Transformer} combines discrete VQ and residual continuous tokens for efficient high-resolution image generation.

\textbf{\textit{(ii) Language-Guided Tokenization and Alignment.}}
NUWA-LIP~\cite{2023_CVPR_NUWA-LIP_NUWA-LIP=Language-guided-Image-Inpainting-with-Defect-free-VQGAN} improves language-guided image inpainting via a defect-free VQGAN~\cite{2021_CVPR_VQGAN_Taming-transformers-for-high-resolution-image-synthesis}, fusing semantic and visual cues.
Similarly,
TexTok~\cite{2024_arXiv_TexTok_Language-Guided-Image-Tokenization-for-Generation} introduces a text-conditioned tokenizer that improves both continuous and discrete tokenization quality.
LG-VQ~\cite{2024_NeurIPS_LG-VQ_LG-VQ=Language-guided-Codebook-Learning} and TokLIP~\cite{2025_arXiv_TokLIP_TokLIP_TokLIP=Marry-Visual-Tokens-to-CLIP-for-Multimodal-Comprehension-and-Generation} both align visual tokens with textual semantics to enhance multimodal understanding, through language-guided codebook learning~\cite{2024_NeurIPS_LG-VQ_LG-VQ=Language-guided-Codebook-Learning} and CLIP-aligned token encoders~\cite{2025_arXiv_TokLIP_TokLIP_TokLIP=Marry-Visual-Tokens-to-CLIP-for-Multimodal-Comprehension-and-Generation}, respectively.
Extending to knowledge graphs,
MyGO~\cite{2025_AAAI_MyGO_MyGO=Discrete-Modality-Information-as-Fine-Grained-Tokens-for-Multi-modal-Knowledge-Graph-Completion} tokenizes multimodal data into fine-grained tokens and boosts entity representations via contrastive learning.

\vspace{0.3em}
\noindent \textbf{(b) Text + Audio.}
Discrete tokenization has been explored for aligning textual and acoustic modalities, particularly in text-to-speech synthesis. These methods leverage quantized speech representations to enable controllable, high-fidelity generation and efficient language-to-audio modeling.

\textbf{\textit{(i) Tokenization-Driven Generative Modeling.}}
Specifically,
AudioGen~\cite{2023_ICLR_AudioGen_AudioGen=Textually-Guided-Audio-Generation} proposes text and audio mixing augmentations for text-to-audio generation, improving compositionality.
Similarly,
NaturalSpeech 3~\cite{2024_ICML_NaturalSpeech-3_NaturalSpeech-3=Zero-Shot-Speech-Synthesis-with-Factorized-Codec-and-Diffusion-Models} introduces a factorized diffusion model for TTS on disentangled subspaces via a factorized neural speech codec (FACodec).
To enable high-quality TTS,
Spectral Codec~\cite{2024_arXiv_Spectral-Codecs=Spectrogram-Based-Audio-Codecs-for-High-Quality-Speech-Synthesis} and Single-Codec~\cite{2024_Interspeech_Single-Codec_Single-Codec=Single-Codebook-Speech-Codec-towards-High-Performance} tokenize mel-spectrograms using FSQ and single-codebook VQ-VAE, respectively.
The SimpleSpeech series~\cite{2024_Interspeech_SimpleSpeech_SimpleSpeech=Towards-Simple-and-Efficient-Text-to-Speech-with-Scalar-Latent-Transformer-Diffusion-Models, 2024_arXiv_SimpleSpeech-2_SimpleSpeech-2=Towards-Simple-and-Efficient-Text-to-Speech-with-Flow-based-Scalar-Latent-Transformer-Diffusion-Models} focuses on efficient TTS with scalar quantization and diffusion.
Concretely, 
SimpleSpeech~\cite{2024_Interspeech_SimpleSpeech_SimpleSpeech=Towards-Simple-and-Efficient-Text-to-Speech-with-Scalar-Latent-Transformer-Diffusion-Models} proposes scalar-quantized speech codec (SQ-Codec) and transformer-based diffusion, while SimpleSpeech 2~\cite{2024_arXiv_SimpleSpeech-2_SimpleSpeech-2=Towards-Simple-and-Efficient-Text-to-Speech-with-Flow-based-Scalar-Latent-Transformer-Diffusion-Models} further introduces Time MoE and flow-based diffusion.

\textbf{\textit{(ii) Codec-Based Language Modeling.}}
VALL-E~\cite{2023_arXiv_VALL-E_Neural-Codec-Language-Models-are-Zero-Shot-Text-to-Speech-Synthesizers} and VALL-E X~\cite{2023_arXiv_VALL-E-X_Speak-Foreign-Languages-with-Your-Own-Voice=Cross-Lingual-Neural-Codec-Language-Modeling} model text-to-speech synthesis (TTS) as conditional language modeling on discrete codec tokens in monolingual and cross-lingual settings, respectively, enabling in-context learning capabilities in zero-shot scenarios.
To enhance robustness,
RALL-E~\cite{2024_arXiv_RALL-E_RALL-E=Robust-Codec-Language-Modeling-with-Chain-of-Thought-Prompting-for-Text-to-Speech-Synthesis} introduces chain-of-thought (CoT) prompting to improve the stability of TTS generation.
Further,
HALL-E~\cite{2024_arXiv_2025_ICLR_HALL-E_HALL-E=Hierarchical-Neural-Codec-Language-Model-for-Minute-Long-Zero-Shot-Text-to-Speech-Synthesis} introduces a post-training approach that hierarchically reorganizes discrete tokens through knowledge distillation, reducing frame rate for minute-long TTS.

\vspace{0.3em}
\noindent \textbf{(c) Audio + Video.}
Discrete tokenization has also been applied to joint audio-video modeling, enabling unified representations for multimodal tasks. Initial efforts demonstrate its potential in audiovisual understanding.  
VQ-MAE-AV~\cite{2025_CVIU_VQ-MAE-AV_A-vector-quantized-masked-autoencoder-for-audiovisual-speech-emotion-recognition} introduces a vector-quantized masked autoencoder for audiovisual speech emotion recognition, which learns discrete audio-visual speech representations via self-supervised multimodal fusion.

\vspace{0.3em}
\noindent \textbf{(d) Audio + Motion.}
In audio-action tasks, discrete tokenization enables mapping speech to compact motion representations. A representative approach is outlined below.
ProTalk~\cite{2024_CVPR_ProTalk_Towards-variable-and-Coordinated-Holistic-Co-speech-Motion-Generation} introduces a PQ-based non-autoregressive framework for generating diverse and coordinated full-body co-speech motions, integrating structured quantization and motion refinement for realism.

\begin{figure}[t]
    \centering
    \includegraphics[width=0.8\linewidth]{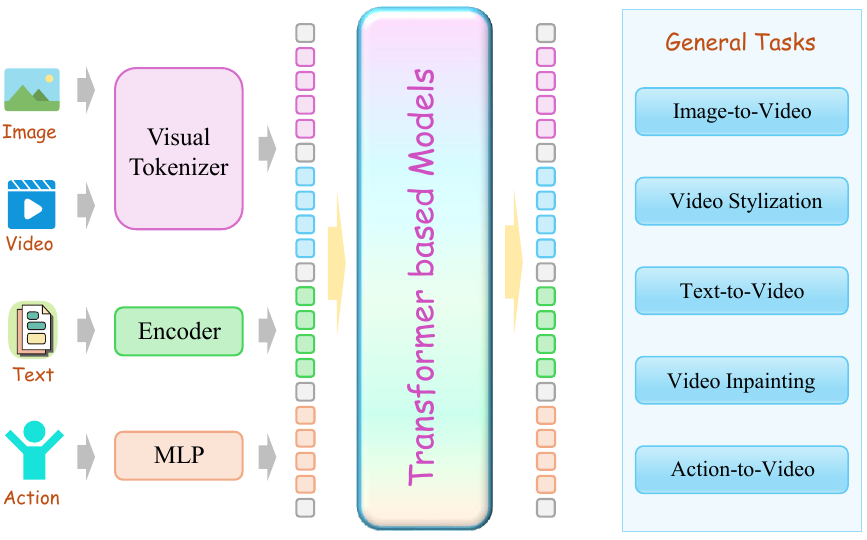}
    \caption{Non-LLM-based multimodal pipeline that encodes each modality via specialized modules before fusion through transformer-based models for different tasks~\cite{2024_arXiv_WorldDreamer_Dreamer=Towards-General-World-Models-for-Video-Generation-via-Predicting-Masked-Tokens}.}
    \label{fig:nonLLM-Text-Image-Video-Action}
\end{figure}

\vspace{0.3em}
\noindent \textbf{(e) Audio + Image + Video.}
In multimodal synthesis involving audio, image, and video, discrete tokenization enables compact control over facial motion.
For instance, VQTalker~\cite{2025_AAAI_VQTalker_VQTalker=Towards-Multilingual-Talking-Avatars-through-Facial-Motion-Tokenization} employs group residual scalar quantization for facial motion tokenization, enabling high-fidelity multilingual talking head synthesis at low bitrates.

\vspace{0.3em}
\noindent \textbf{(f) Text + Image + Video + Action.}
Token-based representations have been extended to unify text, image, video, and action modalities, supporting the modeling of complex spatio-temporal dynamics within a shared discrete space. As illustrated in Fig.~\ref{fig:nonLLM-Text-Image-Video-Action}, such systems typically adopt modality-specific tokenizers followed by a transformer-based fusion module. 
WorldDreamer~\cite{2024_arXiv_WorldDreamer_Dreamer=Towards-General-World-Models-for-Video-Generation-via-Predicting-Masked-Tokens} models various video generation tasks as masked visual token prediction by the proposed spatial-temporal patchwise transformer on discrete visual tokens across general world physics and motions.

\vspace{0.3em}
\noindent \textbf{(g) Recommendation Systems.}
Discrete tokenization in recommendation systems supports compact modeling of behaviors and semantics, enabling more efficient, transferable, and generative recommendation frameworks.
Specifically,
MGQE~\cite{2020_WWW_MGQE_Learning-Multi-granular-Quantized-Embeddings-for-Large-Vocab-Categorical-Features-in-Recommender-Systems} extends DPQ~\cite{2020_ICML_Differentiable-Product-Quantization-(DPQ)_Differentiable-product-quantization-for-end-to-end-embedding-compression} with variable capacities to handle the power-law distribution, learning compact embeddings for recommendation.
ReFRS~\cite{2023_TOIS_ReFRS_ReFRS=Resource-efficient-Federated-Recommender-System-for-Dynamic-and-Diversified-User-Preferences} and VQ-Rec~\cite{2023_WWW_VQ-Rec_Learning-Vector-Quantized-Item-Representation-for-Transferable-Sequential-Recommenders} both employ vector-quantized representations for sequential recommendation. 
Additional works~\cite{2023_TOIS_ReFRS_ReFRS=Resource-efficient-Federated-Recommender-System-for-Dynamic-and-Diversified-User-Preferences, 2023_WWW_VQ-Rec_Learning-Vector-Quantized-Item-Representation-for-Transferable-Sequential-Recommenders, 2023_NeurlPS_TIGER_Recommender-Systems-with-Generative-Retrieval, 2024_RecSys_CoST_CoST=Contrastive-Quantization-based-Semantic-Tokenization-for-Generative-Recommendation, 2024_KDD_EAGER_EAGER=Two-Stream-Generative-Recommender-with-Behavior-Semantic-Collaboration} are discussed in Appendix~\ref{appendix_subsec-non-LLM_RecSys}.

\section{LLMs with Single Modality}
\label{Sec_LLM-based-Single-Modality-Applications}

LLMs have demonstrated remarkable capabilities in generation, understanding, and generalization across various tasks, making them an attractive backbone for modeling other non-text modalities. To benefit from these powerful capabilities of LLMs, recent studies~\cite{2023_NeurlPS_LQAE_Language-Quantized-AutoEncoders=Towards-Unsupervised-Text-Image-Alignment, 2023_NeurlPS_TWIST_Textually-Pretrained-Speech-Language-Models, 2024_KDD_EAGER_EAGER=Two-Stream-Generative-Recommender-with-Behavior-Semantic-Collaboration} have explored how to encode single non-text modalities into LLM-readable tokens via discrete tokenization, e.g., mapping data features into LLMs' vocabulary space without explicit text inputs~\cite{2023_NeurIPS_SPAE_SPAE=Semantic-Pyramid-AutoEncoder-for-Multimoda-Generation-with-Frozen-LLMs}. This section reviews how such discrete tokens serve as a bridge that allows LLMs to complete downstream tasks like node classification~\cite{2025_AAAI_Dr.E_Multi-View-Empowered-Structural-Graph-Wordification-for-Language-Models} and recommendation~\cite{2025_WWW_ED2_Unleash-LLMs-Potential-for-Sequential-Recommendation-by-Coordinating-Dual-Dynamic-Index-Mechanism}. The left side of Fig.~\ref{fig:LLM-Tree} illustrates the evolution of such applications across different single modalities over time. Among them, image and recommendation tasks dominate in volume. 
In terms of model choices, LLaMA~\cite{LLaMA-series_website, 2023_arXiv_LLaMA-2=LLaMA-2=Open-Foundation-and-Fine-tuned-Chat-Models, 2024_arXiv_LLaMA-3_The-LLama-3-Herd-of-Models}-based LLMs are most frequently adopted, followed by T5~\cite{T5-and-variants_Huggingface, 2020_JMLR_T5_Exploring-the-Limits-of-Transfer-Learning-with-A-Unified-Text-to-text-Transformer} variants, Qwen~\cite{Qwen-series_Huggingface, 2025_arXiv_Qwen-3_Qwen3-Technical-Report}, PaLM 2~\cite{PaLM-2_link}, GPT-series~\cite{GPT-series_Huggingface} models and so on. 
The key information and open sources of these applications are summarized in Table~\ref{tabs:LLM-SM} in Appendix.

\vspace{0.3em}
\noindent \textbf{(a) Image.}
Discrete tokenization enables LLMs to process visual inputs by converting image features into semantic tokens, supporting visual alignment, generation, and understanding.
Both LQAE~\cite{2023_NeurlPS_LQAE_Language-Quantized-AutoEncoders=Towards-Unsupervised-Text-Image-Alignment} and SPAE~\cite{2023_NeurIPS_SPAE_SPAE=Semantic-Pyramid-AutoEncoder-for-Multimoda-Generation-with-Frozen-LLMs} leverage pretrained LLM vocabularies to discretize visual signals for efficient visual generation.
LQAE~\cite{2023_NeurlPS_LQAE_Language-Quantized-AutoEncoders=Towards-Unsupervised-Text-Image-Alignment} introduces a VQ-VAE~\cite{2017_NeurlPS_VQ-VAE_Neural-discrete-representation-learning}-style tokenizer that maps images to the token space of frozen LLMs, enabling few-shot multimodal tasks via direct token-level interaction,
while SPAE~\cite{2023_NeurIPS_SPAE_SPAE=Semantic-Pyramid-AutoEncoder-for-Multimoda-Generation-with-Frozen-LLMs} extends this idea with a semantic pyramid token structure that generates variable-length lexical tokens, enabling multimodal in-context learning.
In addition,
LlamaGen~\cite{2024_arXiv_LlamaGen_Autoregressive-Model-Beats-Diffusion=Llama-for-Scalable-Image-Generation} applies the vanilla autoregressive model Llama to image generation, achieving high-quality image tokenization with a downsample ratio of 16.
Similarly, 
StrokeNUWA~\cite{2024_ICML_StrokeNUWA_StrokeNUWA=Tokenizing-Strokes-for-Vector-Graphic-Synthesis} proposes a stroke token as a better visual representation, which is semantically rich, LLM-compatible, and highly compressed, enabling efficient vector graphic synthesis through LLMs.
Both V2L Tokenizer~\cite{2024_CVPR_V2L-Tokenizer_Beyond-Text=Frozen-Large-Language-Models-in-Visual-Signal-Comprehension} and V$^{2}$Flow~\cite{2025_arXiv_V2Flow_V2Flow=Unifying-Visual-Tokenization-and-Large-Language-Model-Vocabularies-for-Autoregressive-Image-Generation} adopt LLM vocabularies as visual codebooks, enabling seamless integration with frozen LLMs.
Concretely,
V2L Tokenizer~\cite{2024_CVPR_V2L-Tokenizer_Beyond-Text=Frozen-Large-Language-Models-in-Visual-Signal-Comprehension} introduces a global-local tokenization scheme to support visual understanding and denoising, while V$^{2}$Flow~\cite{2025_arXiv_V2Flow_V2Flow=Unifying-Visual-Tokenization-and-Large-Language-Model-Vocabularies-for-Autoregressive-Image-Generation} incorporates a vocabulary resampler and rectified-flow decoder for high-quality autoregressive generation.

\vspace{0.3em}
\noindent\textbf{(b) Audio.}
In the audio domain, discrete tokenization has been explored to improve speech generation and recognition by mapping acoustic signals to LLM-compatible token sequences.
TWIST~\cite{2023_NeurlPS_TWIST_Textually-Pretrained-Speech-Language-Models} introduces a warm start from the pre-trained LLM to initialize speech language models for speech generation.
To improve stability and control,
SSVC~\cite{2024_arXiv_SSVC_Enhancing-the-Stability-of-LLM-based-Speech-Generation-Systems-through-Self-supervised-Representation} disentangles speaker identity and linguistic content via self-supervised learning and RVQ.
In addition,
JTFS LM~\cite{2024_arXiv_JTFS-LM_Comparing-Discrete-and-Continuous-Space-LLMs-for-Speech-Recognition} systematically compares discrete and continuous speech representations in LLM-based automatic speech recognition, showing that supervised discrete tokens offer robust performance and better alignment.
GLM-4-Voice~\cite{2024_arXiv_GLM-4-Voice_GLM-4-Voice=Towards-Intelligent-and-Human-like-End-to-end-Spoken-Chatbot} further integrates a single-codebook speech tokenizer into a pre-trained LLM backbone, enabling end-to-end speech-text pretraining and spoken dialogue generation.

\vspace{0.3em}
\noindent \textbf{(c) Graph.}
In graph applications, discrete tokenization helps encode structural information into LLM-compatible tokens for integration and reasoning. 
Specifically,
NT-LLM~\cite{2024_arXiv_NT-LLM_NT-LLM=A-Novel-Node-Tokenizer-for-Integrating-Graph-Structure-into-Large-Language-Models} employs graph anchors for node tokenization, which selects anchors via a greedy algorithm and encodes nodes for LLM input based on anchor-based distance.
Similarly,
Dr.E~\cite{2025_AAAI_Dr.E_Multi-View-Empowered-Structural-Graph-Wordification-for-Language-Models} employs a dual-residual VQ-VAE to discretize graphs into tokens aligned with LLM vocabulary, enabling token-level integration of graph-structured data into LLMs through multi-view structural enhancement.

\vspace{0.3em}
\noindent \textbf{(d) Action.}
For action understanding, discrete tokenization has been used to convert motion sequences into structured token inputs for LLMs.
LLM-AR~\cite{2024_CVPR_LLM-AR_LLMs-are-Good-Action-Recognizers} treats LLMs as action recognizers by projecting skeleton sequences into “action sentences” through a linguistic projection process, where the hyperbolic codebook is designed for the tree-like human skeleton representations.

\vspace{0.3em}
\noindent \textbf{(e) Recommendation Systems.}
In this field, discrete tokenization bridges collaborative and semantic signals, enabling LLMs to handle user-item interactions effectively.

\textbf{\textit{(i) Alignment-Based Semantic-Collaborative Tokenization.}}
To unify item representations, several methods~\cite{2024_ICDE_LC-Rec_Adapting-Large-Language-Models-by-Integrating-Collaborative-Semantics-for-Recommendation, 2024_CIKM_LETTER_Learnable-Item-Tokenization-for-Generative-Recommendation, 2024_CIKM_ColaRec_Content-Based-Collaborative-Generation-for-Recommender-Systems} align semantic content with collaborative signals for better tokenization and recommendation quality.
Specifically,
LC-Rec~\cite{2024_ICDE_LC-Rec_Adapting-Large-Language-Models-by-Integrating-Collaborative-Semantics-for-Recommendation} introduces a tree-structured residual quantizer with uniform semantic mapping and leverages fine-tuning tasks to integrate collaborative semantics.
LETTER~\cite{2024_CIKM_LETTER_Learnable-Item-Tokenization-for-Generative-Recommendation} employs three regularizations to fuse hierarchical semantics, collaborative signals, and code diversity into learnable item tokens.
ColaRec~\cite{2024_CIKM_ColaRec_Content-Based-Collaborative-Generation-for-Recommender-Systems} unifies content and interaction signals through an auxiliary indexing task and a contrastive loss for aligned token learning.
To align input tokens with the representation space of LLMs, several methods~\cite{2024_arXiv_STORE_Streamlining-Semantic-Tokenization-and-Generative-Recommendation-with-A-Single-LLM, 2024_arXiv_META-ID_Improving-LLMs-for-Recommendation-with-Out-Of-Vocabulary-Tokens, 2025_AAAI_Semantic-Convergence=Harmonizing-Recommender-Systems-via-Two-Stage-Alignment-and-Behavioral-Semantic-Tokenization} introduce auxiliary tokens or alignment modules to enhance compatibility and reasoning capacity.
STORE~\cite{2024_arXiv_STORE_Streamlining-Semantic-Tokenization-and-Generative-Recommendation-with-A-Single-LLM} unifies semantic tokenization and generative recommendation using a single LLM;
META ID~\cite{2024_arXiv_META-ID_Improving-LLMs-for-Recommendation-with-Out-Of-Vocabulary-Tokens} introduces out-of-vocabulary tokens via meta-path sampling to align user-item interaction information with LLMs;
and 
Semantic Convergence~\cite{2025_AAAI_Semantic-Convergence=Harmonizing-Recommender-Systems-via-Two-Stage-Alignment-and-Behavioral-Semantic-Tokenization} comprises an alignment tokenization module to synchronize item tokens with input semantic space of LLMs, and an alignment task module to fine-tune LLMs.
In addition, 
QARM~\cite{2024_arXiv_QARM_QARM=Quantitative-Alignment-Multi-Modal-Recommendation-at-Kuaishou} discretizes aligned multi-modal representations into trainable code IDs for downstream tasks.

\textbf{\textit{(ii) Learnable ID Tokenization for Generative Recommendation.}}
TokenRec~\cite{2024_arXiv_Tokenrec_Tokenrec=learning-to-tokenize-id-for-llm-based-generative-recommendation} introduces a masked vector-quantized tokenizer to discretize user and item IDs into LLM-compatible tokens, capturing high-order collaborative knowledge for LLMs, and improves efficiency by only updating GNN.
To enhance LLMs' comprehension towards tokens,
ED$^{2}$~\cite{2025_WWW_ED2_Unleash-LLMs-Potential-for-Sequential-Recommendation-by-Coordinating-Dual-Dynamic-Index-Mechanism} introduces a dual dynamic index mechanism and designs a multi-grained token regulator, unifying index generation and recommendation.
Further,
EAGER-LLM~\cite{2025_WWW_EAGER-LLM_EAGER-LLM=Enhancing-Large-Language-Models-as-Recommenders-through-Exogenous-Behavior-Semantic-Integration} integrates endogenous and exogenous behavioral and semantic signals through dual-source knowledge-rich item indices and multiscale alignment reconstruction tasks.
Recent methods~\cite{2024_arXiv_ETEGRec_Generative-Recommender-with-End-to-End-Learnable-Item-Tokenization, 2025_arXiv_UTGRec_Universal-Item-Tokenization-for-Transferable-Generative-Recommendation} unify tokenization and generation to enhance alignment and transferability.
Specifically,
ETEGRec~\cite{2024_arXiv_ETEGRec_Generative-Recommender-with-End-to-End-Learnable-Item-Tokenization} adopts a dual encoder-decoder architecture to jointly optimize item tokenization and autoregressive recommendation,
while UTGRec~\cite{2025_arXiv_UTGRec_Universal-Item-Tokenization-for-Transferable-Generative-Recommendation} learns a universal item tokenizer across domains using a multimodal LLM and tree-structured codebooks for transferable generation.

\section{LLMs with Multiple Modalities}
\label{Sec_LLM-based-Multi-Modality-Applications}

As LLMs evolve into general-purpose agents, discrete tokenization makes it possible for LLMs to operate in multimodal contexts where modality-specific tokenizers can convert continuous signals to unified token sequences for LLM-based modeling. This section reviews multimodal applications that demand more sophisticated alignment and integration for semantic consistency compared to single-modality applications.
As shown in Fig.~\ref{fig:LLM-Tree}, early exploration began in 2023, followed by rapid expansion in 2024 across increasingly complex modality combinations. Among them, \textit{Text + Image} has seen the most active development, followed by the integration of \textit{Text + Audio}. 
Numerous LLM backbones have been developed, including LLaMA series~\cite{LLaMA-series_website, 2023_arXiv_LLaMA-2=LLaMA-2=Open-Foundation-and-Fine-tuned-Chat-Models, 2024_arXiv_LLaMA-3_The-LLama-3-Herd-of-Models}, T5~\cite{T5-and-variants_Huggingface, 2020_JMLR_T5_Exploring-the-Limits-of-Transfer-Learning-with-A-Unified-Text-to-text-Transformer}, Qwen~\cite{Qwen-series_Huggingface, 2025_arXiv_Qwen-3_Qwen3-Technical-Report}, DeepSeek~\cite{DeepSeek_Huggingface, 2024_arXiv_DeepSeek_DeepSeek-LLM=Scaling-Open-source-Language-Models-with-Longtermism}, Mistral~\cite{Mistral_Huggingface}, Vicuna~\cite{Vicuna-series_Huggingface}, PaLM~\cite{2023_JMLR_PaLM_PaLM=Scaling-Language-Modeling-with-Pathway}, and InternVL~\cite{InternVL_Huggingface, 2024_CVPR_InternVL_InternVL=Scaling-UP-Vision-Foundation-Models-and-Aligning-for-Generic-Visual-linguistic-Tasks}.
The key information and source of these applications are summarized in Table~\ref{tabs:LLM-MM} in Appendix.

\vspace{0.3em}
\noindent\textbf{(a) Text + Image.}
Text and image constitute the most common and extensively explored modality pair in multimodal learning. A key to empowering language models with visual capabilities is to integrate image inputs using the native modeling paradigm of LLMs. Recent studies approach this by discretizing visual signals for unified modeling.

\textbf{\textit{(i) Visual Tokenization for Multimodal Alignment.}}
For alignment with left-to-right autoregressive modeling in LLMs, SEED~\cite{2023_arXiv_SEED_Planting-A-Seed-of-Vision-in-Large-Language-Model} and SEED-LLaMA~\cite{2024_ICLR_SEED-LLaMA_Making-LLaMA-SEE-and-Draw-with-SEED-Tokenizer} generate 1D causal visual tokens by vector quantization, rather than conventional 2D representations.
LaVIT~\cite{2024_ICLR_LaVIT_Unified-Language-vision-Pretraining-in-LLM-with-Dynamic-Discrete-Visual-Tokenization} argues images should be tokenized into discrete tokens to enable LLMs to process images and text indiscriminately, and develops a dynamic variable-length tokenizer for images.
Besides, many studies have focused on aligning visual tokens with language semantics through dedicated tokenizer design, such as the discrete tokenizer with semantic constraints in MUSE-VL~\cite{2024_arXiv_MUSE-VL_MUSE-VL=Modeling-Unified-VLM-through-Semantic-Discrete-Encoding}.
Designed as an early-fusion architecture,
Chameleon~\cite{2024_arXiv_Chameleon_Chameleon=Mixed-modal-Early-fusion-Foundation-Models} can generate interleaved textual and image contents by training mixed-modal discrete tokens.
Also,
ClawMachine~\cite{2024_arXiv_2025_ICLR_ClawMachine_ClawMachine=Fetching-Visual-Tokens-as-An-Entity-for-Referring-and-Grounding} directly embeds discrete visual tokens into text for referential tasks, unifying visual referring and grounding without extra syntax.
Recently, 
QLIP~\cite{2025_arXiv_QLIP_QLIP=Text-Aligned-Visual-Tokenization-Unifies-Auto-Regressive-Multimodal-Understanding-and-Generation} introduces a BSQ~\cite{2024_arXiv_2025_ICLR_BSQ_Image-and-Video-Tokenization-with-Binary-Spherical-Quantization}-based visual tokenizer aligned with text by contrastive and reconstruction learning.
Beyond modality alignment and unification, some studies~\cite{2024_arXiv_Janus_Janus=Decoupling-Visual-Encoding-for-Unified-Multimodal-Understanding-and-Generation, 2025_arXiv_Janus-pro_Janus-pro=Unified-multimodal-understanding-and-generation-with-data-and-model-scaling, 2025_arXiv_UniToken_UniToken=Harmonizing-Multimodal-Understanding-and-Generation-Through-Unified-Visual-Encoding, 2024_arXiv_2025_CVPR_TokenFLow_TokenFlow=Unified-Image_Tokenizer-for-Multimodal} have also considered the gap of information granularities between generation and understanding.
For instance, Janus series~\cite{2024_arXiv_Janus_Janus=Decoupling-Visual-Encoding-for-Unified-Multimodal-Understanding-and-Generation, 2025_arXiv_Janus-pro_Janus-pro=Unified-multimodal-understanding-and-generation-with-data-and-model-scaling} explores decoupled encoding pathways for understanding and generation, where Janus-Pro~\cite{2025_arXiv_Janus-pro_Janus-pro=Unified-multimodal-understanding-and-generation-with-data-and-model-scaling} further scales Janus~\cite{2024_arXiv_Janus_Janus=Decoupling-Visual-Encoding-for-Unified-Multimodal-Understanding-and-Generation} to a larger model and data size.  
In~\cite{2025_arXiv_UniToken_UniToken=Harmonizing-Multimodal-Understanding-and-Generation-Through-Unified-Visual-Encoding}, UniToken also combines VQ-based discrete tokens with continuous features via unified visual encoding.
In addition, TokenFlow~\cite{2024_arXiv_2025_CVPR_TokenFLow_TokenFlow=Unified-Image_Tokenizer-for-Multimodal} decouples semantic and pixel representations through a dual-codebook design and aligns them by shared mapping, unifying understanding and generation.

\textbf{\textit{(ii) Generative Pretraining and Tokenizer Tuning.}}
To improve the synergy between discrete visual tokens and LLMs, recent efforts focus on generative pretraining~\cite{2024_arXiv_Lumina-mGPT_Lumina-mGPT=Illuminate-Flexible-Photorealistic-Text-to-Image-Generation-with-Multimodal-Generative-Pretraining} and tokenizer-level optimization~\cite{2025_arXiv_ETT_End-to-End-Vision-Tokenizer-Tuning}.
Lumina-mGPT~\cite{2024_arXiv_Lumina-mGPT_Lumina-mGPT=Illuminate-Flexible-Photorealistic-Text-to-Image-Generation-with-Multimodal-Generative-Pretraining} advances multimodal generalist capabilities through unambiguous image representation with flexible supervised finetuning strategies.
In addition, ETT~\cite{2025_arXiv_ETT_End-to-End-Vision-Tokenizer-Tuning} jointly trains the vision tokenizer and LLM by feeding codebook embeddings and applying token-level caption supervision.

\textbf{\textit{(iii) Diffusion-Enhanced Vision Decoding.}}
Show-o~\cite{2024_arXiv_2025_ICLR_Show-o_Show-o=One-Single-Transformer-to-Unify-Multimodal-Understanding-and-Generation} and MARS~\cite{2025_AAAI_MARS_MARS=Mixture-of-Auto-regressive-Models-for-Fine-grained-Text-to-image-Synthesis} both adopt autoregressive generation frameworks.
Show-o~\cite{2024_arXiv_2025_ICLR_Show-o_Show-o=One-Single-Transformer-to-Unify-Multimodal-Understanding-and-Generation} uses a single transformer with autoregressive language modeling and discrete diffusion-based image generation,
while MARS~\cite{2025_AAAI_MARS_MARS=Mixture-of-Auto-regressive-Models-for-Fine-grained-Text-to-image-Synthesis} integrates frozen LLMs with trainable visual experts via SemVIE for fine-grained text-to-image generation.
The ILLUME series~\cite{2024_arXiv_ILLUME_ILLUME=Illuminating-Your-LLMs-to-See-Draw-and-Self-enhance, 2025_arXiv_ILLUME+_ILLUME+=Illuminating-Unified-MLLM-with-Dual-Visual-Tokenization-and-Diffusion-Refinement} combines semantic tokenization with diffusion decoding.
Specifically, ILLUME~\cite{2024_arXiv_ILLUME_ILLUME=Illuminating-Your-LLMs-to-See-Draw-and-Self-enhance} introduces a vision tokenizer to enable LLM-based understanding, generation, and self-enhancement,
while ILLUME+~\cite{2025_arXiv_ILLUME+_ILLUME+=Illuminating-Unified-MLLM-with-Dual-Visual-Tokenization-and-Diffusion-Refinement} extends it with a dual-branch tokenizer (DualViTok) and a diffusion decoder for high-fidelity image synthesis and editing.
In addition,
DDT-LLaMA~\cite{2025_arXiv_DDT-LLaMA_Generative-Multimodal-Pretraining-with-Discrete-Diffusion-Timestep-Tokens} introduces discrete diffusion timestep tokens with a recursive structure to enhance visual representation in multimodal generation.
In parallel,
Token-Shuffle~\cite{2025_arXiv_Token-Shuffle_Token-Shuffle=Towards-High-resolution-Image-Generation-with-Autoregressive-Models} designs a plug-and-play spatial token reordering strategy that enhances high-resolution autoregressive generation.

\textbf{\textit{(iv) Advanced Tokenizer Architectures and Integration.}}
Morph-Tokens~\cite{2024_ICML_Morph-Tokens_Auto-encoding-Morph-tokens-for-Multimodal-LLM} and Libra~\cite{2024_ICML_Libra_Libra=Building-Decoupled-Vision-System-on-Large-Language-Models} both decouple visual processing from MLLMs. Morph-Tokens~\cite{2024_ICML_Morph-Tokens_Auto-encoding-Morph-tokens-for-Multimodal-LLM} separates abstract prompts and visual tokens for task-specific comprehension and generation, while Libra~\cite{2024_ICML_Libra_Libra=Building-Decoupled-Vision-System-on-Large-Language-Models} routes inputs through expert modules and cross-modal bridges for discrete autoregressive modeling.
FashionM3~\cite{2025_arXiv_FashionM3_FashionM3=Multimodal-Multitask-and-Multiround-FashionAssistant-based-on-Unified-Vision-Language-Model} fine-tunes the Show-o~\cite{2024_arXiv_2025_ICLR_Show-o_Show-o=One-Single-Transformer-to-Unify-Multimodal-Understanding-and-Generation} model on discrete visual tokens derived from MAGVIT-v2~\cite{2024_ICLR_MAGVIT-v2_Language-Model-Beats-Diffusion-Tokenizer-is-Key-to-Visual-Generation} to support fashion-specific multimodal recommendation and image generation.
HimTok~\cite{2025_arXiv_HiMTok_HiMTok=Learning-Hierarchical-Mask-Tokens-for-Image-Segmentation-with-Large-Multimodal-Model} equips an LLM with hierarchical discrete mask tokens based on TiTok tokenizer~\cite{2024_NeurlPS_TiTok_An-image-is-worth-32-tokens-for-reconstruction-and-generation}, enabling coarse-to-fine segmentation without relying on external decoders.
Both SemHiTok~\cite{2025_arXiv_SemHiTok_SemHiTok=A-unified-Image-Tokenizer-via-Semantic-guided-Hierarchical-Codebook-for-Multimodal-Understanding-and-Generation} and Unicode$^2$~\cite{2025_arXiv_Unicode2_Unicode2=Cascaded-Large-scale-Codebooks-for-Unified-Multimodal-Understanding-and-Generation} adopt hierarchical codebook designs to improve visual tokenization. SemHiTok~\cite{2025_arXiv_SemHiTok_SemHiTok=A-unified-Image-Tokenizer-via-Semantic-guided-Hierarchical-Codebook-for-Multimodal-Understanding-and-Generation} employs semantic guidance to structure the hierarchy for better language alignment, while Unicode$^2$~\cite{2025_arXiv_Unicode2_Unicode2=Cascaded-Large-scale-Codebooks-for-Unified-Multimodal-Understanding-and-Generation} constructs a cascaded 500K-entry codebook to enhance stability.

\vspace{0.3em}
\noindent\textbf{(b) Text + Audio.}
In text-audio applications, discrete tokenization enables LLMs to jointly model speech and language for speech recognition, synthesis, and dialogue.

\textbf{\textit{(i) Discrete Speech Tokenization for Understanding and Generation.}}
For instance, DiscreteSLU~\cite{2024_Interspeech_DiscreteSLU_DiscreteSLU=A-Large-Language-Model-with-Self-supervised-Discrete-Speech-Units-for-Spoken-Language-Understanding} explores applications of spoken language understanding in LLMs by discrete speech units and a speech adapter.
MSRT~\cite{2024_IEEE-Signal-Processing-Letters_MSRT_Tuning-Large-Language-Model-for-Speech-Recognition-with-Mixed-scale-Re-tokenization} introduces a mixed-scale re-tokenization layer, enabling better alignment of multi-granularity speech information with language model inputs for speech recognition.
The CosyVoice series~\cite{2024_arXiv_CosyVoice_CosyVoice=A-Scalable-Multilingual-Zero-shot-Text-to-speech-Synthesizer-Based-on-Supervised-Semantic-Tokens, 2024_arXiv_CosyVoice-2_CosyVoice-2=Scalable-Streaming-Speech-Synthesis-with-Large-Language-Models} improves TTS scalability and expressivity by incorporating multilingual supervision and streaming generation. CosyVoice~\cite{2024_arXiv_CosyVoice_CosyVoice=A-Scalable-Multilingual-Zero-shot-Text-to-speech-Synthesizer-Based-on-Supervised-Semantic-Tokens} leverages supervised speech tokens for multilingual zero-shot synthesis,
while CosyVoice~2~\cite{2024_arXiv_CosyVoice-2_CosyVoice-2=Scalable-Streaming-Speech-Synthesis-with-Large-Language-Models} incorporates streaming techniques for emotional and expressive control.
T5-TTS~\cite{2024_Interspeech_T5TTS_Improving-Robustness-of-LLM-based-Speech-Synthesis-by-Learning-Monotonic-Alignment} exploits attention priors and CTC-based alignment loss with a T5~\cite{2020_JMLR_T5_Exploring-the-Limits-of-Transfer-Learning-with-A-Unified-Text-to-text-Transformer} architecture and spectral codec~\cite{2024_arXiv_Spectral-Codecs=Spectrogram-Based-Audio-Codecs-for-High-Quality-Speech-Synthesis} tokenizer for monotonic alignment between modalities, improving the robustness of TTS.
Similarly, GPT-Talker~\cite{2024_MM_GPT-Talker_Generative-Expressive-Conversational-Speech-Synthesis} introduces semantic and style tokens derived from multimodal dialogue contexts for expressive speech.
For attribute controllability of zero-shot TTS, Spark-TTS~\cite{2025_arXiv_Spark-TTS_Spark-TTS=An-Efficient-LLM-based-Text-to-speech-Model-with-Single-stream-Decoupled-Speech-Tokens} introduces attribute labels and fine-grained attributes and generates tokens by the CoT.

\textbf{\textit{(ii) Real-Time and Dialog-Oriented Speech Modeling.}}
The SpeechGPT series~\cite{2023_EMNLP_SpeechGPT_SpeechGPT=Empowering-Large-Language-Models-with-Intrinsic-Cross-modal-Conversational-Abilities, 2024_arXiv_SpeechGPT-Gen_Speechgpt-gen=Scaling-chain-of-information-speech-generation} supports real-time dialogue through multi-stage training and semantic-perceptual disentanglement. SpeechGPT~\cite{2023_EMNLP_SpeechGPT_SpeechGPT=Empowering-Large-Language-Models-with-Intrinsic-Cross-modal-Conversational-Abilities} adopts a three-stage strategy for cross-modal transfer, while SpeechGPT-Gen~\cite{2024_arXiv_SpeechGPT-Gen_Speechgpt-gen=Scaling-chain-of-information-speech-generation} introduces chain-of-information generation for efficient and expressive speech synthesis.
To reduce latency and computational overhead, IntrinsicVoice~\cite{2024_arXiv_IntrinsicVoice_IntrinsicVoice=Empowering-LLMs-with-Intrinsic-Real-time-Voice-Interaction-Abilities} innovatively reduces the lengths of speech token sequences, and thereby lessens the differences between modalities.
To support full-duplex dialogue~\cite{2024_arXiv_Moshi_Moshi=A-Speech-text-Foundation-Model-for-Real-time-Dialogue, 2024_arXiv_OmniFlatten_OmniFlatten=An-End-to-end-GPT-Model-for-Seamless-Voice-Conversation}, Moshi~\cite{2024_arXiv_Moshi_Moshi=A-Speech-text-Foundation-Model-for-Real-time-Dialogue} generates semantic and acoustic tokens in a streaming and hierarchical manner, while OmniFlatten~\cite{2024_arXiv_OmniFlatten_OmniFlatten=An-End-to-end-GPT-Model-for-Seamless-Voice-Conversation} chunks and flattens speech and text tokens into a single sequence, followed by multi-stage post-training for half- and full-duplex abilities.

\textbf{\textit{(iii) Unified Speech-Language Foundation Models.}}
AudioPaLM~\cite{2023_arXiv_AudioPaLM_AudioPaLM=A-Large-Language-Model-That-Can-Speak-and-Listen} and VoxtLM~\cite{2024_ICASSP_VoxtLM_VoxtLM=Unified-Decoder-Only-Models-for-Consolidating-Speech-Recognition-Synthesis-and-Speech-Text-Continuation-Tasks} extend LLMs with discrete audio tokens and unified vocabularies to support multitask speech-language modeling, including ASR, TTS, and speech-text continuation.
Several models, such as LauraGPT~\cite{2023_arXiv_LauraGPT_LauraGPT=Listen-Attend-Understand-and-Regenerate-Audio-with-GPT} and Kimi-Audio~\cite{2025_arXiv_Kimi-Audio_Kimi-Audio-Technical-Report}, integrate discrete audio tokens with continuous representations for audio understanding, generation, recognition, and conversation.

\vspace{0.3em}
\noindent\textbf{(c) Text + Video.}
In text-video applications, discrete tokenization bridges language and visual dynamics, enabling LLMs to generate or understand videos through unified or hierarchical token sequences.
Loong~\cite{2024_arXiv_Loong_Loong=Generating-Minute-level-Long-Videos-with-Autoregressive-Language-Models} unifies text and video tokens into a single autoregressive sequence and introduces progressive short-to-long training with re-weighted loss and token re-encoding mechanisms, generating coherent minute-level videos.
To support efficient multimodal understanding and generation,
Video-LaVIT~\cite{2024_ICML_Video-LaVIT_Video-LaVIT=Unified-Video-Language-Pre-training-with-Decoupled-Visual-motional-Tokenization} presents a unified video-language pre-training framework that decouples visual and motion information through discrete tokenization.
Building on hierarchical modeling,
HiTVideo~\cite{2025_arXiv_HiTVideo_HiTVideo=Hierarchical-Tokenizers-for-Enhancing-Text-to-Video-Generation-with-Autoregressive-Large-Language-Models} encodes videos into multi-layer discrete tokens to balance compression and reconstruction, enabling efficient text-to-video generation.

\vspace{0.3em}
\noindent\textbf{(d) Text + Graph.}
Recent methods have extended discrete tokenization to specialized domains through domain-specific adaptations.
For molecular modeling, UniMoT~\cite{2024_arXiv_UniMoT_UniMoT=Unified-Molecule-text-Language-Model-with-Discrete-Token-Representation} introduces a unified molecule-text language model that leverages vector quantization to discretize molecular representations, enabling joint sequence modeling and cross-modal generation in a shared token space.
In addition, HIGHT~\cite{2024_ICML-Workshop-HIGHT_Improving-Graph-language-Alignment-with-Hierarchical-Graph-Tokenization} presents hierarchical graph tokenization with node-, motif-, and graph-level tokens to capture multi-scale structural semantics for graph-language alignment.
For electronic health record tasks, MedTok~\cite{2025_arXiv_MedTok_Multimodal-Medical-Code-Tokenizer} proposes a discrete tokenization framework for medical codes by integrating textual descriptions with graph-based relational contexts, supporting multimodal representation learning.
To seamlessly integrate with LLMs for knowledge-aware reasoning, SSQR~\cite{2025_arXiv_SSQR_Self-supervised-Quantized-Representation-for-Seamlessly-Integrating-Knowledge-Graphs-with-Large-Language-Models} develops a self-supervised quantization approach to encode knowledge graphs into discrete tokens.

\vspace{0.3em}
\noindent\textbf{(e) Text + Motion.}
Text-motion applications leverage discrete tokenization to map linguistic instructions to structured motion representations, supporting generation and control across diverse embodiments and tasks.
MotionGlot~\cite{2024_arXiv_MotionGlot_MotionGlot=A-Multi-Embodied-Motion-Generation-Model} introduces a unified Transformer decoder that generates discrete motion tokens for diverse embodiments (e.g., humans, quadrupeds) using embodiment-specific VQ-VAEs and instruction-tuned text prompts.
In~\cite{2024_CVPR_AvatarGPT_AvatarGPT=All-in-one-Framework-for-Motion-Understanding-Planning-Generation-and-Beyond}, AvatarGPT uses VQ-VAE-based motion tokenization and integrates motion tokens into an LLM to unify understanding, planning, and generation tasks through instruction tuning.
SemGrasp~\cite{2024_ECCV_SemGrasp_SemGrasp=Semantic-Grasp-Generation-via-Language-Aligned-Discretization} decomposes grasp generation into a three-level (i.e., orientation, manner, and refinement) token prediction task, using hierarchical VQ-VAE-based discretization aligned with language and point cloud inputs.
Recently, Walk-the-Talk~\cite{2024_IV_Walk-the-Talk_Walk-the-Talk=LLM-Driven-Pedestrian-Motion-Generation} employs VQ-VAE to discretize pedestrian motion and leverages LLMs to generate diverse and realistic behaviors from natural language descriptions.

\begin{figure}[t]
    \centering
    \includegraphics[width=0.86\linewidth]{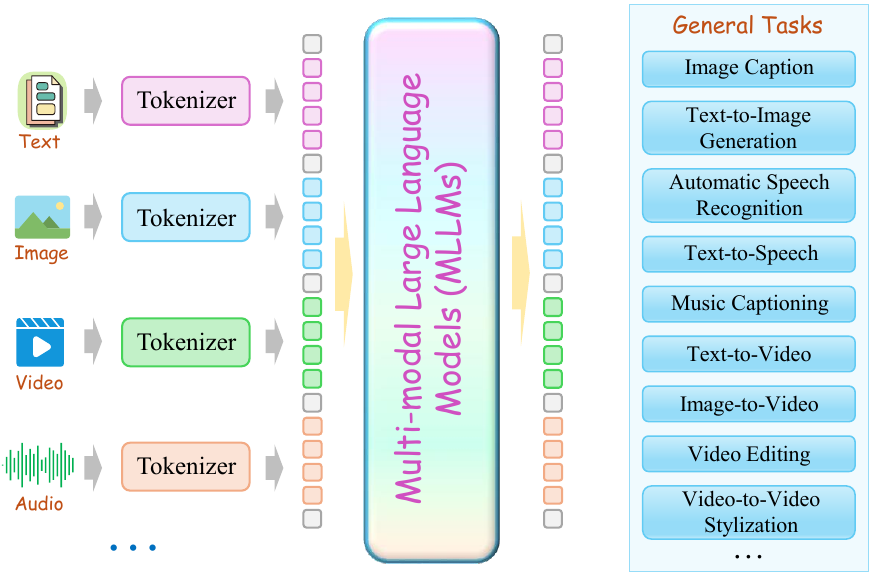}
    \caption{MLLMs process multi-modal inputs by tokenizing text, image, video, and audio into a unified space for diverse tasks~\cite{2024_ICML_VideoPoet_VideoPoet=A-Large-Language-Model-for-Zero-shot-Video-Generation, 2024_arXiv_MIO_MIO=A-foundation-model-on-multimodal-tokens}.}
    \label{fig:MLLM_MM}
\end{figure}

\vspace{0.3em}
\noindent\textbf{(f) Text + Image + Audio.}
Some approaches have explored the multimodal modeling abilities of frozen language models by modality-specific discrete tokenizers, demonstrating the effectiveness of discrete representations in unified multimodal processing.
For instance, TEAL~\cite{2023_arXiv_TEAL_TEAL=Tokenize-and-Embed-All-for-Multi-modal-Large-Language-Models} enables frozen LLMs to process multi-modal data by leveraging VQGAN and Whisper-based tokenizers.
AnyGPT~\cite{2024_ACL_AnyGPT_AnyGPT=Unified-Multimodal-LLM-with-Discrete-Sequence-Modeling} extends this idea by employing SpeechTokenizer~\citep{2024_ICLR_SpeechTokenizer_SpeechTokenizer=Unified-Speech-Tokenizer-for-Speech-Language-Models}, Encodec~\cite{2023_TMLR_EnCodec_High-Fidelity-Neural-Audio-Compression} and SEED~\citep{2023_arXiv_SEED_Planting-A-Seed-of-Vision-in-Large-Language-Model} tokenizers for speech, music, and vision, respectively, achieving unified discrete sequence modeling.
Furthermore, DMLM~\cite{2024_arXiv_DMLM_Discrete-Multimodal-Transformers-with-a-Pretrained-Large-Language-Model-for-Mixed-supervision-Speech-Processing} innovatively normalizes the sequence lengths across modalities and designs mixed supervised and unsupervised training for speech-centric tasks.

\vspace{0.3em}
\noindent\textbf{(g) Text + Image + Video.}
Most visual models still rely on diffusion-based approaches and adopt separate modules for understanding and generation, resulting in suboptimal alignment between perception and generation~\cite{2024_arXiv_Emu3_Emu3=Next-token-Prediction-is-All-You-Need, 2024_arXiv_2025_ICLR_VILA-U_VILA-U=A-Unified-Foundation-Model-Integrating-Visual-Understanding-and-Generation}.
Emu3~\cite{2024_arXiv_Emu3_Emu3=Next-token-Prediction-is-All-You-Need} attempts to eliminate this need for diffusion or compositional architectures by processing all modalities uniformly under the next-token prediction paradigm in a discrete space.
Based on the same paradigm, VILA-U~\cite{2024_arXiv_2025_ICLR_VILA-U_VILA-U=A-Unified-Foundation-Model-Integrating-Visual-Understanding-and-Generation} also aligns visual tokens with textual inputs by contrastive learning.
Additionally, LWM~\cite{2025_ICLR_LWM_World-Model-on-Million-length-Video-and-Language-with-Blockwise-Ringattention} extends the scalability of such models to long video modeling by introducing Blockwise RingAttention, supporting sequences exceeding one million tokens with efficient memory and compute optimization.

\vspace{0.3em}
\noindent\textbf{(h) Text + Audio + Motion.}
Cross-modal tasks can be formulated as sequence-to-sequence translation by discrete tokenization, enabling flexible any-to-any generation and allowing models to leverage the strengths of autoregressive sequence modeling.
Building on this idea, LLM Gesticulator~\cite{2025_ICCGV_LLM-Gesticulator_LLM-Gesticulator=Leveraging-Large-Language-Models-for-Scalable-and-Controllable-Co-speech-Gesture-Synthesis} generates rhythmically aligned and editable full-body co-speech gestures from audio signals with the aid of residual vector quantization, demonstrating scalability and controllability in gesture synthesis.

\vspace{0.3em}
\noindent\textbf{(i) Text + Image + Audio + Video.}
Building upon the foundation of discrete tokenization across modalities, some models extend support to audio and video, forming fully multimodal generative systems. As illustrated in Fig.~\ref{fig:MLLM_MM}, such models unify diverse modality streams via shared token spaces to support both cross-modal understanding and generation. 
\citet{2024_ICML_VideoPoet_VideoPoet=A-Large-Language-Model-for-Zero-shot-Video-Generation} propose VideoPoet, introducing additional audio modality for video generation, encoding visual and audio signals into the discrete space by MAGVIT-v2~\cite{2024_ICLR_MAGVIT-v2_Language-Model-Beats-Diffusion-Tokenizer-is-Key-to-Visual-Generation} and SoundStream~\cite{2022_TASLP_SoundStream_SoundStream=An-End-to-End-Neural-Audio-Codec} tokenizers.
To support multimodal interleaved sequence generation on discrete tokens, MIO~\cite{2024_arXiv_MIO_MIO=A-foundation-model-on-multimodal-tokens} introduces alignment pre-training, interleaved pre-training and speech-enhanced pre-training followed by supervised fine-tuning for multimodal foundation models.

\vspace{0.3em}
\noindent\textbf{(j) Text + Image + Audio + Action.}
Beyond incorporating diverse modalities into a shared space, achieving stable training across multiple modalities, especially stable training from scratch, has also attracted increasing attention.
Unified-IO 2~\cite{2024_CVPR_Unified-IO-2_Unified-IO-2=Scaling-Autoregressive-Multimodal-Models-with-Vision-Language-Audio-and-Action} applies 2D rotary embeddings, QK normalization and scaled cosine attention mechanisms for stability, scaling Unified-IO~\cite{2023_ICLR_Unified-IO_Unified-IO=A-Unified-Model-for-Vision-Language-and-Multi-modal-Tasks} to audio and action modalities under a multimodal mixture of denoisers objective.

\section{Challenges and Future Directions}
\label{Sec_Challenges and Opportunities}

Despite recent progress, discrete tokenization still faces challenges hindering its effectiveness and generalization. In this section, we discuss key challenges and future directions.

\noindent \textbf{(a) Codebook Utilization.}
Underutilized codebooks result in inefficient representations, limiting the expressiveness of discrete tokens. Although techniques like reparameterization tricks~\cite{2023_ICML_Straightening-Out-the-Straight-Through-Estimator=Overcoming-Optimization-Challenges-in-Vector-Quantized-Networks, 2024_arXiv_SimVQ_Addressing-representation-collapse-in-vector-quantized-models-with-one-linear-layer} and diversity regularization~\cite{2023_ICML_VQ-WAE_Vector-Quantized-Wasserstein-Auto-Encoder, 2023_CVPR_Reg-VQ_Regularized-vector-quantization-for-tokenized-image-synthesis, 2024_ICLR_MAGVIT-v2_Language-Model-Beats-Diffusion-Tokenizer-is-Key-to-Visual-Generation} help improve token usage, they often compromise stability. 
Future research could focus on approaches that balance token diversity and coverage with stability, ensuring better codebook utilization without sacrificing performance. For instance, it is worth exploring curriculum-based code activation schedules to promote balanced code usage, and hybrid codebook designs that integrate multiple structural priors (e.g., semantic or spherical organization) to enhance flexibility while maintaining robustness.

\noindent
\textbf{(b) Information Loss.}
Discrete quantization inevitably introduces information reduction~\cite{2024_L4DC_SCQ_Soft-Convex-Quantization=Revisiting-Vector-Quantization-with-Convex-Optimization, 2025_ICLR_DnD-Transformer_A-spark-of-vision-language-intelligence=2-dimensional-autoregressive-transformer-for-efficient-finegrained-image-generation, 2025_CVPR_MergeVQ_MergeVQ=A-Unified-Framework-for-Visual-Generation-and-Representation-with-Disentangled-Token-Merging-and-Quantization, 2023_arXiv_LauraGPT_LauraGPT=Listen-Attend-Understand-and-Regenerate-Audio-with-GPT}, as high-dimensional continuous manifolds are projected onto a finite codebook. When distinct embeddings map to the same code, fine-grained semantic cues may collapse, degrading both discriminative understanding and generative fidelity. This issue is exacerbated in low-codebook settings or under codebook collapse~\cite{2020_NeurlPS_HQA_Hierarchical-quantized-autoencoders,2024_arXv_VQGAN-LC_Scaling-the-codebook-size-of-vqgan-to-100000-with-a-utilization-rate-of-99}, limiting representation completeness compared to raw pixels or dense encoders (e.g., ViT). From a rate-distortion perspective, improved compactness and modularity reduce representational entropy. While it is beneficial for autoregressive modeling, discretization may hinder tasks requiring precise spatial discrimination or high-fidelity reconstruction. Task- and modality-aware capacity allocation, adaptive or hierarchical coding, and hybrid discrete-continuous designs offer promising mitigation strategies.

\noindent
\textbf{(c) Unification of Discrete and Continuous Tokens.}
Discrete and continuous representations exhibit complementary strengths: discrete tokens provide compactness, modular structure, and interpretability, whereas continuous embeddings retain fine-grained information and enable smooth gradient-based optimization~\citep{2024_Preprints_Survey_Continuous-or-Discrete-That-Is-the-Question=A-Survey-on-Large-Multi-modal-Models-from-the-Perspective-of-Input-output-Space-Extension}. 
However, most approaches treat them separately, limiting cross-space synergy. Recently, only a few studies have explored unified frameworks that integrate both representations~\cite{2024_ICLR_LaVIT_Unified-Language-vision-Pretraining-in-LLM-with-Dynamic-Discrete-Visual-Tokenization, 2025_arXiv_TokenBridge_Bridging-Continuous-and-Discrete-Tokens-for-Autoregressive-Visual-Generation, 2025_arXiv_DisCon_Rethinking-Discrete-Tokens=Treating-Them-as-Conditions-for-Continuous-Autoregressive-Image-Synthesis}. 
Notably, many state-of-the-art MLLMs rely primarily on continuous feature alignment to preserve subtle visual distinctions and high-resolution semantics, whereas discrete quantization may struggle with fine-grained discrimination and precise cross-modal alignment due to information loss and limited codebook granularity.
Hybrid architectures that jointly optimize discrete and continuous tokens therefore offer a promising direction, combining expressive precision with structured abstraction.

Additional discussions are provided in the Appendix.

\section{Conclusion}
\label{Sec_Conclusion}

This survey presents a unified overview of discrete tokenization for integrating multimodal data with LLMs. We introduce a taxonomy of VQ methods, analyze their adaptation across modalities, and discuss key challenges and limitations. By synthesizing classical and modern advances, we outline future research directions toward efficient and interpretable multimodal foundation models.

{\footnotesize
\bibliographystyle{plainnat}
\bibliography{Survey}
}

\clearpage

\appendix

\section*{A. Supplement for Classic Applications without LLMs}
\label{appendix_sec-non-LLM}

This section serves as a complementary resource to Section~\ref{Sec_Earlier-Tokenization}. Due to space constraints in the main text, several representative papers on classic applications without large language models are included in this appendix to maintain the completeness of the survey and ensure a comprehensive coverage of the topic.

\subsection*{A.1 Image}
\label{appendix_subsec-non-LLM_Image}

As an extension to Section~\ref{sec_non-LLM_Image}, this section summarizes several representative works on image tokenization.
MoVQ~\cite{2022_NeurlPS_MoVQ_Movq=Modulating-quantized-vectors-for-high-fidelity-image-generation} integrates conditional normalization and multichannel quantization into VQGAN~\cite{2021_CVPR_VQGAN_Taming-transformers-for-high-resolution-image-synthesis} for spatially variant image information.
To unify image generation and representation learning,
MAGE~\cite{2023_CVPR_MAGE_Mage=Masked-generative-encoder-to-unify-representation-learning-and-image-synthesis} uses variable masking ratios,
VQ-KD~\cite{2024_Neurl_VQ-KD_Image-understanding-makes-for-a-good-tokenizer-for-image-generation} distills knowledge from pretrained image understanding encoders (e.g., CLIP),
and VAR~\cite{2024_NeurlPS_VAR_Visual-autoregressive-modeling=Scalable-image-generation-via-next-scale-prediction} proposes a GPT-style autoregressive model using next-scale prediction with multi-scale VQ-VAE quantization.
Additionally, SeQ-GAN~\cite{2024_CVPR_SeQ-GAN_Rethinking-the-objectives-of-vector-quantized-tokenizers-for-image-synthesis} balances semantic compression and detail with perceptual loss and fine-tuning of the decoder.
Recently, MergeVQ~\cite{2025_CVPR_MergeVQ_MergeVQ=A-Unified-Framework-for-Visual-Generation-and-Representation-with-Disentangled-Token-Merging-and-Quantization} introduces token merging in VQ-based generative models, designing a semantically richer tokenizer and boosting image generation quality.
TiTok~\cite{2024_NeurlPS_TiTok_An-image-is-worth-32-tokens-for-reconstruction-and-generation} and FlowMo~\cite{2025_arXiv_FlowMo_Flow-to-the-Mode=Mode-Seeking-Diffusion-Autoencoders-for-State-of-the-Art-Image-Tokenization} are 1D tokenizers, tokenizing images into 1D latent codes. In particular, FlowMo innovatively employs a transformer-based diffusion autoencoder.
MaskBit~\cite{2024_TMLR_MaskBit_MaskBit=Embedding-free-Image-Generation-via-Bit-Tokens} generates bit tokens via LFQ, enabling embedding-free image generation.

\subsection*{A.2 Audio}
\label{appendix_subsec-non-LLM_Audio}

To complement the main discussion in Section~\ref{sec_non-LLM_Audio}, this appendix provides a more detailed overview of several representative methods in audio tokenization, highlighting diverse strategies in codec design, quantization mechanisms, and model architecture choices.
UniCodec~\cite{2025_arXiv_UniCodec_UniCodec=Unified-Audio-Codec-with-Single-Domain-Adaptive-Codebook} proposes a unified audio codec with a single codebook and introduces a partitioned domain-adaptive codebook based on a domain Mixture-of-Experts (MoE) strategy, where different experts are assigned to distinct audio domains while sharing a global quantization space. 
Similarly,
QinCodec~\cite{2025_arXiv_QinCodec_QINCODEC=Neural-Audio-Compression-with-Implicit-Neural-Codebooks} follows a three-stage training pipeline consisting of autoencoder pretraining, offline quantization using QINCo2~\cite{2025_arXiv_QinCo2_Qinco2-Vector-Compression-and-Search-with-Improved-Implicit-Neural-Codebooks}, and decoder fine-tuning, decoupling codebook learning from end-to-end optimization through implicit neural codebooks. 
Meanwhile, 
TAAE~\cite{2025_ICLR_TAAE_Scaling-transformers-for-low-bitrate-high-quality-speech-coding} scales transformer-based encoder–decoder architectures with a flexible discrete bottleneck based on FSQ~\cite{2024_ICLR_FSQ_Finite-scalar-quantization-Vq-vae-made-simple}, enhancing global sequence modeling in low-bitrate speech coding, whereas LFSC~\cite{2025_ICASSP_LFSC_Low-Frame-rate-Speech-Codec=A-Codec-Designed-for-Fast-High-quality-Speech-LLM-Training-and-Inference} leverages finite scalar quantization together with a low frame-rate design and speech-language-model-based discrimination to stabilize codec training under reduced temporal resolution.

\begin{figure*}
    \centering
    \includegraphics[width=0.9\linewidth]{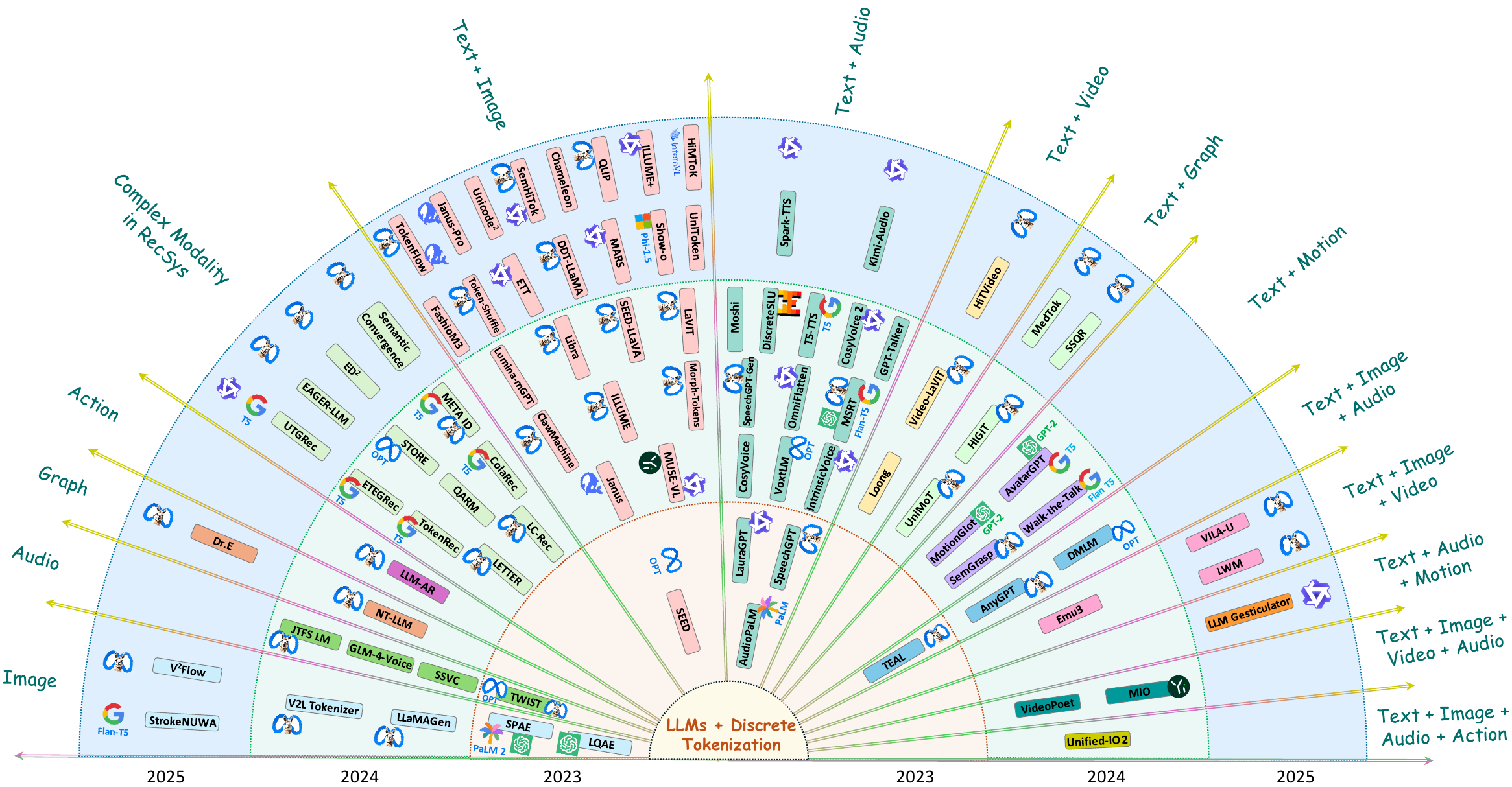}
    \caption{Discrete tokenization with (M)LLMs emerges in 2023 and gains widespread adoption in 2024, especially in LLaMA~\cite{LLaMA-series_website, 2023_arXiv_LLaMA-2=LLaMA-2=Open-Foundation-and-Fine-tuned-Chat-Models, 2024_arXiv_LLaMA-3_The-LLama-3-Herd-of-Models}-based models. The trend continues to accelerate, showing strong momentum in research and applications.}
    \label{fig:LLM-Tree}
\end{figure*}

\begin{table*}[t]
\caption{LLM-based applications with discrete tokenization on single modality.}
\label{tabs:LLM-SM}
\centering
\renewcommand{\arraystretch}{1.6}
\rowcolors{0}{mycolor_tab-1}{mycolor_tab-2}
\scalebox{0.675}{
\begin{tabular}{c c c c c}
\toprule
\textbf{Model} & \textbf{Quantized Modality} & \textbf{VQ Technique} & \textbf{LLM} & \textbf{Open Source} \\
\midrule

SPAE~\cite{2023_NeurIPS_SPAE_SPAE=Semantic-Pyramid-AutoEncoder-for-Multimoda-Generation-with-Frozen-LLMs}            
& Image             
& VQ        
& PaLM 2~\cite{PaLM-2_link}, GPT-3.5~\cite{GPT-series_Huggingface}                           
& - \\ 

LQAE~\cite{2023_NeurlPS_LQAE_Language-Quantized-AutoEncoders=Towards-Unsupervised-Text-Image-Alignment}              
& Image             
& VQ        
& GPT-3~\cite{GPT-series_Huggingface}, InstructGPT~\cite{GPT-series_Huggingface}                        
& \url{https://github.com/haoliuhl/language-quantized-autoencoders} \\ 

V2L Tokenizer~\cite{2024_CVPR_V2L-Tokenizer_Beyond-Text=Frozen-Large-Language-Models-in-Visual-Signal-Comprehension} 
& Image 
& VQ 
& \makecell[c]{LLaMA 2-7B~\cite{LLaMA-series_website}, LLaMA 2-13B~\cite{LLaMA-series_website}, LLaMA 2-70B~\cite{LLaMA-series_website}} & \url{https://github.com/zh460045050/V2L-Tokenizer} \\ 

LlamaGen~\cite{2024_arXiv_LlamaGen_Autoregressive-Model-Beats-Diffusion=Llama-for-Scalable-Image-Generation}        
& Image             
& VQ        
& LLaMA~\cite{LLaMA-series_website} architecture                        
& \url{https://github.com/FoundationVision/LlamaGen} \\ 

V$^2$Flow~\cite{2025_arXiv_V2Flow_V2Flow=Unifying-Visual-Tokenization-and-Large-Language-Model-Vocabularies-for-Autoregressive-Image-Generation}        
& Image             
& VQ        
& LLaMA 2-7B~\cite{LLaMA-series_website}                                 
& \url{https://github.com/zhangguiwei610/V2Flow} \\ 

StrokeNUWA~\cite{2024_ICML_StrokeNUWA_StrokeNUWA=Tokenizing-Strokes-for-Vector-Graphic-Synthesis}      
& Image             
& RVQ       
& Flan-T5 3B~\cite{T5-and-variants_Huggingface}                                
& - \\ 

TWIST~\cite{2023_NeurlPS_TWIST_Textually-Pretrained-Speech-Language-Models}	                           
& Audio	           
& k-means   
& OPT-{125M, 350M, 1.3B}~\cite{OPT-series_Huggingface}, LLaMA-{7B, 13B}~\cite{LLaMA-series_website}        
& \url{https://pages.cs.huji.ac.il/adiyoss-lab/twist/}   \\ 

SSVC~\cite{2024_arXiv_SSVC_Enhancing-the-Stability-of-LLM-based-Speech-Generation-Systems-through-Self-supervised-Representation} 
& Audio             
& RVQ       
& from scratch                              
& - \\ 

JTFS LM~\cite{2024_arXiv_JTFS-LM_Comparing-Discrete-and-Continuous-Space-LLMs-for-Speech-Recognition}         
& Audio             
& k-means   
& LLaMA 2-7B~\cite{LLaMA-series_website}                                 
& \url{https://github.com/xuyaoxun/ASRCompare} \\

GLM-4-Voice~\cite{2024_arXiv_GLM-4-Voice_GLM-4-Voice=Towards-Intelligent-and-Human-like-End-to-end-Spoken-Chatbot}
& Audio 
& VQ 
& GLM-4-9B~\cite{2024_arXiv_GLM-4-Voice_GLM-4-Voice=Towards-Intelligent-and-Human-like-End-to-end-Spoken-Chatbot} 
& \url{https://github.com/zai-org/GLM-4-Voice}\\

NT-LLM~\cite{2024_arXiv_NT-LLM_NT-LLM=A-Novel-Node-Tokenizer-for-Integrating-Graph-Structure-into-Large-Language-Models}          
& Graph             
& GART         
& LLaMA 3-8B~\cite{LLaMA-series_website}             
& - \\ 

Dr.E~\cite{2025_AAAI_Dr.E_Multi-View-Empowered-Structural-Graph-Wordification-for-Language-Models}            
& Graph             
& RVQ                           
& LLaMA 2-7B~\cite{LLaMA-series_website}             
& - \\ 

LLM-AR~\cite{2024_CVPR_LLM-AR_LLMs-are-Good-Action-Recognizers}          
& Action            
& VQ        
& LLaMA-13B~\cite{LLaMA-series_website}             
& - \\ 

LC-Rec~\cite{2024_ICDE_LC-Rec_Adapting-Large-Language-Models-by-Integrating-Collaborative-Semantics-for-Recommendation}          
& Complex modalities in RecSys            
& RVQ       
& LLaMA-7B~\cite{LLaMA-series_website}              
& \url{https://github.com/RUCAIBox/LC-Rec/} \\ 

LETTER~\cite{2024_CIKM_LETTER_Learnable-Item-Tokenization-for-Generative-Recommendation}          
& Complex modalities in RecSys            
& RVQ       
& LLaMA-7B~\cite{LLaMA-series_website}              
& \url{https://github.com/HonghuiBao2000/LETTER} \\ 

ColaRec~\cite{2024_CIKM_ColaRec_Content-Based-Collaborative-Generation-for-Recommender-Systems}         
& Complex modalities in RecSys            
& k-means   
& T5-small~\cite{T5-and-variants_Huggingface} 
& \url{https://github.com/Junewang0614/ColaRec} \\ 

STORE~\cite{2024_arXiv_STORE_Streamlining-Semantic-Tokenization-and-Generative-Recommendation-with-A-Single-LLM}           
& Complex modalities in RecSys            
& k-means   
& OPT-base~\cite{OPT-series_Huggingface}              
& - \\ 

QARM~\cite{2024_arXiv_QARM_QARM=Quantitative-Alignment-Multi-Modal-Recommendation-at-Kuaishou}            
& Complex modalities in RecSys            
& VQ, RVQ   
& 
& - \\ 

META ID~\cite{2024_arXiv_META-ID_Improving-LLMs-for-Recommendation-with-Out-Of-Vocabulary-Tokens}         
& Complex modalities in RecSys            
& k-means   
& T5-small~\cite{T5-and-variants_Huggingface}, LLaMA-7B~\cite{LLaMA-series_website}   
& - \\ 

TokenRec~\cite{2024_arXiv_Tokenrec_Tokenrec=learning-to-tokenize-id-for-llm-based-generative-recommendation}        
& Complex modalities in RecSys            
& VQ        
& T5-small~\cite{T5-and-variants_Huggingface}              
& - \\ 

ETEGRec~\cite{2024_arXiv_ETEGRec_Generative-Recommender-with-End-to-End-Learnable-Item-Tokenization}         
& Complex modalities in RecSys            
& RVQ       
& T5~\cite{T5-and-variants_Huggingface}                    
& - \\ 

Semantic Convergence~\cite{2025_AAAI_Semantic-Convergence=Harmonizing-Recommender-Systems-via-Two-Stage-Alignment-and-Behavioral-Semantic-Tokenization}    
& Complex modalities in RecSys    
& RVQ       
& LLaMA-7B~\cite{LLaMA-series_website}              
& - \\ 

ED$^2$~\cite{2025_WWW_ED2_Unleash-LLMs-Potential-for-Sequential-Recommendation-by-Coordinating-Dual-Dynamic-Index-Mechanism}           
& Complex modalities in RecSys            
& VQ        
& LLaMA 2~\cite{LLaMA-series_website}               
& \url{https://github.com/Esperanto-mega/ED2} \\ 

EAGER-LLM~\cite{2025_WWW_EAGER-LLM_EAGER-LLM=Enhancing-Large-Language-Models-as-Recommenders-through-Exogenous-Behavior-Semantic-Integration}       
& Complex modalities in RecSys            
& k-means   
& LLaMA-7B~\cite{LLaMA-series_website}              
& \url{https://github.com/Indolent-Kawhi/EAGER-LLM} \\ 

UTGRec~\cite{2025_arXiv_UTGRec_Universal-Item-Tokenization-for-Transferable-Generative-Recommendation}          
& Complex modalities in RecSys            
& RVQ       
& Qwen-VL-2B~\cite{Qwen-series_Huggingface}, T5~\cite{T5-and-variants_Huggingface}        
& - \\ 
\bottomrule
\end{tabular}
}
\end{table*}

\begin{table*}[th]
\caption{LLM-based applications with discrete tokenization on multiple modalities.}
\rowcolors{2}{mycolor_tab-1}{mycolor_tab-2}  
\label{tabs:LLM-MM}
\centering
\renewcommand{\arraystretch}{1.38}
\scalebox{0.675}{
\begin{tabular}{c >{\centering\arraybackslash}p{3cm} >{\centering\arraybackslash}p{2cm} c c}
\toprule

\textbf{Model} & \textbf{Quantized Modality} & \textbf{VQ Technique} & \textbf{LLM} & \textbf{Open Source} \\
\midrule

SEED~\cite{2023_arXiv_SEED_Planting-A-Seed-of-Vision-in-Large-Language-Model} 
& Image 
& VQ 
& OPT-2.7B~\cite{OPT-series_Huggingface, 2022_arXiv_OPT_OPT=Open-Pre-trained-Transformer-Language-Models} 
& \url{https://github.com/AILab-CVC/SEED} \\ 

{Chameleon~\cite{2024_arXiv_Chameleon_Chameleon=Mixed-modal-Early-fusion-Foundation-Models}} 
& Image 
& VQ 
& 7B from scratch 
& \url{https://github.com/facebookresearch/chameleon} \\ 

Lumina-mGPT~\cite{2024_arXiv_Lumina-mGPT_Lumina-mGPT=Illuminate-Flexible-Photorealistic-Text-to-Image-Generation-with-Multimodal-Generative-Pretraining}	
& Image	   
& VQ	   
& Chameleon-7B, 30B~\cite{Chameleon-series_Huggingface, 2024_arXiv_Chameleon_Chameleon=Mixed-modal-Early-fusion-Foundation-Models}	    
& \url{https://github.com/Alpha-VLLM/Lumina-mGPT} \\ 

ILLUME~\cite{2024_arXiv_ILLUME_ILLUME=Illuminating-Your-LLMs-to-See-Draw-and-Self-enhance} 
& Image 
& VQ 
& Vicuna-7B~\cite{Vicuna-series_Huggingface} 
& - \\ 

Janus~\cite{2024_arXiv_Janus_Janus=Decoupling-Visual-Encoding-for-Unified-Multimodal-Understanding-and-Generation}  
& Image 
& VQ  
& DeepSeek-LLM (1.3B)~\cite{DeepSeek_Huggingface}      
& \url{https://github.com/deepseek-ai/janus} \\ 

Janus-Pro~\cite{2025_arXiv_Janus-pro_Janus-pro=Unified-multimodal-understanding-and-generation-with-data-and-model-scaling}	    
& Image	    
& VQ	   
& DeepSeek-LLM (1.5B, 7B)~\cite{DeepSeek_Huggingface}	  
& \url{https://github.com/deepseek-ai/janus}    \\ 

MUSE-VL~\cite{2024_arXiv_MUSE-VL_MUSE-VL=Modeling-Unified-VLM-through-Semantic-Discrete-Encoding} 
& Image 
& VQ  
& Qwen2.5-7B~\cite{Qwen-series_Huggingface}, Qwen2.5-32B~\cite{Qwen-series_Huggingface}, Yi-1.5-9B~\cite{Yi-series_Huggingface}, Yi-1.5-34B~\cite{Yi-series_Huggingface} 
& - \\ 

Morph-Tokens~\cite{2024_ICML_Morph-Tokens_Auto-encoding-Morph-tokens-for-Multimodal-LLM} 
& Image 
& VQ  
& Vicuna~\cite{Vicuna-series_Huggingface} 
& \url{https://github.com/DCDmllm/MorphTokens} \\ 

LaVIT~\cite{2024_ICLR_LaVIT_Unified-Language-vision-Pretraining-in-LLM-with-Dynamic-Discrete-Visual-Tokenization} 
& Image 
& VQ 
& LLaMA-7B~\cite{LLaMA-series_website} 
& \url{https://github.com/jy0205/LaVIT} \\ 

SEED-LLaMA~\cite{2024_ICLR_SEED-LLaMA_Making-LLaMA-SEE-and-Draw-with-SEED-Tokenizer} 
& Image 
& VQ  
& Vicuna-7B~\cite{Vicuna-series_Huggingface}, LLaMA 2-13B-Chat~\cite{LLaMA-series_website} 
& \url{https://github.com/AILab-CVC/SEED} \\ 

Libra~\cite{2024_ICML_Libra_Libra=Building-Decoupled-Vision-System-on-Large-Language-Models} 
& Image 
& LFQ 
& LLaMA 2-7B-Chat~\cite{LLaMA-series_website} 
& \url{https://github.com/YifanXu74/Libra} \\ 

Show-o~\cite{2024_arXiv_2025_ICLR_Show-o_Show-o=One-Single-Transformer-to-Unify-Multimodal-Understanding-and-Generation} 
& Image 
& LFQ 
& Phi1.5-1.3B~\cite{Phi-and-variants_Huggingface} 
& \url{https://github.com/showlab/Show-o} \\ 

TokenFlow~\cite{2024_arXiv_2025_CVPR_TokenFLow_TokenFlow=Unified-Image_Tokenizer-for-Multimodal} 
& Image 
& VQ 
& Vicuna-v1.5-13B~\cite{Vicuna-series_Huggingface}, Qwen2.5-14B~\cite{Qwen-series_Huggingface}, LLaMA 2-7B~\cite{LLaMA-series_website} 
& \url{https://bytevisionlab.github.io/TokenFlow/} \\ 

ClawMachine~\cite{2024_arXiv_2025_ICLR_ClawMachine_ClawMachine=Fetching-Visual-Tokens-as-An-Entity-for-Referring-and-Grounding} 
& Image 
& VQ 
& LaVIT-7B~\cite{2024_ICLR_LaVIT_Unified-Language-vision-Pretraining-in-LLM-with-Dynamic-Discrete-Visual-Tokenization} 
& \url{https://github.com/martian422/ClawMachine} \\ 

DDT-LLaMA~\cite{2025_arXiv_DDT-LLaMA_Generative-Multimodal-Pretraining-with-Discrete-Diffusion-Timestep-Tokens} 
& Image 
& VQ  
& LLaMA 3-8B~\cite{LLaMA-series_website} 
& \url{https://ddt-llama.github.io/} \\ 

FashionM3~\cite{2025_arXiv_FashionM3_FashionM3=Multimodal-Multitask-and-Multiround-FashionAssistant-based-on-Unified-Vision-Language-Model} 
& Image 
& LFQ 
& fine-tuning Show-o~\cite{2024_arXiv_2025_ICLR_Show-o_Show-o=One-Single-Transformer-to-Unify-Multimodal-Understanding-and-Generation} 
& - \\ 

HiMTok~\cite{2025_arXiv_HiMTok_HiMTok=Learning-Hierarchical-Mask-Tokens-for-Image-Segmentation-with-Large-Multimodal-Model} 
& Image 
& VQ 
& InternVL-2.5-8B~\cite{InternVL_Huggingface} 
& \url{https://github.com/yayafengzi/LMM-HiMTok} \\ 

ILLUME+~\cite{2025_arXiv_ILLUME+_ILLUME+=Illuminating-Unified-MLLM-with-Dual-Visual-Tokenization-and-Diffusion-Refinement} 
& Image 
& VQ 
& Qwen2.5-3B~\cite{Qwen-series_Huggingface} 
& \url{https://illume-unified-mllm.github.io/} \\ 

QLIP~\cite{2025_arXiv_QLIP_QLIP=Text-Aligned-Visual-Tokenization-Unifies-Auto-Regressive-Multimodal-Understanding-and-Generation} 
& Image 
& BSQ 
& LLaMA 3~\cite{LLaMA-series_website}  
& - \\ 

SemHiTok~\cite{2025_arXiv_SemHiTok_SemHiTok=A-unified-Image-Tokenizer-via-Semantic-guided-Hierarchical-Codebook-for-Multimodal-Understanding-and-Generation} 
& Image 
& VQ 
& Vicuna-v1.5-7B~\cite{Vicuna-series_Huggingface}, Qwen2.5-3B~\cite{Qwen-series_Huggingface} 
& - \\ 

UniToken~\cite{2025_arXiv_UniToken_UniToken=Harmonizing-Multimodal-Understanding-and-Generation-Through-Unified-Visual-Encoding} 
& Image 
& VQ  
& Chameleon 7B~\cite{Chameleon-series_Huggingface, 2024_arXiv_Chameleon_Chameleon=Mixed-modal-Early-fusion-Foundation-Models} 
& \url{https://github.com/SxJyJay/UniToken} \\ 

Token-Shuffle~\cite{2025_arXiv_Token-Shuffle_Token-Shuffle=Towards-High-resolution-Image-Generation-with-Autoregressive-Models} 
& Image 
& VQ  
& LLaMA-2.7B~\cite{LLaMA-series_website} 
& - \\ 

MARS~\cite{2025_AAAI_MARS_MARS=Mixture-of-Auto-regressive-Models-for-Fine-grained-Text-to-image-Synthesis} 
& Image 
& VQ  
& Qwen-7B~\cite{Qwen-series_Huggingface} 
& \url{https://github.com/fusiming3/MARS} \\ 

ETT~\cite{2025_arXiv_ETT_End-to-End-Vision-Tokenizer-Tuning} 
& Image 
& VQ 
& Qwen2.5-1.5B~\cite{Qwen-series_Huggingface} 
& - \\ 

Unicode$^2$~\cite{2025_arXiv_Unicode2_Unicode2=Cascaded-Large-scale-Codebooks-for-Unified-Multimodal-Understanding-and-Generation} 
& Image 
& k-means 
& Qwen2.5-7B-Instruct~\cite{Qwen-series_Huggingface} 
& - \\

AudioPaLM~\cite{2023_arXiv_AudioPaLM_AudioPaLM=A-Large-Language-Model-That-Can-Speak-and-Listen} 
& Audio 
& k-means 
& PaLM-2 8B~\cite{PaLM-2_link} 
& \url{https://google-research.github.io/seanet/audiopalm/examples/} \\ 

LauraGPT~\cite{2023_arXiv_LauraGPT_LauraGPT=Listen-Attend-Understand-and-Regenerate-Audio-with-GPT} 
& Audio 
& RVQ 
& Qwen-1.8B~\cite{Qwen-series_Huggingface} 
& \url{https://lauragpt.github.io/} \\ 

SpeechGPT~\cite{2023_EMNLP_SpeechGPT_SpeechGPT=Empowering-Large-Language-Models-with-Intrinsic-Cross-modal-Conversational-Abilities} 
& Audio 
& k-means 
& LLaMA-13B~\cite{LLaMA-series_website} 
& 
 \url{https://0nutation.github.io/SpeechGPT.github.io/}
\\ 

SpeechGPT-Gen~\cite{2024_arXiv_SpeechGPT-Gen_Speechgpt-gen=Scaling-chain-of-information-speech-generation}    
& Audio     
& RVQ      
& LLaMA 2-7B-Chat~\cite{LLaMA-series_website}         
& \url{https://github.com/0nutation/SpeechGPT}   \\ 

CosyVoice~\cite{2024_arXiv_CosyVoice_CosyVoice=A-Scalable-Multilingual-Zero-shot-Text-to-speech-Synthesizer-Based-on-Supervised-Semantic-Tokens} 
& Audio 
& VQ 
& from scratch  
& 
\begin{tabular}[c]{@{}l@{}}
\url{https://github.com/FunAudioLLM/CosyVoice}
\end{tabular} \\ 

CosyVoice 2~\cite{2024_arXiv_CosyVoice-2_CosyVoice-2=Scalable-Streaming-Speech-Synthesis-with-Large-Language-Models} 
& Audio 
& FSQ 
& Qwen2.5-0.5B~\cite{Qwen-series_Huggingface} 
& \url{https://funaudiollm.github.io/cosyvoice2} \\ 

IntrinsicVoice~\cite{2024_arXiv_IntrinsicVoice_IntrinsicVoice=Empowering-LLMs-with-Intrinsic-Real-time-Voice-Interaction-Abilities} 
& Audio 
& k-means 
& Qwen2-7B-Instruct~\cite{Qwen-series_Huggingface} 
& - \\ 

Moshi~\cite{2024_arXiv_Moshi_Moshi=A-Speech-text-Foundation-Model-for-Real-time-Dialogue} 
& Audio 
& RVQ 
& 7B from scratch 
& \url{https://github.com/kyutai-labs/moshi} \\ 

OmniFlatten~\cite{2024_arXiv_OmniFlatten_OmniFlatten=An-End-to-end-GPT-Model-for-Seamless-Voice-Conversation} 
& Audio 
& VQ 
& Qwen2-0.5B~\cite{Qwen-series_Huggingface} 
& \url{https://omniflatten.github.io/} \\ 

MSRT~\cite{2024_IEEE-Signal-Processing-Letters_MSRT_Tuning-Large-Language-Model-for-Speech-Recognition-with-Mixed-scale-Re-tokenization} 
& Audio 
& k-means 
& Flan-T5~\cite{T5-and-variants_Huggingface}, GPT-2 medium~\cite{GPT-series_Huggingface}, LLaMA 2-7B~\cite{LLaMA-series_website} 
& - \\ 

T5-TTS~\cite{2024_Interspeech_T5TTS_Improving-Robustness-of-LLM-based-Speech-Synthesis-by-Learning-Monotonic-Alignment} 
& Audio 
& FSQ 
& T5~\cite{T5-and-variants_Huggingface} 
& - \\ 

DiscreteSLU~\cite{2024_Interspeech_DiscreteSLU_DiscreteSLU=A-Large-Language-Model-with-Self-supervised-Discrete-Speech-Units-for-Spoken-Language-Understanding} 
& Audio 
& k-means 
& Mistral-7B~\cite{Mistral_Huggingface} 
& - \\ 

GPT-Talker~\cite{2024_MM_GPT-Talker_Generative-Expressive-Conversational-Speech-Synthesis} 
& Audio 
& k-means, VQ 
& from scratch  
& \url{https://github.com/AI-S2-Lab/GPT-Talker} \\ 

VoxtLM~\cite{2024_ICASSP_VoxtLM_VoxtLM=Unified-Decoder-Only-Models-for-Consolidating-Speech-Recognition-Synthesis-and-Speech-Text-Continuation-Tasks}	
& Audio	  
& k-means   
& OPT~\cite{OPT-series_Huggingface, 2022_arXiv_OPT_OPT=Open-Pre-trained-Transformer-Language-Models}	   
& \url{https://soumimaiti.github.io/icassp24_voxtlm/}   \\ 

Spark-TTS~\cite{2025_arXiv_Spark-TTS_Spark-TTS=An-Efficient-LLM-based-Text-to-speech-Model-with-Single-stream-Decoupled-Speech-Tokens} 
& Audio 
& VQ, FSQ 
& Qwen2.5-0.5B-Instruct~\cite{Qwen-series_Huggingface} 
& \url{https://github.com/SparkAudio/Spark-TTS} \\ 

Kimi-Audio~\cite{2025_arXiv_Kimi-Audio_Kimi-Audio-Technical-Report} 
& Audio 
& VQ 
& Qwen2.5-7B~\cite{Qwen-series_Huggingface} 
& \url{https://github.com/MoonshotAI/Kimi-Audio} \\ 

Loong~\cite{2024_arXiv_Loong_Loong=Generating-Minute-level-Long-Videos-with-Autoregressive-Language-Models} 
& Video 
& VQ 
& 700M, 3B, 7B from scratch 
& \url{https://yuqingwang1029.github.io/Loong-video/} \\ 

Video-LaVIT~\cite{2024_ICML_Video-LaVIT_Video-LaVIT=Unified-Video-Language-Pre-training-with-Decoupled-Visual-motional-Tokenization} 
& Video 
& VQ  
& LLaMA 2-7B~\cite{LLaMA-series_website} 
& \url{https://video-lavit.github.io/} \\ 

HiTVideo~\cite{2025_arXiv_HiTVideo_HiTVideo=Hierarchical-Tokenizers-for-Enhancing-Text-to-Video-Generation-with-Autoregressive-Large-Language-Models} 
& Video 
& LFQ 
& LLaMA-3B~\cite{LLaMA-series_website} 
& \url{https://ziqinzhou66.github.io/project/HiTVideo/} \\ 

UniMoT~\cite{2024_arXiv_UniMoT_UniMoT=Unified-Molecule-text-Language-Model-with-Discrete-Token-Representation} 
& Graph 
& VQ  
& LLaMA 2-7B~\cite{LLaMA-series_website} 
& \url{https://uni-mot.github.io/} \\ 

HIGHT~\cite{2024_ICML-Workshop-HIGHT_Improving-Graph-language-Alignment-with-Hierarchical-Graph-Tokenization} 
& Graph 
& VQ  
& Vicuna-v1.3-7B~\cite{Vicuna-series_Huggingface} 
& \url{https://higraphllm.github.io/} \\ 

MedTok~\cite{2025_arXiv_MedTok_Multimodal-Medical-Code-Tokenizer} 
& Graph 
& VQ 
& LLaMA 3.1–8B~\cite{LLaMA-series_website} 
& - \\ 

SSQR~\cite{2025_arXiv_SSQR_Self-supervised-Quantized-Representation-for-Seamlessly-Integrating-Knowledge-Graphs-with-Large-Language-Models} 
& Graph 
& VQ  
& LLaMA 2-7B~\cite{LLaMA-series_website}, LLaMA 3.1-8B~\cite{LLaMA-series_website} 
& - \\ 

MotionGlot~\cite{2024_arXiv_MotionGlot_MotionGlot=A-Multi-Embodied-Motion-Generation-Model} 
& Motion 
& VQ  
& GPT-2 small~\cite{GPT-series_Huggingface} 
& \url{https://ivl.cs.brown.edu/research/motionglot.html} \\ 

AvatarGPT~\cite{2024_CVPR_AvatarGPT_AvatarGPT=All-in-one-Framework-for-Motion-Understanding-Planning-Generation-and-Beyond} 
& Motion 
& VQ  
& GPT-2 large~\cite{GPT-series_Huggingface}, T5-Large~\cite{T5-and-variants_Huggingface} 
& \url{https://zixiangzhou916.github.io/AvatarGPT/} \\ 

Semgrasp~\cite{2024_ECCV_SemGrasp_SemGrasp=Semantic-Grasp-Generation-via-Language-Aligned-Discretization} 
& Motion 
& VQ  
& Vicuna-7B~\cite{Vicuna-series_Huggingface} 
& \url{https://kailinli.github.io/SemGrasp/} \\ 

Walk-the-Talk~\cite{2024_IV_Walk-the-Talk_Walk-the-Talk=LLM-Driven-Pedestrian-Motion-Generation} 
& Motion 
& VQ  
& Flan-T5-Base~\cite{T5-and-variants_Huggingface} 
& \url{https://iv.ee.hm.edu/publications/w-the-t/} \\ 

TEAL~\cite{2023_arXiv_TEAL_TEAL=Tokenize-and-Embed-All-for-Multi-modal-Large-Language-Models} 
& \makecell[c]{Image \\ Audio} 
& \makecell[c]{VQ \\ k-means}  
& LLaMA~\cite{LLaMA-series_website} 
& - \\ 

DMLM~\cite{2024_arXiv_DMLM_Discrete-Multimodal-Transformers-with-a-Pretrained-Large-Language-Model-for-Mixed-supervision-Speech-Processing} 
& \makecell[c]{Image \\ Audio} 
& \makecell[c]{VQ \\ RVQ} 
& {OPT~\cite{OPT-series_Huggingface, 2022_arXiv_OPT_OPT=Open-Pre-trained-Transformer-Language-Models}} 
& - \\ 

AnyGPT~\cite{2024_ACL_AnyGPT_AnyGPT=Unified-Multimodal-LLM-with-Discrete-Sequence-Modeling} 
& \makecell[c]{Image \\ Audio} 
& \makecell[c]{VQ \\ RVQ}  
& LLaMA 2-7B~\cite{LLaMA-series_website} 
& \url{https://junzhan2000.github.io/AnyGPT.github.io/} \\ 

Emu3~\cite{2024_arXiv_Emu3_Emu3=Next-token-Prediction-is-All-You-Need} 
& \makecell[c]{Image \\ Video} 
& \makecell[c]{VQ \\ VQ}  
& 8B from scratch
& \url{https://emu.baai.ac.cn/about} \\ 

VILA-U~\cite{2024_arXiv_2025_ICLR_VILA-U_VILA-U=A-Unified-Foundation-Model-Integrating-Visual-Understanding-and-Generation} 
& \makecell[c]{Image \\ Video} 
& \makecell[c]{RVQ \\ RVQ} 
& LLaMA 2-7B~\cite{LLaMA-series_website} 
& \url{https://github.com/mit-han-lab/vila-u} \\ 

LWM~\cite{2025_ICLR_LWM_World-Model-on-Million-length-Video-and-Language-with-Blockwise-Ringattention} 
& \makecell[c]{Image \\ Video} 
& \makecell[c]{VQ \\ VQ} 
& LLaMA 2-7B~\cite{LLaMA-series_website}
& \url{https://largeworldmodel.github.io/lwm/} \\ 

LLM Gesticulator~\cite{2025_ICCGV_LLM-Gesticulator_LLM-Gesticulator=Leveraging-Large-Language-Models-for-Scalable-and-Controllable-Co-speech-Gesture-Synthesis} 
& \makecell[c]{Audio \\ Motion} 
& \makecell[c]{RVQ \\ RVQ} 
& Qwen1.5-0.5B~\cite{Qwen-series_Huggingface}, Qwen1.5-1.8B~\cite{Qwen-series_Huggingface}, Qwen1.5-4B~\cite{Qwen-series_Huggingface}, Qwen1.5-7B~\cite{Qwen-series_Huggingface}
& - \\ 

VideoPoet~\cite{2024_ICML_VideoPoet_VideoPoet=A-Large-Language-Model-for-Zero-shot-Video-Generation} 
& \makecell[c]{Image \\ Video \\ Audio} 
& \makecell[c]{LFQ \\ LFQ \\ RVQ} 
& 1B, 8B from scratch
& \url{https://sites.research.google/videopoet/} \\ 

MIO~\cite{2024_arXiv_MIO_MIO=A-foundation-model-on-multimodal-tokens}
& \makecell[c]{Image \\ Video \\Audio} 
& \makecell[c]{VQ \\ VQ \\ RVQ} 
&  Yi-6B-Base~\cite{Yi-series_Huggingface} 
& \url{https://github.com/MIO-Team/MIO} \\ 

Unified-IO 2~\cite{2024_CVPR_Unified-IO-2_Unified-IO-2=Scaling-Autoregressive-Multimodal-Models-with-Vision-Language-Audio-and-Action}
& \makecell[c]{Image \\ Audio} 
& \makecell[c]{VQ \\ VQ}  
& 1.1B, 3.2B, 6.8B from scratch
& 
\url{https://unified-io-2.allenai.org/}
 \\ 

\bottomrule
\end{tabular}
}
\end{table*}

\subsection*{A.3 Graph}
\label{appendix_subsec-non-LLM_Graph}

We include here several additional graph tokenization methods that further illustrate the diversity of discrete modeling strategies beyond those discussed in Section~\ref{sec_non-LLM_Graph}.
VQGraph~\cite{2024_ICLR_VQGraph_VQGraph=Rethinking-Graph-Representation-Space-for-Bridging-GNNs-and-MLPs} introduces a structure-aware tokenizer that encodes local substructures into discrete codes for effective GNN-to-MLP distillation.
Along similar lines,
GFT~\cite{2024_NeurlPS_GFT_Graph-Foundation-Model-with-Transferable-Tree-Vocabulary} treats computation trees as a discrete tree vocabulary via tree reconstruction, unifying tasks into tree classification for graph foundation model.
In contrast, 
HQA-GAE~\cite{2025_WWW_HQA-GAE_Hierarchical-Vector-Quantized-Graph-Autoencoder-with-Annealing-Based-Code-Selection} applies VQ-VAE~\cite{2017_NeurlPS_VQ-VAE_Neural-discrete-representation-learning} to graphs with a hierarchical codebook and annealing-based selection to address underutilization and sparsity.
Complementarily,
GQT~\cite{2025_ICLR_GQT_Learning-Graph-Quantized-Tokenizers} leverages multi-task self-supervised learning and RVQ to generate hierarchical graph tokens for efficient, generalizable tokenization.
Aiming for efficiency,
GT-SVQ~\cite{2025_arXiv_GT-SVQ_GT-SVQ=A-Linear-Time-Graph-Transformer-for-Node-Classification-Using-Spiking-Vector-Quantization} builds a linear-time graph transformer using spiking vector quantization, where spike count embeddings act as codewords to guide attention.

\subsection*{A.4 Video}
\label{appendix_subsec-non-LLM_Video}

This section provides additional representative works that extend video tokenization techniques beyond those discussed in Section~\ref{sec_non-LLM_Video}.
Phenaki~\cite{2023_ICLR_Phenaki_Phenaki=Variable-Length-Video-Generation-from-Open-Domain-Textual-Description} proposes a discrete video tokenizer with causal temporal attention to compress variable-length videos into compact token sequences.
OmniTokenizer~\cite{2024_NeurlPS_OmniTokenizer_OmniTokenizer=A-Joint-Image-Video-Tokenizer-for-Visual-Generation} introduces a spatial-temporal transformer and progressive training for joint image and video tokenization.
LARP~\cite{2024_arXiv_LARP_LARP=Tokenizing-Videos-with-A-Learned-Autoregressive-Generative-Prior} introduces a holistic video tokenizer based on stochastic quantization, enhancing generation performance with the autoregressive prior model.
Focusing on improved quantization strategies,
VidTok~\cite{2024_arXiv_VidTok_VidTok=A-Versatile-and-Open-source-Video-Tokenizer} is an open-source video tokenizer that uses FSQ~\cite{2024_ICLR_FSQ_Finite-scalar-quantization-Vq-vae-made-simple} to improve discrete representation and mitigate codebook collapse in VQ methods.
To better capture hierarchical video structure,
VQ-NeRV~\cite{2024_arXiv_VQ-NeRV_VQ-NeRV=A-Vector-Quantized-Neural-Representation-for-Videos} adopts a U-shaped architecture with VQ-based blocks to discretize shallow and inter-frame residual features.
Complementarily,
SweetTok~\cite{2024_arXiv_SweetTokenizer_SweetTokenizer=Semantic-aware-Spatial-temporal-Tokenizer-for-Compact-Visual-Discretization} introduces a semantic-aware tokenizer that captures spatial-temporal cues to produce compact, informative video tokens.
Recently,
TVC~\cite{2025_arXiv_TVC_TCV=Tokenized-Video-Compression-with-Ultra-low-Bitrate} combines discrete and continuous token compression for ultra-low bitrate video reconstruction with high fidelity.

\subsection*{A.5 Recommendation Systems}
\label{appendix_subsec-non-LLM_RecSys}

This section provides additional representative works that extend beyond those discussed in Section~\ref{sec:non-LLM_MM}~(g).
ReFRS~\cite{2023_TOIS_ReFRS_ReFRS=Resource-efficient-Federated-Recommender-System-for-Dynamic-and-Diversified-User-Preferences} emphasizes privacy-preserving federated learning via VQ-VAE-based temporal embeddings and semantic clustering, while VQ-Rec~\cite{2023_WWW_VQ-Rec_Learning-Vector-Quantized-Item-Representation-for-Transferable-Sequential-Recommenders} focuses on transferability through contrastive pretraining and permutation-based OPQ alignment.
TIGER~\cite{2023_NeurlPS_TIGER_Recommender-Systems-with-Generative-Retrieval} and CoST~\cite{2024_RecSys_CoST_CoST=Contrastive-Quantization-based-Semantic-Tokenization-for-Generative-Recommendation} both target generative recommendation via semantic tokenization. Concretely, TIGER~\cite{2023_NeurlPS_TIGER_Recommender-Systems-with-Generative-Retrieval} adopts RQ-VAE~\cite{2022_CVPR_RQ-VAE_RQ-Transformer_Autoregressive-image-generation-using-residual-quantization} to represent items as semantic IDs and autoregressively predicts the next item, while CoST~\cite{2024_RecSys_CoST_CoST=Contrastive-Quantization-based-Semantic-Tokenization-for-Generative-Recommendation} improves token quality through contrastive quantization.
In~\cite{2024_KDD_EAGER_EAGER=Two-Stream-Generative-Recommender-with-Behavior-Semantic-Collaboration}, EAGER integrates behavior and semantic tokens via contrastive learning and semantic-guided transfer in a two-stream framework.

\section*{B. Supplement for LLMs with Single-Modality Applications}

Table~\ref{tabs:LLM-SM} serves as a complementary resource to Section~\ref{Sec_LLM-based-Single-Modality-Applications}, presenting a structured summary of LLM-based applications that leverage discrete tokenization on a single modality. Each entry is categorized by the quantized modality (e.g., image, audio, graph, action, or complex modalities in recommendation), the employed quantization technique (e.g., VQ~\cite{2017_NeurlPS_VQ-VAE_Neural-discrete-representation-learning}, PQ~\citep{2011_TPAMI_PQ_Product-Quantization-for-Nearest-Neighbor-Search}, LFQ~\cite{2024_ICLR_MAGVIT-v2_Language-Model-Beats-Diffusion-Tokenizer-is-Key-to-Visual-Generation}), the backbone LLM, and the availability of open-source implementations. This table provides concrete instances of the methods discussed in the main text, offering a practical reference for how vector quantization techniques have been integrated into single-modality LLM pipelines.

\section*{C. Supplement for LLMs with Multiple-Modality Applications}

Table~\ref{tabs:LLM-MM} serves as a complementary resource to Section~\ref{Sec_LLM-based-Multi-Modality-Applications}, presenting a structured summary of LLM-based applications that leverage discrete tokenization across multiple modalities. Each entry is categorized by the quantized modality (e.g., image, audio, video, motion), the employed quantization technique (e.g., VQ~\cite{2017_NeurlPS_VQ-VAE_Neural-discrete-representation-learning}, k-means, LFQ~\cite{2024_ICLR_MAGVIT-v2_Language-Model-Beats-Diffusion-Tokenizer-is-Key-to-Visual-Generation}), the backbone LLM, and the availability of open-source implementations. This table provides concrete instances of the methods discussed in the main text, offering a practical reference for how vector quantization techniques have been integrated into multimodal LLM pipelines.

\section*{D. Supplement for Challenges and Future Directions}
\label{appendix_sec-challenge-and-future-directions}

This section serves as a complementary resource to Section~\ref{Sec_Challenges and Opportunities}. It highlights several important challenges that merit further discussion and attention. Due to space limitations in the main text, we provide a more detailed analysis here to ensure a more thorough and nuanced treatment of these issues.

\noindent \textbf{(d) Gradient Propagation.}
Discrete quantization breaks the differentiability of neural networks, making it difficult to propagate gradients through discrete latent variables. To enable end-to-end training, common approximations such as the Straight-Through Estimator (STE)~\cite{2013_arXiv_STE, 2017_NeurlPS_VQ-VAE_Neural-discrete-representation-learning} and Gumbel-Softmax~\cite{2016_arXiv_Gumbel-Softmax, 2019_arXiv_2020_ICLR_vq-wav2vec_vq-wav2vec-Self-supervised-learning-of-discrete-speech-representations} are widely adopted. However, these methods can introduce estimation bias, gradient variance, and convergence instability, especially in complex downstream tasks.
An alternative is to design principled, stable gradient approximations for discrete token spaces. Promising approaches include score-based estimators, hybrid relaxations, and RL-inspired methods aligned with token selection. Task-specific gradient flows and regularization may further improve robustness and generalization.

\noindent \textbf{(e) Granularity and Semantic Alignment.}
Balancing token granularity is crucial: coarse tokens may miss details, while overly fine-grained ones inflate sequence length and cost~\cite{2022_CVPR_RQ-VAE_RQ-Transformer_Autoregressive-image-generation-using-residual-quantization, 2023_CVPR_DQ-VAE_Towards-accurate-image-coding-Improved-autoregressive-image-generation-with-dynamic-vector-quantization, 2025_arXiv_Spark-TTS_Spark-TTS=An-Efficient-LLM-based-Text-to-speech-Model-with-Single-stream-Decoupled-Speech-Tokens}. Existing methods also struggle to align with semantic boundaries, especially in continuous modalities such as image or audio, where meaningful units are often ambiguous or task-specific~\cite{2024_arXiv_2025_CVPR_TokenFLow_TokenFlow=Unified-Image_Tokenizer-for-Multimodal, 2023_CVPR_MQ-VAE_Not-all-image-regions-matter=Masked-vector-quantization-for-autoregressive-image-generation, 2025_arXiv_SemHiTok_SemHiTok=A-unified-Image-Tokenizer-via-Semantic-guided-Hierarchical-Codebook-for-Multimodal-Understanding-and-Generation}.
To address these issues, promising directions include adaptive and hierarchical quantization that modulates granularity based on content complexity and semantics. Techniques, such as dynamic masking, multi-scale encoding, and attention-guided segmentation, may better align tokens with structure, leading to more efficient and interpretable representations.

\noindent \textbf{(f) Modality and Task Transferability.}
Numerous tokenizers are handcrafted or domain-specific, limiting their applicability across tasks. A promising direction is to develop generalizable tokenization methods that work across domains. This could be achieved through cross-modal pretraining or unified discrete spaces~\cite{2024_ACL_AnyGPT_AnyGPT=Unified-Multimodal-LLM-with-Discrete-Sequence-Modeling, 2024_ICML_VideoPoet_VideoPoet=A-Large-Language-Model-for-Zero-shot-Video-Generation, 2024_CVPR_Unified-IO-2_Unified-IO-2=Scaling-Autoregressive-Multimodal-Models-with-Vision-Language-Audio-and-Action}, enabling tokenization to function seamlessly across multiple data types.
While several works have demonstrated success in aligning two modalities such as text and image, efforts to support three~\cite{2023_arXiv_TEAL_TEAL=Tokenize-and-Embed-All-for-Multi-modal-Large-Language-Models, 2024_ACL_AnyGPT_AnyGPT=Unified-Multimodal-LLM-with-Discrete-Sequence-Modeling, 2024_arXiv_DMLM_Discrete-Multimodal-Transformers-with-a-Pretrained-Large-Language-Model-for-Mixed-supervision-Speech-Processing, 2024_arXiv_Emu3_Emu3=Next-token-Prediction-is-All-You-Need, 2024_arXiv_2025_ICLR_VILA-U_VILA-U=A-Unified-Foundation-Model-Integrating-Visual-Understanding-and-Generation, 2025_ICLR_LWM_World-Model-on-Million-length-Video-and-Language-with-Blockwise-Ringattention, 2025_ICCGV_LLM-Gesticulator_LLM-Gesticulator=Leveraging-Large-Language-Models-for-Scalable-and-Controllable-Co-speech-Gesture-Synthesis} or more~\cite{2024_ICML_VideoPoet_VideoPoet=A-Large-Language-Model-for-Zero-shot-Video-Generation, 2024_CVPR_Unified-IO-2_Unified-IO-2=Scaling-Autoregressive-Multimodal-Models-with-Vision-Language-Audio-and-Action} modalities in a unified discrete representation remain rare, presenting a compelling direction for future research on multimodal tokenization.

\noindent \textbf{(g) Interpretability and Controllability.}
Learned tokens often lack transparency, making them difficult to interpret and control. This is a significant challenge for applications requiring human-understandable representations. Prior work using discrete latent spaces has shown that while tokens can capture meaningful patterns, they often do not correspond to interpretable or manipulable concepts~\cite{2023_ICML_Muse_Muse=Text-to-image-Generation-via-Masked-Generative-Transformers, 2023_CVPR_NUWA-LIP_NUWA-LIP=Language-guided-Image-Inpainting-with-Defect-free-VQGAN, 2025_ICLR_NID_Node-Identifiers=Compact-Discrete-Representations-for-Efficient-Graph-Learning}. 
Future directions include aligning discrete tokens with human-centric semantics to enhance transparency and usability. This may involve concept-grounded codebooks, token-level editing, or interpretable priors during training. Improving interpretability can also help with debugging, personalization, and safety in downstream tasks.






\end{document}